%% file: Papadopoulos-BDS.tex
\documentclass[conference]{IEEEtran}
\usepackage[utf8]{inputenc}
\usepackage[english]{babel}
\usepackage{todonotes}
\usepackage{soul}
\usepackage{cite}

\usepackage{graphicx}     
\usepackage{caption}      
\usepackage{subcaption}   
\usepackage{balance}
\usepackage{tabularx}
\usepackage{url}

\usepackage{booktabs}
\newcolumntype{Y}{>{\centering\arraybackslash}X} 

\usepackage{multirow}
\usepackage{amssymb}      
\usepackage{amsmath,amsfonts}
\usepackage{textcomp}     

\usepackage[
colorlinks=true,
allcolors=blue
]{hyperref}               

\usepackage{pgf}

\everymath=\expandafter{\the\everymath\displaystyle}
\makeatletter\@ifpackageloaded{underscore}{}{\usepackage[strings]{underscore}}\makeatother

\definecolor{figurered}{RGB}{255, 0, 0}
\definecolor{figureblue}{RGB}{0, 16, 255}
\definecolor{figuregreen}{RGB}{4, 128, 0}

\begin{document}

\title{X-ray illicit object detection using hybrid CNN-transformer neural network architectures\thanks{The research leading to these results has received funding from the European Union’s Horizon Europe research and innovation programme under grant agreement No 101073876 (Ceasefire). This publication reflects only the authors views. The European Union is not liable for any use that may be made of the information contained therein.}}

\author{
  \IEEEauthorblockN{
    Jorgen Cani\IEEEauthorrefmark{1},
    Christos Diou\IEEEauthorrefmark{1},
    Spyridon Evangelatos\IEEEauthorrefmark{2},
    Panagiotis Radoglou-Grammatikis\IEEEauthorrefmark{3}\IEEEauthorrefmark{4},
    Vasileios Argyriou\IEEEauthorrefmark{5},\\
    Panagiotis Sarigiannidis\IEEEauthorrefmark{3},
    Iraklis Varlamis\IEEEauthorrefmark{1} and
    Georgios Th. Papadopoulos\IEEEauthorrefmark{1}
  }

  \IEEEauthorblockA{
    \IEEEauthorrefmark{1}
    Department of Informatics and Telematics, Harokopio University of Athens, Athens, Greece\\
  }

    \IEEEauthorblockA{
    \IEEEauthorrefmark{2}
    Research \& Innovation Development Department, Netcompany-Intrasoft S.A., Luxembourg, Luxembourg\\
  }

  \IEEEauthorblockA{
    \IEEEauthorrefmark{3}
    Department of Electrical and Computer Engineering, University of Western Macedonia, Kozani, Greece\\
  }

  \IEEEauthorblockA{
    \IEEEauthorrefmark{4}
    K3Y Ltd, Sofia, Bulgaria\\
  }

  \IEEEauthorblockA{
    \IEEEauthorrefmark{5}
    Department of Networks and Digital Media, Kingston University, London, United Kingdom\\
  }

  \IEEEauthorblockA{
    Email: \IEEEauthorrefmark{1}\{cani,cdiou,varlamis,g.th.papadopoulos\}@hua.gr,
    \IEEEauthorrefmark{2}sevangelatos@netcompany.com,\\
    \IEEEauthorrefmark{3}Vasileios.Argyriou@kingston.ac.uk,
    \IEEEauthorrefmark{4}psarigiannidis@uowm.gr,
  \IEEEauthorrefmark{5}pradoglou@k3y.bg
  }

}
\IEEEoverridecommandlockouts

\maketitle

\IEEEpubidadjcol

\maketitle
\begin{abstract}
In the field of X-ray security applications, even the smallest details can significantly impact outcomes. Objects that are heavily occluded or intentionally concealed pose a great challenge for detection, whether by human observation or through advanced technological applications. While certain Deep Learning (DL) architectures demonstrate strong performance in processing local information, such as Convolutional Neural Networks (CNNs), others excel in handling distant information, e.g., transformers. In X-ray security imaging the literature has been dominated by the use of CNN-based methods, while the integration of the two aforementioned leading architectures has not been sufficiently explored. In this paper, various hybrid CNN-transformer architectures are evaluated against a common CNN object detection baseline, namely YOLOv8. In particular, a CNN (HGNetV2) and a hybrid CNN-transformer (Next-ViT-S) backbone are combined with different CNN/transformer detection heads (YOLOv8 and RT-DETR). The resulting architectures are comparatively evaluated on three challenging public X-ray inspection datasets, namely EDS, HiXray, and PIDray. Interestingly, while the YOLOv8 detector with its default backbone (CSP-DarkNet53) is generally shown to be advantageous on the HiXray and PIDray datasets, when a domain distribution shift is incorporated in the X-ray images (as happens in the EDS datasets), hybrid CNN-transformer architectures exhibit increased robustness. Detailed comparative evaluation results, including object-level detection performance and object-size error analysis, demonstrate the strengths and weaknesses of each architectural combination and suggest guidelines for future research. The source code and network weights of the models employed in this study are available at \url{https://github.com/jgenc/xray-comparative-evaluation}.

\end{abstract}

\begin{IEEEkeywords}
Object detection, X-ray imaging, convolutional neural networks, vision transformers, hybrid architectures
\end{IEEEkeywords}


\section{Introduction}

Automated X-ray screening for prohibited item detection constitutes a crucial task for public safety and yet remains open to several challenges, such as object occlusion, cluttered scenes, high intra-class variation, and the inherent visual ambiguity of grayscale X-ray imagery \cite{zhang2023pidray}. Under these conditions, relying on manual inspection can lead to fatigue and errors in high-throughput environments. In recent years, deep neural networks have driven breakthroughs in generic object detection. Consequently, deep learning methods, such as Convolutional Neural Networks (CNNs), have introduced notable advances in automatic object detection in X-ray images \cite{batsis2023illicit}\cite{rafiei2023computer}\cite{mademlis2023visual}. 

In parallel, Vision Transformers (ViTs) have shown increased capabilities in modeling global visual context, by attending to whole image patches via multi-head self-attention \cite{dosovitskiy2020image}\cite{cani2024illicit}. In particular, DETR \cite{carion2020end} formulates the task of object detection as a set prediction problem. Additionally, sparse DETR \cite{roh2021sparse} employs a Swin transformer backbone and adjusts accordingly token usage across various configurations. Moreover, DINO \cite{zhang2022dino} introduces contrastive denoising, mixed query anchors, and dual look-ahead box prediction, along with test time augmentation \cite{alimisis2025advances}. 

More recently, hybrid CNN-transformer architectures have been introduced, which combine the complementary merits of CNNs and ViTs to address various computer vision tasks more efficiently \cite{khan2023survey}. In particular, when CNNs' excellence in capturing local detail is combined with ViTs' ability to capture long-range relationships across the visual scene, promising object detection performance can be observed \cite{hendria2023combining}. In this context, Next-ViT-S \cite{li2022next} comprises a leading example that alternates specialized convolution blocks with attention-based ones at each stage and outperforms comparable CNNs and ViTs on standard benchmarks. More specifically, Next-ViT-S retains the inductive biases and speed of convolution, while incorporating global receptive fields through the use of self-attention; Moreover, it is explicitly engineered for efficient deployment, thus being an ideal backbone network for object detection tasks in realistic scenarios.

Despite the rise of ViTs and hybrid CNN-transformer architectures in natural image analysis, the X-ray imaging community still focuses on CNN-based approaches \cite{wu2023object}. This is mainly because ViT architectural components often require large training datasets and can be computationally heavy. To this end, the advantageous characteristics of hybrid CNN-transformer architectures have not yet been sufficiently explored in X-ray security imaging \cite{rafiei2023computer}.

In this paper, the issue of applying hybrid CNN-transformer architectures for the detection of illicit objects \cite{mademlis2024invisible} in X-ray inspection imaging is investigated, aiming at examining their performance against the typical CNN-only detection baseline. In particular, various hybrid architectures are formed by combining a CNN (HGNetV2 \cite{HGNetv2}) and a hybrid CNN-transformer (Next-ViT-S \cite{li2022next}) backbone with different CNN/transformer detection heads (namely, YOLOv8 \cite{Jocher_Ultralytics_YOLO_2023} and RT-DETR \cite{zhao2024detrs}, respectively). The formed architectures are comparatively evaluated against a common and well-performing CNN-only object detection baseline, namely YOLOv8 with its default backbone (CSP-DarkNet53) \cite{Jocher_Ultralytics_YOLO_2023}. All methods are evaluated on three challenging public X-ray inspection datasets, namely EDS \cite{tao2022exploring}, HiXray \cite{tao2021towards}, and PIDray \cite{wang2021towards}. Compared to HiXray and PIDray, EDS introduces a distributional shift in the collected data, due to the use of three different X-ray scanners during the capturing process, simulating in this way real-world deployment challenges and emphasizing on the evaluation of models' robustness. Interestingly, the YOLOv8 detector performs better in the HiXray and PIDray datasets; however, hybrid architectures are shown to be advantageous in the EDS one. Moreover, detailed comparative evaluation results, also including object-level detection performance and object-size error analysis, demonstrate the pros and cons of each architectural combination and suggest guidelines for future research.

The remainder of the paper is organized as follows: Section \ref{sec:relwork} presents an overview of previous work in the field of DL-based X-ray object detection. Section \ref{sec:methodology} details the hybrid CNN-transformer neural network architectures considered in this work. Section \ref{sec:experimental-results} discusses the
obtained experimental results and highlights key findings. Section \ref{sec:conclusions} draws conclusions and outlines future research directions in the field.

\section{Previous work}
\label{sec:relwork}

Automatic X-ray security inspection systems typically rely on the use of DL-based object detection methods that, so far, have been predominantly based on CNN architectures. In particular, notable early studies, such as DOAM \cite{wei2020occluded} and LIM \cite{tao2021towards}, adapt popular CNN-based object detectors, to suit the characteristics of the X-ray imaging domain. On another direction, detection heads such as SSD, FCOS, YOLOv3 and YOLOv5 have also been explored \cite{wu2023object}, aiming at addressing real-time analysis requirements. More elaborate approaches follow the respective advancements in generic CNN-based detection schemes. In particular, EM-YOLO \cite{jing2023yolo} enhances YOLOv7 with X-ray specific pre-processing for handling occlusion, low contrast, and class imbalance. SC-YOLOv8 \cite{han2023sc} modifies the backbone to adapt to object position and shape. Additionally, Wang et al. \cite{wang2024lightweight} propose two YOLOv8 variants with changes to the neck and head networks, maintaining size while boosting performance. YOLOv8-GEMA \cite{wang2024improved} adjusts the backbone and neck for improved results. Moreover, TinyRay \cite{zhang2025lightweight} uses a YOLOv7 variant with the lightweight FasterNet backbone, while Ren et al. \cite{ren2025feature} apply distillation to train compact YOLOv4 and RetinaNet models.

Following the application of hybrid CNN-transformer architectures to natural RGB images, the same methods have been used in X-ray-based illicit object detection. In particular, the Trans2ray \cite{meng2024transformer} approach comprises a dual-branch hybrid framework for dual-view X-ray imaging, achieving notable performance against other dual-view detectors. Additionally, the EslaXDET \cite{wu2024eslaxdet} method applies self-supervised training \cite{konstantakos2025self} on a ViT and introduces a detection head for creating multi-scale feature maps.

Although hybrid CNN-transformer architectures show promise, they remain less adopted in X-ray inspection compared to conventional CNNs. Their limited uptake is mainly due to challenges like increased computational overhead, which is critical for high-throughput systems. In this context, lightweight adaptations, like TinyViT \cite{wu2022tinyvit} and Efficient Hybrid DETR \cite{zhao2024detrs}, explore parameter reduction strategies without compromising performance, while leveraging Neural Architecture Search (NAS) for balancing accuracy and latency aspects. On the other hand, the demand of transformer-based components for increased training datasets (compared to the respective CNN case) introduces further concerns and obstacles. In this respect, the lack of sufficiently large X-ray datasets and the presence of biases in the publicly available ones (e.g. due to the inherent scarcity of security threats in publicly annotated benchmarks), require the development of appropriate techniques, like semi-supervised learning and synthetic data augmentation. Towards this direction, the approaches of Lin et al. \cite{lin2025detection} and Huang et al. \cite{huang2024adaptxray} demonstrate how hybrid architectures benefit from cross-domain pretraining and attention-based few-shot learning, respectively, reducing in that way the reliance on large labeled datasets.

\section{Hybrid CNN-transformer-based object detection}
\label{sec:methodology}

This section details the hybrid CNN-transformer architectures investigated in this work, providing the basis for their performance comparison against a conventional CNN-only detection baseline. In particular, various hybrid architectures are formed by combining recent and well-performing CNN (HGNetV2 \cite{HGNetv2}) and hybrid CNN-transformer (Next-ViT-S \cite{li2022next}) backbones with different CNN/transformer detection heads (namely, YOLOv8 \cite{Jocher_Ultralytics_YOLO_2023} and RT-DETR \cite{zhao2024detrs}, respectively). Denoting each formed object detector as D(head, backbone), the considered composite architectures in this work are as follows: D(YOLOv8, Next-ViT-S), D(RT-DETR, HGNetV2), and D(RT-DETR, Next-ViT-S). The latter are comparatively evaluated against a common and well-performing CNN-only object detection baseline, namely D(YOLOv8, CSP-DarkNet53), i.e. the YOLOv8 detector with its default CSP-DarkNet53 backbone \cite{Jocher_Ultralytics_YOLO_2023}.

In the following sections, the architectural modifications to the key neural components (namely, detectors YOLOv8 and RT-DETR, and backbone Next-ViT-S) that are required for forming the D(YOLOv8, Next-ViT-S) and D(RT-DETR, Next-ViT-S) object detectors are detailed.

\subsection{Next-ViT backbone network}

Next-ViT \cite{li2022next} is a hybrid model combining convolutional and transformer network components, designed so as to achieve an optimal balance between latency and accuracy. It adopts a hierarchical pyramidal architecture, which comprises multiple stages that are adaptable for various downstream tasks, including object detection and segmentation. Next-ViT includes an initial stem stage (incorporating standard convolutional layers), followed by four primary ones (labeled $S_1$ to $S_4$), where the feature output size is subsequently reduced by half at each stage. Each primary stage incorporates Next Convolutional Blocks (NCB) and Next Transformer Blocks (NTB), in order to model both short- and long-term dependencies within the input visual data. In the current study, the smallest \textit{Next-ViT-S} configuration is considered for ensuring real-time performance, using model weights\footnote{\url{https://github.com/bytedance/Next-ViT?tab=readme-ov-file\#image-classification}} pre-trained on the ImageNet dataset.

\subsection{Hybrid YOLOv8-based object detector}
\label{sub-sec:yolov8-mod}

In order to form the D(YOLOv8, Next-ViT-S) object detector, the CSP-DarkNet53 backbone, included in the standard YOLOv8 architecture, is substituted by the Next-ViT-S one. This requires modifications to the connections of the original YOLOv8 neck network, in order to handle the varying feature map sizes and spatial resolutions of the intermediate layers of the Next-ViT-S backbone.

\begin{figure}
    \centering
    \includegraphics[width=1\linewidth]{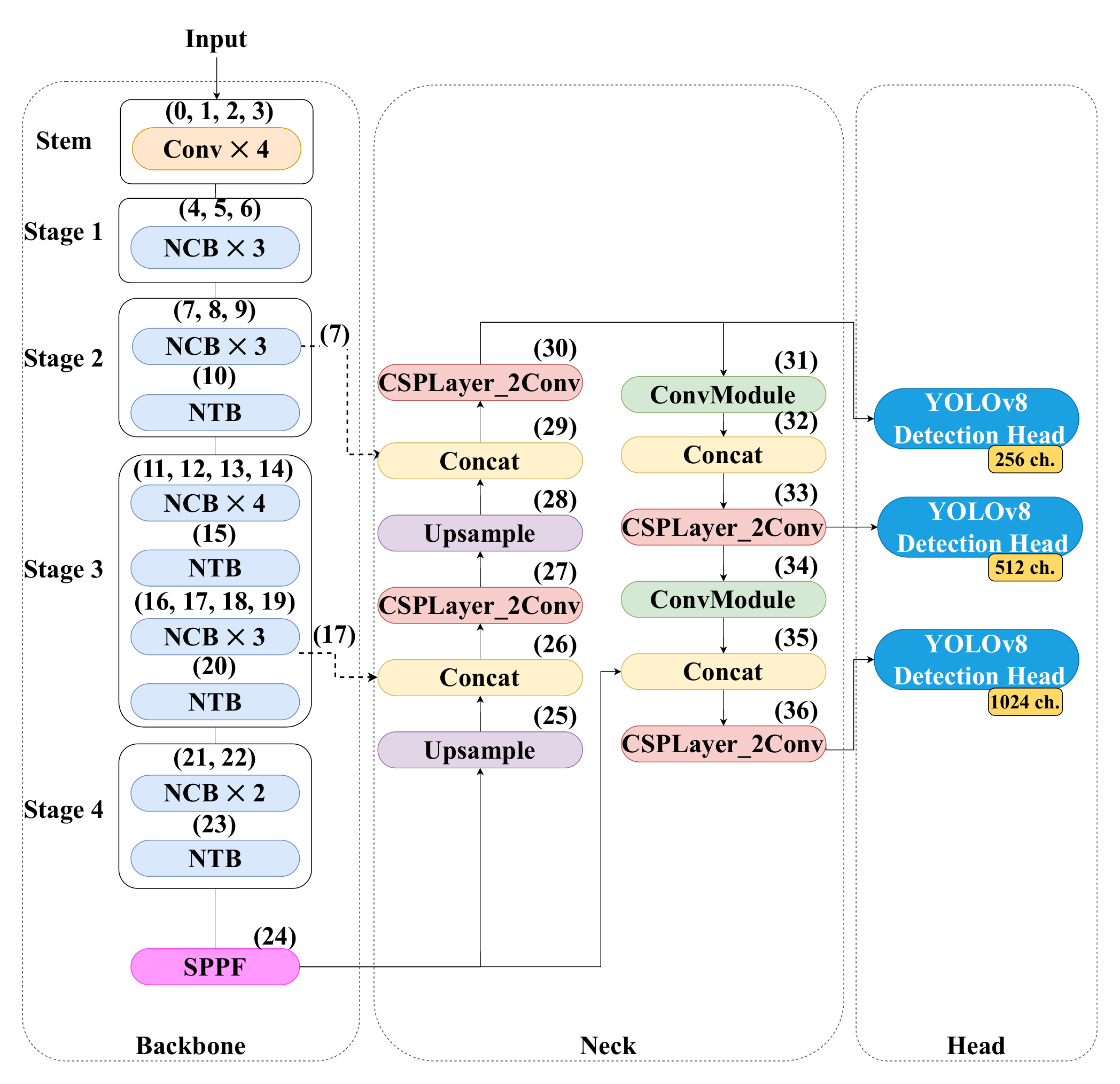}
    \caption{Architecture of the D(YOLOv8, Next-ViT-S) detector.}
    \label{fig:nextvit-yolov8}
\end{figure}

The default YOLOv8 implementation (with the CSPDarkNet-53 backbone) utilizes multi-scale feature maps that are organized in five stages, labeled $P_1$ to $P_5$. In order to effectively utilize these features, YOLOv8 incorporates skip connections (known as residual links) from the backbone to the neck module, a process facilitated by the so called Path Aggregation Network (PAFPN); these skip connections merge and upscale feature maps. By convention, the output of each backbone block is referred to as a `layer', numbered sequentially throughout the network, in order to differentiate from the $P_k$ feature levels. YOLOv8 utilizes connections to layers $4$ and $6$, corresponding to the outputs of the layers before $P_3$ and $P_4$. In order to integrate the Next-ViT-S backbone to the YOLOv8 detector head, three different combinations of skip connections are considered that are denoted $C(x, y)$, where $x$ is the index of the first skip layer and $y$ the next one, and graphically illustrated in Fig. \ref{fig:nextvit-yolov8}:
\begin{itemize}
    \item $C(10, 20)$: Layer 10 represents the first transformer block (NTB) in $S_2$, incorporating both basic visual features and broad contextual information. Layer 20, on the other hand, is the third NTB block in $S_3$, which bears more profound features.
    \item $C(9, 19)$: Layer 9 is the last convolutional block (NCB) in $S_2$, possessing basic visual information, but lacking comprehensive contextual information. Layer 19 is the last NCB block in $S_3$, now integrating contextual information from the preceding NTB layers.
    \item $C(7, 17)$: Layer 7 is the first NCB block in $S_2$, incorporating features from $S_1$. Layer 17 is located two layers after the second NTB block, bearing contextual information that has not yet been refined by subsequent NCB layers.
\end{itemize}

The final output features of the Next-ViT-S backbone are processed by the YOLOv8 SPPF layer (a modified version of the SPP-Network) and then by the original neck network. After extensive experimental evaluation, configuration $C(7, 17)$ was shown to lead to superior detection performance and utilized in all experiments in this study.

\subsection{Hybrid RT-DETR-based object detector}

\begin{figure}
    \centering
    \includegraphics[width=1\linewidth]{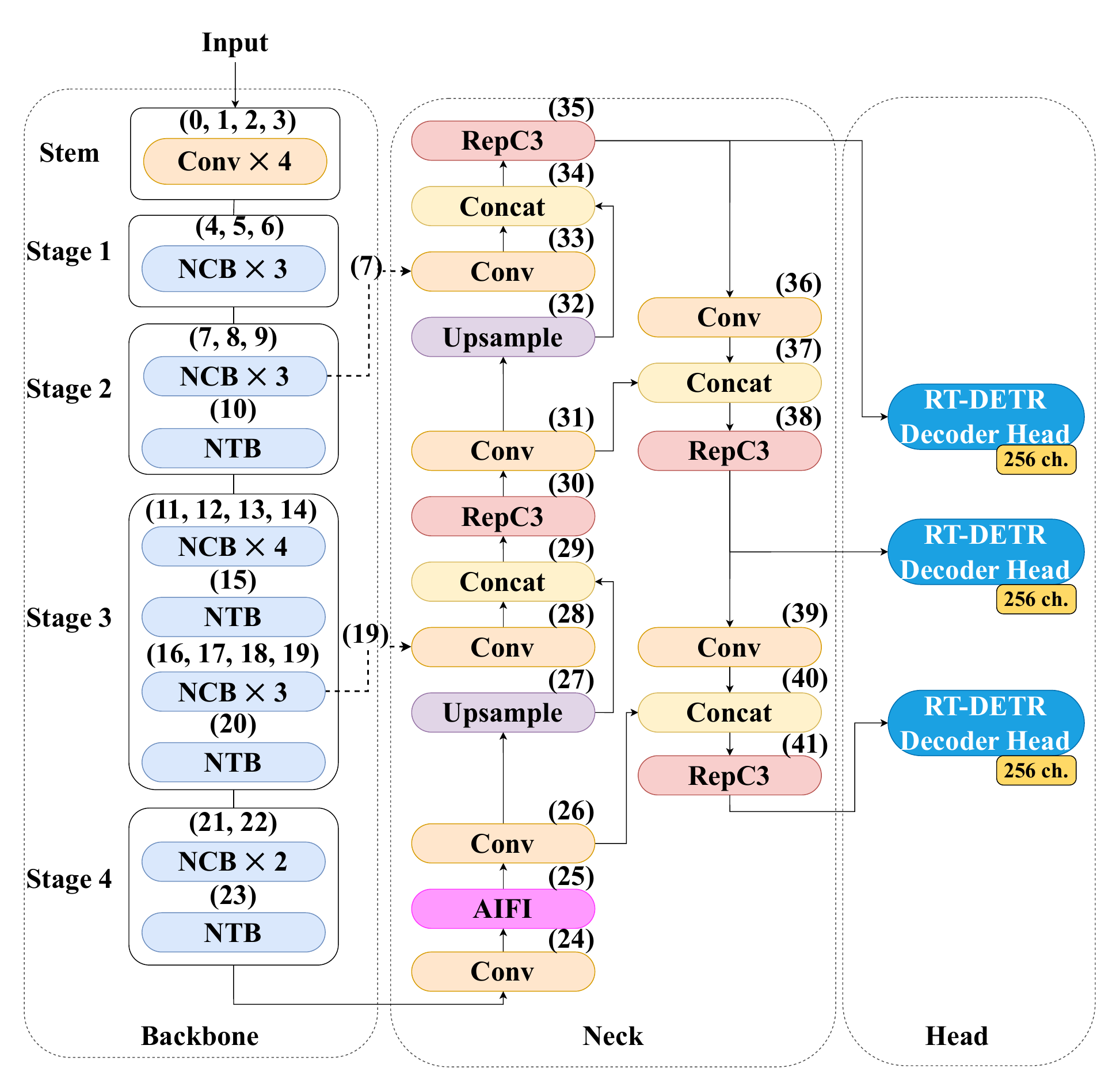}
    \caption{Architecture of the D(RT-DETR, Next-ViT-S) detector.}
    \label{fig:nextvit-rtdetr}
\end{figure}

In order to formulate the D(RT-DETR, Next-ViT-S) detector, the default backbone (HGNetV2 \cite{HGNetv2}) of RT-DETR \cite{zhao2024detrs} is replaced by the Next-ViT-S one. The latter requires modifications to the connections of the original RT-DETR neck network towards the backbone for accounting for the different feature map scales and spatial resolutions of Next-ViT-S. In particular, the CNN-only HGNetV2 network comprises a hierarchical architecture with four stages, denoted `stage 1-4'. Two residual connections are established from the backbone (stages 2 and 3) to the detection head (layers 3 and 7). In order to integrate the Next-ViT-S backbone to the RT-DETR detection head, two skip connections of different feature scales are defined from the backbone to the neck network. Similarly to \ref{sub-sec:yolov8-mod}, the same skip connections $C(10, 20)$, $C(9, 19)$, and $C(7, 17)$ are again considered. The head network, consisting of three RT-DETR decoder heads, remains unchanged. The overall D(RT-DETR, Next-ViT-S) architecture is illustrated in Fig. \ref{fig:nextvit-rtdetr}. Following thorough experimental assessment, configuration $C(9, 19)$ was shown to be the most efficient one and was used in all experiments carried out in this study.

\section{Experimental results}
\label{sec:experimental-results}

This section details the defined experimental framework for evaluating the performance of the considered object detectors, while the obtained results, along with their corresponding assessment discussion, are subsequently provided.

\subsection{Experimental framework}
\label{sub-sec:experimental-setup}

\subsubsection{Datasets}

\begin{table*}[t]
  \centering
  \caption{Object detection results for the EDS, HiXray and PIDray datasets (left column: mAP$^{50}$, right column: mAP$^{50:95}$)}
  \begin{tabular}{l *{3}{c}}
    \toprule
    \textbf{Detector}          & \textbf{EDS}           & \textbf{HiXray}        & \textbf{PIDray}        \\
    \midrule
    D(YOLOv8, CSP-DarkNet53) & 0.547 / 0.386          & 0.845 / \textbf{0.564} & 0.897 / \textbf{0.807} \\
    D(YOLOv8, Next-ViT-S)     & 0.588 / 0.408          & 0.841 / 0.551          & 0.898 / 0.801          \\
    \midrule
    D(RT-DETR, HGNetV2)      & 0.573 / \textbf{0.410} & 0.839 / 0.510          & 0.835 / 0.720          \\
    D(RT-DETR, Next-ViT-S)    & 0.504 / 0.322          & 0.818 / 0.483          & 0.879 / 0.773          \\
    \bottomrule
  \end{tabular}
  \label{tab:results}
\end{table*}

\begin{table*}[t]
  \centering
  \caption{Object detection results for the various sessions of the EDS dataset (left column: mAP$^{50}$, right column: mAP$^{50:95}$)}
  \begin{tabular}{>{\raggedright\arraybackslash}p{3.7cm} *{7}{c}}
    \toprule
    \textbf{Detector}        & \boldmath$\mathcal{D}_{1\rightarrow2}$ & \boldmath$\mathcal{D}_{1\rightarrow3}$ & \boldmath$\mathcal{D}_{2\rightarrow1}$ & \boldmath$\mathcal{D}_{2\rightarrow3}$ & \boldmath$\mathcal{D}_{3\rightarrow1}$ & \boldmath$\mathcal{D}_{3\rightarrow2}$ & \textbf{Avg.}          \\
    \midrule
    D(YOLOv8, CSP-DarkNet53) & 0.482 / 0.340                          & 0.555 / 0.410                          & 0.454 / 0.295                          & 0.619 / 0.449                          & 0.587 / 0.411                          & 0.590 / 0.411                          & 0.547 / 0.386 \\
    D(YOLOv8, Next-ViT-S)    & 0.512 / 0.347                          & 0.603 / \textbf{0.441}                 & 0.515 / 0.341                          & 0.648 / 0.454                          & 0.624 / \textbf{0.434}                 & 0.626 / \textbf{0.431}                 & 0.588 / 0.408 \\
    D(RT-DETR, HGNetV2)      & 0.506 / \textbf{0.352}                 & 0.569 / 0.424                          & 0.506 / \textbf{0.350}                 & 0.648 / \textbf{0.471}                 & 0.595 / 0.431                          & 0.616 / 0.429                          & 0.573 / \textbf{0.410} \\
    D(RT-DETR, Next-ViT-S)   & 0.446 / 0.292                          & 0.545 / 0.343                          & 0.372 / 0.217                          & 0.450 / 0.286                          & 0.578 / 0.377                          & 0.636 / 0.419                          & 0.504 / 0.322 \\
    \bottomrule
  \end{tabular}
  \label{tab:results-eds}
\end{table*}

\begin{table*}[t]
  \centering
  \caption{Object detection results for the various subsets of the PIDray dataset (left column: mAP$^{50}$, right column: mAP$^{50:95}$)}
  \begin{tabular}{l *{4}{c}}
    \toprule
    \textbf{Detector}   & \textbf{easy}          & \textbf{hard}          & \textbf{hidden}        & \textbf{overall}       \\
    \midrule
    D(YOLOv8, CSP-DarkNet53) & 0.911 / \textbf{0.846} & 0.914 / \textbf{0.812} & 0.797 / 0.682          & 0.897 / \textbf{0.807} \\
    D(YOLOv8, Next-ViT-S)     & 0.912 / 0.837          & 0.910 / 0.799          & 0.803 / \textbf{0.685} & 0.898 / 0.801          \\
    \midrule
    D(RT-DETR, HGNetV2)      & 0.864 / 0.780          & 0.864 / 0.724          & 0.681 / 0.548          & 0.835 / 0.720          \\
    D(RT-DETR, Next-ViT-S)    & 0.898 / 0.824          & 0.898 / 0.770          & 0.779 / 0.646          & 0.879 / 0.773          \\
    \bottomrule
  \end{tabular}
  \label{tab:results-pidray}
\end{table*}

The EDS \cite{tao2022exploring} dataset focuses on the challenge of domain shift that is inherent in X-ray imaging, due to factors like varying parameters across different scanning devices. In particular, three different X-ray scanners are employed, resulting into variations in the captured color, depth and texture information channels, mainly introduced by the different device specs and wear levels. The packages used during the scanning process were artificially prepared. EDS supports ten classes of common daily-life objects, namely \textit{Plastic bottle (DB)}, \textit{Knife (KN)}, \textit{Scissor (SC)}, \textit{Laptop (LA)}, \textit{Umbrella (UM)}, \textit{Lighter (LI)}, \textit{Device (SE)}, \textit{Power bank (PB)}, \textit{Pressure (PR)}, and \textit{Glass bottle (GB)}. The dataset comprises $14,219$ images containing $31,654$ object instances from three domains (X-ray machines), resulting in $\sim2.22$ instances per image on average. The defined experimental protocol dictates the training of a detection model in a single domain and its subsequent evaluation in a different one, resulting in a total of six performed experimental sessions.

The HiXray \cite{tao2021towards} dataset contains real-world X-ray scans collected from an international airport, where the image annotations were provided by the airport security personnel. The dataset comprises $45,364$ images that include a total of $102,928$ prohibited items, i.e. $\sim2.27$ instances per image. The dataset supports eight classes, namely \textit{Portable charger 1 (lithium-ion prismatic cell) (PO1)}, \textit{Portable charger 2 (lithium-ion cylindrical cell) (PO2)}, \textit{Tablet (TA)}, \textit{Mobile phone (MP)}, \textit{Laptop (LA)}, \textit{Cosmetic (CO)}, \textit{Water (WA)}, and \textit{Nonmetallic Lighter (NL)}. HiXray is split into a training ($80\%$ of images) and a test ($20\%$ of images) set.

The PIDray \cite{wang2021towards} dataset focuses on deliberately hidden items, mimicking real-world scenarios where prohibited objects are intentionally concealed. The latter fact adds an extra level of complexity to the object detection task, since it is required to identify hidden items (and not `simply' detecting objects obscured by other items and/or environmental factors). All scan samples are collected under real-world settings, namely at airport, subway, and railway station security checkpoints. PIDray includes twelve classes of prohibited items, namely \textit{Baton (BA)}, \textit{Pliers (PL)}, \textit{Hammer (HA)}, \textit{Power-bank (PB)}, \textit{Scissors (SC)}, \textit{Wrench (WR)}, \textit{Gun (GU)}, \textit{Bullet (BU)}, \textit{Sprayer (SP)}, \textit{Handcuffs (HC)}, \textit{Knife (KN)}, and \textit{Lighter (LI)}. The dataset is split into a training ($29,457$ samples, $\sim60\%$ of images) and a test ($18,220$ samples, $\sim40\%$ of images) set. Moreover, the test set is further divided into three sub-sets, namely an easy (the images contain only one prohibited object), a hard (the images contain more than one illicit items), and a hidden (the images contain deliberately hidden objects) one, with $9,482$, $3,733$, and $5,055$ images, respectively.

\subsubsection{Performance metrics}

Mean Average Precision (mAP) constitutes the most commonly used metric in object detection applications, which estimates a comprehensive and aggregated evaluation of the examined model's performance across different confidence levels and object classes. Among the different variants regarding how mAP is calculated, especially with respect to the selected  Intersection over Union (IoU) threshold (that assesses the spatial overlap between the predicted bounding box (generated by a detector model) and the corresponding ground truth one (that defines the actual location of the object) for determining true positive detection, the following ones have been considered in this work: a) mAP$^{50}$: This refers to the mAP score calculated using an IoU threshold of $0.5$. b) mAP$^{50:95}$: This involves a more rigorous evaluation protocol, which calculates mAP by averaging AP scores over a range of defined IoU thresholds, typically from $0.5$ to $0.95$ with a step of $0.05$, and subsequently averaging the computed results across all object classes. This metric provides a more comprehensive assessment of the model's localization accuracy, by considering its performance at different levels of overlap with the ground truth annotation. In general, higher mAP scores, which receive values ranging from $0$ to $1$, indicate better performance.

\subsubsection{Implementation details}

All reported experiments were conducted using a PC with Ubuntu Linux 22.04 OS, equipped with an Intel Core i9-13900K CPU and two NVIDIA GeForce RTX 4070 Ti GPUs. For the YOLOv8 and RT-DETR detectors the implementations and layer weights available in the Ultralytics framework\footnote{\url{https://github.com/ultralytics/ultralytics}} were used, while for the Next-ViT-S\footnote{\url{https://github.com/bytedance/next-vit}} the respective publicly available source code and weights were also utilized. Regarding hyperparameter setting, all detectors were trained using a batch size equal to $18$. For the EDS dataset, D(RT-DETR, Next-ViT-S) was trained using Stochastic Gradient Descent (SGD) with a learning rate of $0.01$ and a momentum of $0.937$, while for the remaining detectors the AdamW optimizer was employed with a learning rate of $0.000714$. For the HiXray and PIDray datasets, due to their larger size, all detectors were trained using SGD with a learning rate of $0.01$ and a momentum of $0.937$. All experiments incorporated an early stopping mechanism to prevent overfitting.

\subsection{Evaluation results and discussion}

\begin{figure} [t]
\centering
\begin{tabular}{p{0.01\linewidth}c}

    \begin{tabular}{c}\end{tabular}& 
    \begin{tabular}{c}
        \begin{subfigure}[t]{0.25\textwidth}
        \centering
        \includegraphics[width=\linewidth]{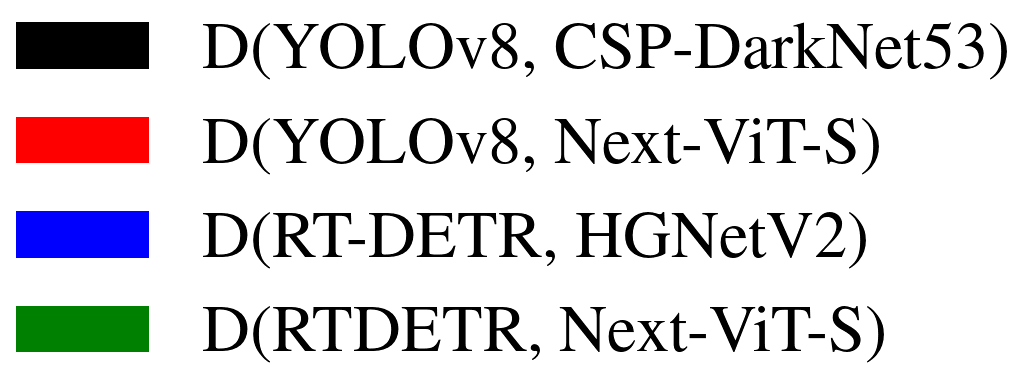}
        \end{subfigure}
    \end{tabular}\\

    \begin{tabular}{c}a)\end{tabular}& 
    \begin{tabular}{c}
        \begin{subfigure}[t]{0.35\textwidth}
        \centering
        \resizebox{\linewidth}{!}{\input{./EDS-classwise.pgf}}
        \end{subfigure}
    \end{tabular}\\

    \begin{tabular}{c}b)\end{tabular}& 
    \begin{tabular}{c}
        \begin{subfigure}[t]{0.35\textwidth}
        \centering
        \resizebox{\linewidth}{!}{\input{./HIXray-classwise.pgf}}
        \end{subfigure}
    \end{tabular}\\

    \begin{tabular}{c}c)\end{tabular}& 
    \begin{tabular}{c}
        \begin{subfigure}[t]{0.35\textwidth}
        \centering
        \resizebox{\linewidth}{!}{\input{./PIDray-classwise.pgf}}
        \end{subfigure}
    \end{tabular}\\
 
\end{tabular}
\caption{Object-level performance (mAP$^{50:95}$ metric) for datasets: a) EDS, b) HiXray, and c) PIDray.}
\label{fig:object-level_figures}
\end{figure}

\begin{figure} [t]
\centering
\begin{tabular}{p{0.01\linewidth}c}

    \begin{tabular}{c}\end{tabular}& 
    \begin{tabular}{c}
        \begin{subfigure}[t]{0.25\textwidth}
        \centering
        \includegraphics[width=\linewidth]{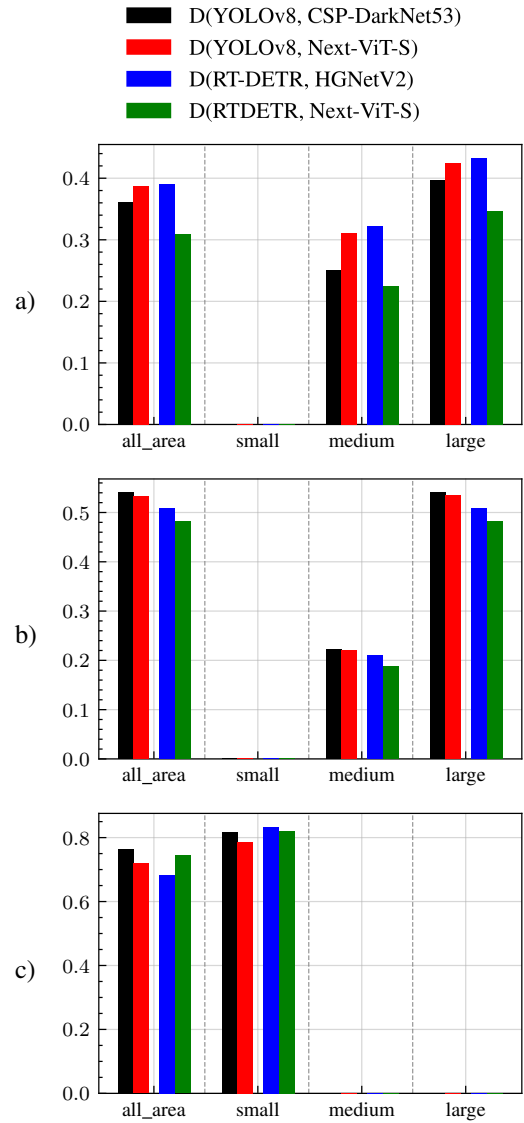}
        \end{subfigure}
    \end{tabular}\\

    \begin{tabular}{c}a)\end{tabular}& 
    \begin{tabular}{c}
        \begin{subfigure}[t]{0.35\textwidth}
        \centering
        \resizebox{\linewidth}{!}{\input{./EDS-object-size.pgf}}
        \end{subfigure}
    \end{tabular}\\

    \begin{tabular}{c}b)\end{tabular}& 
    \begin{tabular}{c}
        \begin{subfigure}[t]{0.35\textwidth}
        \centering
        \resizebox{\linewidth}{!}{\input{./HIXray-object-size.pgf}}
        \end{subfigure}
    \end{tabular}\\

    \begin{tabular}{c}c)\end{tabular}& 
    \begin{tabular}{c}
        \begin{subfigure}[t]{0.35\textwidth}
        \centering
        \resizebox{\linewidth}{!}{\input{./PIDray-object-size.pgf}}
        \end{subfigure}
    \end{tabular}\\
 
\end{tabular}
\caption{Object scale-related performance (mAP$^{50:95}$ metric) for datasets: a) EDS, b) HiXray, and c) PIDray.}
\label{fig:object-size_figures}
\end{figure}

\begin{figure*}[t]
    \centering
    \includegraphics[width=0.90\linewidth]{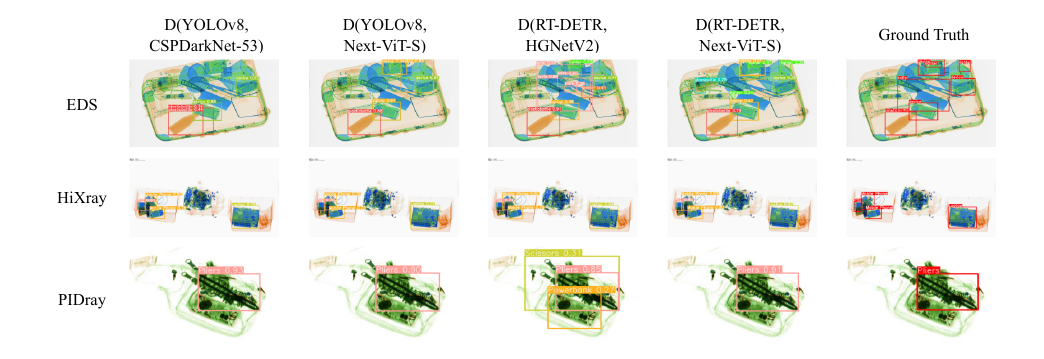}
    \caption{Indicative object detection results in the EDS, HiXray, and PIDray datasets.}
    \label{fig:visual_comparison}
\end{figure*}

Table \ref{tab:results} summarizes the object detection results obtained for all considered detectors for all the datasets employed, where both the mAP$^{50}$ and mAP$^{50:95}$ metrics are provided. Additionally, Tables \ref{tab:results-eds} and \ref{tab:results-pidray} present the detailed detection performance for the various experimental sessions and subsets of the EDS and the PIDray datasets, respectively. Object-level performance (mAP$^{50:95}$ metric) for each dataset is illustrated in Fig. \ref{fig:object-level_figures}. Moreover, Fig. \ref{fig:object-size_figures} demonstrates object scale-related performance (mAP$^{50:95}$ metric) for each dataset, where the COCO \cite{lin2014microsoft} dataset object scale definitions for `small', `medium', and `large' were considered. Furthermore, indicative detection results of the various detectors are illustrated in Fig. \ref{fig:visual_comparison}.

From the presented results, several critical observations and key insights can be extracted, the most important of which are summarized as follows:
\begin{itemize}
    \item Overall, the D(YOLOv8, CSP-DarkNet53) detector, i.e. a CNN-only architecture, achieves the best performance (Table \ref{tab:results}), exhibiting the highest recognition rate in $2$ out of the $3$ considered datasets (namely, HiXray and PIDray). This demonstrates the increased capability of convolutional operators in modeling appearance patterns in X-ray images. It needs to be highlighted though that both HiXray and PIDray contain data collected from a single scanner each.
    \item Interestingly, for the EDS dataset, where multiple/different scanners are employed and the defined experimental sessions target the evaluation under domain shifts in the underlying data distributions, most hybrid CNN-transformer detectors outperform the D(YOLOv8, CSP-DarkNet53) one, with D(RT-DETR, HGNetV2) showcasing the highest performance (Table \ref{tab:results}). The latter suggests that transformer-based components result into increased robustness to data distribution shifts, mainly due to their increased capability in incorporating global contextual information in the extracted features (compared to the respective convolutional-only blocks).
    \item Across all experiments (Tables \ref{tab:results}-\ref{tab:results-pidray}), YOLOv8-based detectors with a Next-ViT-S backbone consistently outperform their RT-DETR-based counterparts, demonstrating the superiority of the YOLOv8 detection head over the RT-DETR one. The latter indicates that transformer-based detection heads are not advantageous for the analysis of X-ray security images, which inherently exhibit no particular spatial structure (i.e. objects in containers are usually positioned without a specific/consistent spatial order).
    \item Replacing a CNN backbone (either CSP-DarkNet53 and HGNetV2) with a hybrid one (Next-ViT-S) is shown not to always lead to improved performance (Table \ref{tab:results}), regardless of the employed detection head (either YOLOv8- or RT-DETR-based one). This demonstrates the need for careful design and comprehensive experimentation regarding compatibility aspects, when incorporating a hybrid backbone in an object detection scheme.
    \item Examining the detection results in the EDS dataset (Table \ref{tab:results-eds}), it can be seen that the CNN-only D(YOLOv8, CSP-DarkNet53) detector generally leads to inferior performance in all experimental sessions, compared to most hybrid architectures (with the best D(YOLOv8, Next-ViT-S) and D(RT-DETR, HGNetV2) detectors performing almost equally well), as already discussed and explained above.
    \item Investigating the performance for the various subsets in the PID dataset (Table \ref{tab:results-pidray}), it can be seen that D(YOLOv8, CSP-DarkNet53) performs best for the `easy' and `hard' ones. However, for the `hidden' partition, where the objects of interest are intentionally concealed and significant occlusions are present, the hybrid D(YOLOv8, Next-ViT-S) detector demonstrates a slightly improved recognition rate.
    \item Analyzing the performance with respect to individual object types (Fig. \ref{fig:object-level_figures}), it can be observed that the D(YOLOv8, CSP-DarkNet53) detector outperforms all remaining ones for all classes in the HiXray and PIDray datasets. However, in the EDS dataset, D(YOLOv8, CSP-DarkNet53) performs inferior for all object types than most hybrid detectors, especially for classes `Plastic bottle' (DB), `Scissor' (SC), `Lighter' (LI), `Power bank' (PB), and `Pressure' (PR), i.e. both objects types with fine-grained local patterns (SC, LI, PB, and PR) as well as with broader motifs in their captured chemical composition appearance.
    \item Examining the impact of the objects' scale (Fig. \ref{fig:object-size_figures}), it can be seen that in the EDS dataset that D(YOLOv8, CSP-DarkNet53) is significantly outperformed by D(YOLOv8, Next-ViT-S) and D(RT-DETR, HGNetV2) for `medium' size objects, while this difference is decreased for `large' instances. On the other hand, for the HiXray and PIDray datasets (where D(YOLOv8, CSP-DarkNet53) performs best), the object scale does not seem to significantly affect the performance for all detectors.
\end{itemize}

\section{Conclusion}
\label{sec:conclusions}

In this paper, various hybrid CNN-transformer architectures were introduced and evaluated against a common CNN object detection baseline, namely YOLOv8. In particular, a CNN (HGNetV2) and a hybrid CNN-transformer (Next-ViT-S) backbone were combined with different CNN/transformer detection heads (YOLOv8 and RT-DETR). The resulting architectures were comparatively evaluated on three challenging public X-ray inspection datasets, namely EDS, HiXray, and PIDray. One of the key observations concerned the fact that while the YOLOv8 detector with its default backbone (CSP-DarkNet53) was shown to be advantageous on the HiXray and PIDray datasets, when a domain distribution shift is incorporated in the X-ray images (EDS datasets), hybrid CNN-transformer architectures demonstrated increased robustness. Future research includes the investigation of additional hybrid CNN-transformer configurations and broader experimental evaluation in additional datasets.

\balance
\bibliographystyle{ieeetr}
\bibliography{Papadopoulos-BDS.bib}

\end{document}

%% file: EDS-object-size.pgf
\begingroup%
\makeatletter%
\begin{pgfpicture}%
\pgfpathrectangle{\pgfpointorigin}{\pgfqpoint{2.624462in}{1.787377in}}%
\pgfusepath{use as bounding box, clip}%
\begin{pgfscope}%
\pgfsetbuttcap%
\pgfsetmiterjoin%
\definecolor{currentfill}{rgb}{1.000000,1.000000,1.000000}%
\pgfsetfillcolor{currentfill}%
\pgfsetlinewidth{0.000000pt}%
\definecolor{currentstroke}{rgb}{1.000000,1.000000,1.000000}%
\pgfsetstrokecolor{currentstroke}%
\pgfsetdash{}{0pt}%
\pgfpathmoveto{\pgfqpoint{0.000000in}{0.000000in}}%
\pgfpathlineto{\pgfqpoint{2.624462in}{0.000000in}}%
\pgfpathlineto{\pgfqpoint{2.624462in}{1.787377in}}%
\pgfpathlineto{\pgfqpoint{0.000000in}{1.787377in}}%
\pgfpathlineto{\pgfqpoint{0.000000in}{0.000000in}}%
\pgfpathclose%
\pgfusepath{fill}%
\end{pgfscope}%
\begin{pgfscope}%
\pgfsetbuttcap%
\pgfsetmiterjoin%
\definecolor{currentfill}{rgb}{1.000000,1.000000,1.000000}%
\pgfsetfillcolor{currentfill}%
\pgfsetlinewidth{0.000000pt}%
\definecolor{currentstroke}{rgb}{0.000000,0.000000,0.000000}%
\pgfsetstrokecolor{currentstroke}%
\pgfsetstrokeopacity{0.000000}%
\pgfsetdash{}{0pt}%
\pgfpathmoveto{\pgfqpoint{0.249462in}{0.197377in}}%
\pgfpathlineto{\pgfqpoint{2.574462in}{0.197377in}}%
\pgfpathlineto{\pgfqpoint{2.574462in}{1.737377in}}%
\pgfpathlineto{\pgfqpoint{0.249462in}{1.737377in}}%
\pgfpathlineto{\pgfqpoint{0.249462in}{0.197377in}}%
\pgfpathclose%
\pgfusepath{fill}%
\end{pgfscope}%
\begin{pgfscope}%
\pgfpathrectangle{\pgfqpoint{0.249462in}{0.197377in}}{\pgfqpoint{2.325000in}{1.540000in}}%
\pgfusepath{clip}%
\pgfsetrectcap%
\pgfsetroundjoin%
\pgfsetlinewidth{0.501875pt}%
\definecolor{currentstroke}{rgb}{0.690196,0.690196,0.690196}%
\pgfsetstrokecolor{currentstroke}%
\pgfsetstrokeopacity{0.500000}%
\pgfsetdash{}{0pt}%
\pgfpathmoveto{\pgfqpoint{0.552226in}{0.197377in}}%
\pgfpathlineto{\pgfqpoint{0.552226in}{1.737377in}}%
\pgfusepath{stroke}%
\end{pgfscope}%
\begin{pgfscope}%
\pgfsetbuttcap%
\pgfsetroundjoin%
\definecolor{currentfill}{rgb}{0.000000,0.000000,0.000000}%
\pgfsetfillcolor{currentfill}%
\pgfsetlinewidth{0.501875pt}%
\definecolor{currentstroke}{rgb}{0.000000,0.000000,0.000000}%
\pgfsetstrokecolor{currentstroke}%
\pgfsetdash{}{0pt}%
\pgfsys@defobject{currentmarker}{\pgfqpoint{0.000000in}{0.000000in}}{\pgfqpoint{0.000000in}{0.041667in}}{%
\pgfpathmoveto{\pgfqpoint{0.000000in}{0.000000in}}%
\pgfpathlineto{\pgfqpoint{0.000000in}{0.041667in}}%
\pgfusepath{stroke,fill}%
}%
\begin{pgfscope}%
\pgfsys@transformshift{0.552226in}{0.197377in}%
\pgfsys@useobject{currentmarker}{}%
\end{pgfscope}%
\end{pgfscope}%
\begin{pgfscope}%
\definecolor{textcolor}{rgb}{0.000000,0.000000,0.000000}%
\pgfsetstrokecolor{textcolor}%
\pgfsetfillcolor{textcolor}%
\pgftext[x=0.552226in,y=0.148766in,,top]{\color{textcolor}{\rmfamily\fontsize{8.000000}{9.600000}\selectfont\catcode`\^=\active\def^{\ifmmode\sp\else\^{}\fi}\catcode`\%=\active\def
\end{pgfscope}%
\begin{pgfscope}%
\pgfpathrectangle{\pgfqpoint{0.249462in}{0.197377in}}{\pgfqpoint{2.325000in}{1.540000in}}%
\pgfusepath{clip}%
\pgfsetrectcap%
\pgfsetroundjoin%
\pgfsetlinewidth{0.501875pt}%
\definecolor{currentstroke}{rgb}{0.690196,0.690196,0.690196}%
\pgfsetstrokecolor{currentstroke}%
\pgfsetstrokeopacity{0.500000}%
\pgfsetdash{}{0pt}%
\pgfpathmoveto{\pgfqpoint{1.123479in}{0.197377in}}%
\pgfpathlineto{\pgfqpoint{1.123479in}{1.737377in}}%
\pgfusepath{stroke}%
\end{pgfscope}%
\begin{pgfscope}%
\pgfsetbuttcap%
\pgfsetroundjoin%
\definecolor{currentfill}{rgb}{0.000000,0.000000,0.000000}%
\pgfsetfillcolor{currentfill}%
\pgfsetlinewidth{0.501875pt}%
\definecolor{currentstroke}{rgb}{0.000000,0.000000,0.000000}%
\pgfsetstrokecolor{currentstroke}%
\pgfsetdash{}{0pt}%
\pgfsys@defobject{currentmarker}{\pgfqpoint{0.000000in}{0.000000in}}{\pgfqpoint{0.000000in}{0.041667in}}{%
\pgfpathmoveto{\pgfqpoint{0.000000in}{0.000000in}}%
\pgfpathlineto{\pgfqpoint{0.000000in}{0.041667in}}%
\pgfusepath{stroke,fill}%
}%
\begin{pgfscope}%
\pgfsys@transformshift{1.123479in}{0.197377in}%
\pgfsys@useobject{currentmarker}{}%
\end{pgfscope}%
\end{pgfscope}%
\begin{pgfscope}%
\definecolor{textcolor}{rgb}{0.000000,0.000000,0.000000}%
\pgfsetstrokecolor{textcolor}%
\pgfsetfillcolor{textcolor}%
\pgftext[x=1.123479in,y=0.148766in,,top]{\color{textcolor}{\rmfamily\fontsize{8.000000}{9.600000}\selectfont\catcode`\^=\active\def^{\ifmmode\sp\else\^{}\fi}\catcode`\%=\active\def
\end{pgfscope}%
\begin{pgfscope}%
\pgfpathrectangle{\pgfqpoint{0.249462in}{0.197377in}}{\pgfqpoint{2.325000in}{1.540000in}}%
\pgfusepath{clip}%
\pgfsetrectcap%
\pgfsetroundjoin%
\pgfsetlinewidth{0.501875pt}%
\definecolor{currentstroke}{rgb}{0.690196,0.690196,0.690196}%
\pgfsetstrokecolor{currentstroke}%
\pgfsetstrokeopacity{0.500000}%
\pgfsetdash{}{0pt}%
\pgfpathmoveto{\pgfqpoint{1.694732in}{0.197377in}}%
\pgfpathlineto{\pgfqpoint{1.694732in}{1.737377in}}%
\pgfusepath{stroke}%
\end{pgfscope}%
\begin{pgfscope}%
\pgfsetbuttcap%
\pgfsetroundjoin%
\definecolor{currentfill}{rgb}{0.000000,0.000000,0.000000}%
\pgfsetfillcolor{currentfill}%
\pgfsetlinewidth{0.501875pt}%
\definecolor{currentstroke}{rgb}{0.000000,0.000000,0.000000}%
\pgfsetstrokecolor{currentstroke}%
\pgfsetdash{}{0pt}%
\pgfsys@defobject{currentmarker}{\pgfqpoint{0.000000in}{0.000000in}}{\pgfqpoint{0.000000in}{0.041667in}}{%
\pgfpathmoveto{\pgfqpoint{0.000000in}{0.000000in}}%
\pgfpathlineto{\pgfqpoint{0.000000in}{0.041667in}}%
\pgfusepath{stroke,fill}%
}%
\begin{pgfscope}%
\pgfsys@transformshift{1.694732in}{0.197377in}%
\pgfsys@useobject{currentmarker}{}%
\end{pgfscope}%
\end{pgfscope}%
\begin{pgfscope}%
\definecolor{textcolor}{rgb}{0.000000,0.000000,0.000000}%
\pgfsetstrokecolor{textcolor}%
\pgfsetfillcolor{textcolor}%
\pgftext[x=1.694732in,y=0.148766in,,top]{\color{textcolor}{\rmfamily\fontsize{8.000000}{9.600000}\selectfont\catcode`\^=\active\def^{\ifmmode\sp\else\^{}\fi}\catcode`\%=\active\def
\end{pgfscope}%
\begin{pgfscope}%
\pgfpathrectangle{\pgfqpoint{0.249462in}{0.197377in}}{\pgfqpoint{2.325000in}{1.540000in}}%
\pgfusepath{clip}%
\pgfsetrectcap%
\pgfsetroundjoin%
\pgfsetlinewidth{0.501875pt}%
\definecolor{currentstroke}{rgb}{0.690196,0.690196,0.690196}%
\pgfsetstrokecolor{currentstroke}%
\pgfsetstrokeopacity{0.500000}%
\pgfsetdash{}{0pt}%
\pgfpathmoveto{\pgfqpoint{2.265985in}{0.197377in}}%
\pgfpathlineto{\pgfqpoint{2.265985in}{1.737377in}}%
\pgfusepath{stroke}%
\end{pgfscope}%
\begin{pgfscope}%
\pgfsetbuttcap%
\pgfsetroundjoin%
\definecolor{currentfill}{rgb}{0.000000,0.000000,0.000000}%
\pgfsetfillcolor{currentfill}%
\pgfsetlinewidth{0.501875pt}%
\definecolor{currentstroke}{rgb}{0.000000,0.000000,0.000000}%
\pgfsetstrokecolor{currentstroke}%
\pgfsetdash{}{0pt}%
\pgfsys@defobject{currentmarker}{\pgfqpoint{0.000000in}{0.000000in}}{\pgfqpoint{0.000000in}{0.041667in}}{%
\pgfpathmoveto{\pgfqpoint{0.000000in}{0.000000in}}%
\pgfpathlineto{\pgfqpoint{0.000000in}{0.041667in}}%
\pgfusepath{stroke,fill}%
}%
\begin{pgfscope}%
\pgfsys@transformshift{2.265985in}{0.197377in}%
\pgfsys@useobject{currentmarker}{}%
\end{pgfscope}%
\end{pgfscope}%
\begin{pgfscope}%
\definecolor{textcolor}{rgb}{0.000000,0.000000,0.000000}%
\pgfsetstrokecolor{textcolor}%
\pgfsetfillcolor{textcolor}%
\pgftext[x=2.265985in,y=0.148766in,,top]{\color{textcolor}{\rmfamily\fontsize{8.000000}{9.600000}\selectfont\catcode`\^=\active\def^{\ifmmode\sp\else\^{}\fi}\catcode`\%=\active\def
\end{pgfscope}%
\begin{pgfscope}%
\pgfpathrectangle{\pgfqpoint{0.249462in}{0.197377in}}{\pgfqpoint{2.325000in}{1.540000in}}%
\pgfusepath{clip}%
\pgfsetrectcap%
\pgfsetroundjoin%
\pgfsetlinewidth{0.501875pt}%
\definecolor{currentstroke}{rgb}{0.690196,0.690196,0.690196}%
\pgfsetstrokecolor{currentstroke}%
\pgfsetstrokeopacity{0.500000}%
\pgfsetdash{}{0pt}%
\pgfpathmoveto{\pgfqpoint{0.249462in}{0.197377in}}%
\pgfpathlineto{\pgfqpoint{2.574462in}{0.197377in}}%
\pgfusepath{stroke}%
\end{pgfscope}%
\begin{pgfscope}%
\pgfsetbuttcap%
\pgfsetroundjoin%
\definecolor{currentfill}{rgb}{0.000000,0.000000,0.000000}%
\pgfsetfillcolor{currentfill}%
\pgfsetlinewidth{0.501875pt}%
\definecolor{currentstroke}{rgb}{0.000000,0.000000,0.000000}%
\pgfsetstrokecolor{currentstroke}%
\pgfsetdash{}{0pt}%
\pgfsys@defobject{currentmarker}{\pgfqpoint{0.000000in}{0.000000in}}{\pgfqpoint{0.041667in}{0.000000in}}{%
\pgfpathmoveto{\pgfqpoint{0.000000in}{0.000000in}}%
\pgfpathlineto{\pgfqpoint{0.041667in}{0.000000in}}%
\pgfusepath{stroke,fill}%
}%
\begin{pgfscope}%
\pgfsys@transformshift{0.249462in}{0.197377in}%
\pgfsys@useobject{currentmarker}{}%
\end{pgfscope}%
\end{pgfscope}%
\begin{pgfscope}%
\definecolor{textcolor}{rgb}{0.000000,0.000000,0.000000}%
\pgfsetstrokecolor{textcolor}%
\pgfsetfillcolor{textcolor}%
\pgftext[x=0.050000in, y=0.158796in, left, base]{\color{textcolor}{\rmfamily\fontsize{8.000000}{9.600000}\selectfont\catcode`\^=\active\def^{\ifmmode\sp\else\^{}\fi}\catcode`\%=\active\def
\end{pgfscope}%
\begin{pgfscope}%
\pgfpathrectangle{\pgfqpoint{0.249462in}{0.197377in}}{\pgfqpoint{2.325000in}{1.540000in}}%
\pgfusepath{clip}%
\pgfsetrectcap%
\pgfsetroundjoin%
\pgfsetlinewidth{0.501875pt}%
\definecolor{currentstroke}{rgb}{0.690196,0.690196,0.690196}%
\pgfsetstrokecolor{currentstroke}%
\pgfsetstrokeopacity{0.500000}%
\pgfsetdash{}{0pt}%
\pgfpathmoveto{\pgfqpoint{0.249462in}{0.536016in}}%
\pgfpathlineto{\pgfqpoint{2.574462in}{0.536016in}}%
\pgfusepath{stroke}%
\end{pgfscope}%
\begin{pgfscope}%
\pgfsetbuttcap%
\pgfsetroundjoin%
\definecolor{currentfill}{rgb}{0.000000,0.000000,0.000000}%
\pgfsetfillcolor{currentfill}%
\pgfsetlinewidth{0.501875pt}%
\definecolor{currentstroke}{rgb}{0.000000,0.000000,0.000000}%
\pgfsetstrokecolor{currentstroke}%
\pgfsetdash{}{0pt}%
\pgfsys@defobject{currentmarker}{\pgfqpoint{0.000000in}{0.000000in}}{\pgfqpoint{0.041667in}{0.000000in}}{%
\pgfpathmoveto{\pgfqpoint{0.000000in}{0.000000in}}%
\pgfpathlineto{\pgfqpoint{0.041667in}{0.000000in}}%
\pgfusepath{stroke,fill}%
}%
\begin{pgfscope}%
\pgfsys@transformshift{0.249462in}{0.536016in}%
\pgfsys@useobject{currentmarker}{}%
\end{pgfscope}%
\end{pgfscope}%
\begin{pgfscope}%
\definecolor{textcolor}{rgb}{0.000000,0.000000,0.000000}%
\pgfsetstrokecolor{textcolor}%
\pgfsetfillcolor{textcolor}%
\pgftext[x=0.050000in, y=0.497436in, left, base]{\color{textcolor}{\rmfamily\fontsize{8.000000}{9.600000}\selectfont\catcode`\^=\active\def^{\ifmmode\sp\else\^{}\fi}\catcode`\%=\active\def
\end{pgfscope}%
\begin{pgfscope}%
\pgfpathrectangle{\pgfqpoint{0.249462in}{0.197377in}}{\pgfqpoint{2.325000in}{1.540000in}}%
\pgfusepath{clip}%
\pgfsetrectcap%
\pgfsetroundjoin%
\pgfsetlinewidth{0.501875pt}%
\definecolor{currentstroke}{rgb}{0.690196,0.690196,0.690196}%
\pgfsetstrokecolor{currentstroke}%
\pgfsetstrokeopacity{0.500000}%
\pgfsetdash{}{0pt}%
\pgfpathmoveto{\pgfqpoint{0.249462in}{0.874656in}}%
\pgfpathlineto{\pgfqpoint{2.574462in}{0.874656in}}%
\pgfusepath{stroke}%
\end{pgfscope}%
\begin{pgfscope}%
\pgfsetbuttcap%
\pgfsetroundjoin%
\definecolor{currentfill}{rgb}{0.000000,0.000000,0.000000}%
\pgfsetfillcolor{currentfill}%
\pgfsetlinewidth{0.501875pt}%
\definecolor{currentstroke}{rgb}{0.000000,0.000000,0.000000}%
\pgfsetstrokecolor{currentstroke}%
\pgfsetdash{}{0pt}%
\pgfsys@defobject{currentmarker}{\pgfqpoint{0.000000in}{0.000000in}}{\pgfqpoint{0.041667in}{0.000000in}}{%
\pgfpathmoveto{\pgfqpoint{0.000000in}{0.000000in}}%
\pgfpathlineto{\pgfqpoint{0.041667in}{0.000000in}}%
\pgfusepath{stroke,fill}%
}%
\begin{pgfscope}%
\pgfsys@transformshift{0.249462in}{0.874656in}%
\pgfsys@useobject{currentmarker}{}%
\end{pgfscope}%
\end{pgfscope}%
\begin{pgfscope}%
\definecolor{textcolor}{rgb}{0.000000,0.000000,0.000000}%
\pgfsetstrokecolor{textcolor}%
\pgfsetfillcolor{textcolor}%
\pgftext[x=0.050000in, y=0.836076in, left, base]{\color{textcolor}{\rmfamily\fontsize{8.000000}{9.600000}\selectfont\catcode`\^=\active\def^{\ifmmode\sp\else\^{}\fi}\catcode`\%=\active\def
\end{pgfscope}%
\begin{pgfscope}%
\pgfpathrectangle{\pgfqpoint{0.249462in}{0.197377in}}{\pgfqpoint{2.325000in}{1.540000in}}%
\pgfusepath{clip}%
\pgfsetrectcap%
\pgfsetroundjoin%
\pgfsetlinewidth{0.501875pt}%
\definecolor{currentstroke}{rgb}{0.690196,0.690196,0.690196}%
\pgfsetstrokecolor{currentstroke}%
\pgfsetstrokeopacity{0.500000}%
\pgfsetdash{}{0pt}%
\pgfpathmoveto{\pgfqpoint{0.249462in}{1.213295in}}%
\pgfpathlineto{\pgfqpoint{2.574462in}{1.213295in}}%
\pgfusepath{stroke}%
\end{pgfscope}%
\begin{pgfscope}%
\pgfsetbuttcap%
\pgfsetroundjoin%
\definecolor{currentfill}{rgb}{0.000000,0.000000,0.000000}%
\pgfsetfillcolor{currentfill}%
\pgfsetlinewidth{0.501875pt}%
\definecolor{currentstroke}{rgb}{0.000000,0.000000,0.000000}%
\pgfsetstrokecolor{currentstroke}%
\pgfsetdash{}{0pt}%
\pgfsys@defobject{currentmarker}{\pgfqpoint{0.000000in}{0.000000in}}{\pgfqpoint{0.041667in}{0.000000in}}{%
\pgfpathmoveto{\pgfqpoint{0.000000in}{0.000000in}}%
\pgfpathlineto{\pgfqpoint{0.041667in}{0.000000in}}%
\pgfusepath{stroke,fill}%
}%
\begin{pgfscope}%
\pgfsys@transformshift{0.249462in}{1.213295in}%
\pgfsys@useobject{currentmarker}{}%
\end{pgfscope}%
\end{pgfscope}%
\begin{pgfscope}%
\definecolor{textcolor}{rgb}{0.000000,0.000000,0.000000}%
\pgfsetstrokecolor{textcolor}%
\pgfsetfillcolor{textcolor}%
\pgftext[x=0.050000in, y=1.174715in, left, base]{\color{textcolor}{\rmfamily\fontsize{8.000000}{9.600000}\selectfont\catcode`\^=\active\def^{\ifmmode\sp\else\^{}\fi}\catcode`\%=\active\def
\end{pgfscope}%
\begin{pgfscope}%
\pgfpathrectangle{\pgfqpoint{0.249462in}{0.197377in}}{\pgfqpoint{2.325000in}{1.540000in}}%
\pgfusepath{clip}%
\pgfsetrectcap%
\pgfsetroundjoin%
\pgfsetlinewidth{0.501875pt}%
\definecolor{currentstroke}{rgb}{0.690196,0.690196,0.690196}%
\pgfsetstrokecolor{currentstroke}%
\pgfsetstrokeopacity{0.500000}%
\pgfsetdash{}{0pt}%
\pgfpathmoveto{\pgfqpoint{0.249462in}{1.551935in}}%
\pgfpathlineto{\pgfqpoint{2.574462in}{1.551935in}}%
\pgfusepath{stroke}%
\end{pgfscope}%
\begin{pgfscope}%
\pgfsetbuttcap%
\pgfsetroundjoin%
\definecolor{currentfill}{rgb}{0.000000,0.000000,0.000000}%
\pgfsetfillcolor{currentfill}%
\pgfsetlinewidth{0.501875pt}%
\definecolor{currentstroke}{rgb}{0.000000,0.000000,0.000000}%
\pgfsetstrokecolor{currentstroke}%
\pgfsetdash{}{0pt}%
\pgfsys@defobject{currentmarker}{\pgfqpoint{0.000000in}{0.000000in}}{\pgfqpoint{0.041667in}{0.000000in}}{%
\pgfpathmoveto{\pgfqpoint{0.000000in}{0.000000in}}%
\pgfpathlineto{\pgfqpoint{0.041667in}{0.000000in}}%
\pgfusepath{stroke,fill}%
}%
\begin{pgfscope}%
\pgfsys@transformshift{0.249462in}{1.551935in}%
\pgfsys@useobject{currentmarker}{}%
\end{pgfscope}%
\end{pgfscope}%
\begin{pgfscope}%
\definecolor{textcolor}{rgb}{0.000000,0.000000,0.000000}%
\pgfsetstrokecolor{textcolor}%
\pgfsetfillcolor{textcolor}%
\pgftext[x=0.050000in, y=1.513355in, left, base]{\color{textcolor}{\rmfamily\fontsize{8.000000}{9.600000}\selectfont\catcode`\^=\active\def^{\ifmmode\sp\else\^{}\fi}\catcode`\%=\active\def
\end{pgfscope}%
\begin{pgfscope}%
\pgfsetbuttcap%
\pgfsetroundjoin%
\definecolor{currentfill}{rgb}{0.000000,0.000000,0.000000}%
\pgfsetfillcolor{currentfill}%
\pgfsetlinewidth{0.501875pt}%
\definecolor{currentstroke}{rgb}{0.000000,0.000000,0.000000}%
\pgfsetstrokecolor{currentstroke}%
\pgfsetdash{}{0pt}%
\pgfsys@defobject{currentmarker}{\pgfqpoint{0.000000in}{0.000000in}}{\pgfqpoint{0.020833in}{0.000000in}}{%
\pgfpathmoveto{\pgfqpoint{0.000000in}{0.000000in}}%
\pgfpathlineto{\pgfqpoint{0.020833in}{0.000000in}}%
\pgfusepath{stroke,fill}%
}%
\begin{pgfscope}%
\pgfsys@transformshift{0.249462in}{0.265105in}%
\pgfsys@useobject{currentmarker}{}%
\end{pgfscope}%
\end{pgfscope}%
\begin{pgfscope}%
\pgfsetbuttcap%
\pgfsetroundjoin%
\definecolor{currentfill}{rgb}{0.000000,0.000000,0.000000}%
\pgfsetfillcolor{currentfill}%
\pgfsetlinewidth{0.501875pt}%
\definecolor{currentstroke}{rgb}{0.000000,0.000000,0.000000}%
\pgfsetstrokecolor{currentstroke}%
\pgfsetdash{}{0pt}%
\pgfsys@defobject{currentmarker}{\pgfqpoint{0.000000in}{0.000000in}}{\pgfqpoint{0.020833in}{0.000000in}}{%
\pgfpathmoveto{\pgfqpoint{0.000000in}{0.000000in}}%
\pgfpathlineto{\pgfqpoint{0.020833in}{0.000000in}}%
\pgfusepath{stroke,fill}%
}%
\begin{pgfscope}%
\pgfsys@transformshift{0.249462in}{0.332833in}%
\pgfsys@useobject{currentmarker}{}%
\end{pgfscope}%
\end{pgfscope}%
\begin{pgfscope}%
\pgfsetbuttcap%
\pgfsetroundjoin%
\definecolor{currentfill}{rgb}{0.000000,0.000000,0.000000}%
\pgfsetfillcolor{currentfill}%
\pgfsetlinewidth{0.501875pt}%
\definecolor{currentstroke}{rgb}{0.000000,0.000000,0.000000}%
\pgfsetstrokecolor{currentstroke}%
\pgfsetdash{}{0pt}%
\pgfsys@defobject{currentmarker}{\pgfqpoint{0.000000in}{0.000000in}}{\pgfqpoint{0.020833in}{0.000000in}}{%
\pgfpathmoveto{\pgfqpoint{0.000000in}{0.000000in}}%
\pgfpathlineto{\pgfqpoint{0.020833in}{0.000000in}}%
\pgfusepath{stroke,fill}%
}%
\begin{pgfscope}%
\pgfsys@transformshift{0.249462in}{0.400560in}%
\pgfsys@useobject{currentmarker}{}%
\end{pgfscope}%
\end{pgfscope}%
\begin{pgfscope}%
\pgfsetbuttcap%
\pgfsetroundjoin%
\definecolor{currentfill}{rgb}{0.000000,0.000000,0.000000}%
\pgfsetfillcolor{currentfill}%
\pgfsetlinewidth{0.501875pt}%
\definecolor{currentstroke}{rgb}{0.000000,0.000000,0.000000}%
\pgfsetstrokecolor{currentstroke}%
\pgfsetdash{}{0pt}%
\pgfsys@defobject{currentmarker}{\pgfqpoint{0.000000in}{0.000000in}}{\pgfqpoint{0.020833in}{0.000000in}}{%
\pgfpathmoveto{\pgfqpoint{0.000000in}{0.000000in}}%
\pgfpathlineto{\pgfqpoint{0.020833in}{0.000000in}}%
\pgfusepath{stroke,fill}%
}%
\begin{pgfscope}%
\pgfsys@transformshift{0.249462in}{0.468288in}%
\pgfsys@useobject{currentmarker}{}%
\end{pgfscope}%
\end{pgfscope}%
\begin{pgfscope}%
\pgfsetbuttcap%
\pgfsetroundjoin%
\definecolor{currentfill}{rgb}{0.000000,0.000000,0.000000}%
\pgfsetfillcolor{currentfill}%
\pgfsetlinewidth{0.501875pt}%
\definecolor{currentstroke}{rgb}{0.000000,0.000000,0.000000}%
\pgfsetstrokecolor{currentstroke}%
\pgfsetdash{}{0pt}%
\pgfsys@defobject{currentmarker}{\pgfqpoint{0.000000in}{0.000000in}}{\pgfqpoint{0.020833in}{0.000000in}}{%
\pgfpathmoveto{\pgfqpoint{0.000000in}{0.000000in}}%
\pgfpathlineto{\pgfqpoint{0.020833in}{0.000000in}}%
\pgfusepath{stroke,fill}%
}%
\begin{pgfscope}%
\pgfsys@transformshift{0.249462in}{0.603744in}%
\pgfsys@useobject{currentmarker}{}%
\end{pgfscope}%
\end{pgfscope}%
\begin{pgfscope}%
\pgfsetbuttcap%
\pgfsetroundjoin%
\definecolor{currentfill}{rgb}{0.000000,0.000000,0.000000}%
\pgfsetfillcolor{currentfill}%
\pgfsetlinewidth{0.501875pt}%
\definecolor{currentstroke}{rgb}{0.000000,0.000000,0.000000}%
\pgfsetstrokecolor{currentstroke}%
\pgfsetdash{}{0pt}%
\pgfsys@defobject{currentmarker}{\pgfqpoint{0.000000in}{0.000000in}}{\pgfqpoint{0.020833in}{0.000000in}}{%
\pgfpathmoveto{\pgfqpoint{0.000000in}{0.000000in}}%
\pgfpathlineto{\pgfqpoint{0.020833in}{0.000000in}}%
\pgfusepath{stroke,fill}%
}%
\begin{pgfscope}%
\pgfsys@transformshift{0.249462in}{0.671472in}%
\pgfsys@useobject{currentmarker}{}%
\end{pgfscope}%
\end{pgfscope}%
\begin{pgfscope}%
\pgfsetbuttcap%
\pgfsetroundjoin%
\definecolor{currentfill}{rgb}{0.000000,0.000000,0.000000}%
\pgfsetfillcolor{currentfill}%
\pgfsetlinewidth{0.501875pt}%
\definecolor{currentstroke}{rgb}{0.000000,0.000000,0.000000}%
\pgfsetstrokecolor{currentstroke}%
\pgfsetdash{}{0pt}%
\pgfsys@defobject{currentmarker}{\pgfqpoint{0.000000in}{0.000000in}}{\pgfqpoint{0.020833in}{0.000000in}}{%
\pgfpathmoveto{\pgfqpoint{0.000000in}{0.000000in}}%
\pgfpathlineto{\pgfqpoint{0.020833in}{0.000000in}}%
\pgfusepath{stroke,fill}%
}%
\begin{pgfscope}%
\pgfsys@transformshift{0.249462in}{0.739200in}%
\pgfsys@useobject{currentmarker}{}%
\end{pgfscope}%
\end{pgfscope}%
\begin{pgfscope}%
\pgfsetbuttcap%
\pgfsetroundjoin%
\definecolor{currentfill}{rgb}{0.000000,0.000000,0.000000}%
\pgfsetfillcolor{currentfill}%
\pgfsetlinewidth{0.501875pt}%
\definecolor{currentstroke}{rgb}{0.000000,0.000000,0.000000}%
\pgfsetstrokecolor{currentstroke}%
\pgfsetdash{}{0pt}%
\pgfsys@defobject{currentmarker}{\pgfqpoint{0.000000in}{0.000000in}}{\pgfqpoint{0.020833in}{0.000000in}}{%
\pgfpathmoveto{\pgfqpoint{0.000000in}{0.000000in}}%
\pgfpathlineto{\pgfqpoint{0.020833in}{0.000000in}}%
\pgfusepath{stroke,fill}%
}%
\begin{pgfscope}%
\pgfsys@transformshift{0.249462in}{0.806928in}%
\pgfsys@useobject{currentmarker}{}%
\end{pgfscope}%
\end{pgfscope}%
\begin{pgfscope}%
\pgfsetbuttcap%
\pgfsetroundjoin%
\definecolor{currentfill}{rgb}{0.000000,0.000000,0.000000}%
\pgfsetfillcolor{currentfill}%
\pgfsetlinewidth{0.501875pt}%
\definecolor{currentstroke}{rgb}{0.000000,0.000000,0.000000}%
\pgfsetstrokecolor{currentstroke}%
\pgfsetdash{}{0pt}%
\pgfsys@defobject{currentmarker}{\pgfqpoint{0.000000in}{0.000000in}}{\pgfqpoint{0.020833in}{0.000000in}}{%
\pgfpathmoveto{\pgfqpoint{0.000000in}{0.000000in}}%
\pgfpathlineto{\pgfqpoint{0.020833in}{0.000000in}}%
\pgfusepath{stroke,fill}%
}%
\begin{pgfscope}%
\pgfsys@transformshift{0.249462in}{0.942384in}%
\pgfsys@useobject{currentmarker}{}%
\end{pgfscope}%
\end{pgfscope}%
\begin{pgfscope}%
\pgfsetbuttcap%
\pgfsetroundjoin%
\definecolor{currentfill}{rgb}{0.000000,0.000000,0.000000}%
\pgfsetfillcolor{currentfill}%
\pgfsetlinewidth{0.501875pt}%
\definecolor{currentstroke}{rgb}{0.000000,0.000000,0.000000}%
\pgfsetstrokecolor{currentstroke}%
\pgfsetdash{}{0pt}%
\pgfsys@defobject{currentmarker}{\pgfqpoint{0.000000in}{0.000000in}}{\pgfqpoint{0.020833in}{0.000000in}}{%
\pgfpathmoveto{\pgfqpoint{0.000000in}{0.000000in}}%
\pgfpathlineto{\pgfqpoint{0.020833in}{0.000000in}}%
\pgfusepath{stroke,fill}%
}%
\begin{pgfscope}%
\pgfsys@transformshift{0.249462in}{1.010112in}%
\pgfsys@useobject{currentmarker}{}%
\end{pgfscope}%
\end{pgfscope}%
\begin{pgfscope}%
\pgfsetbuttcap%
\pgfsetroundjoin%
\definecolor{currentfill}{rgb}{0.000000,0.000000,0.000000}%
\pgfsetfillcolor{currentfill}%
\pgfsetlinewidth{0.501875pt}%
\definecolor{currentstroke}{rgb}{0.000000,0.000000,0.000000}%
\pgfsetstrokecolor{currentstroke}%
\pgfsetdash{}{0pt}%
\pgfsys@defobject{currentmarker}{\pgfqpoint{0.000000in}{0.000000in}}{\pgfqpoint{0.020833in}{0.000000in}}{%
\pgfpathmoveto{\pgfqpoint{0.000000in}{0.000000in}}%
\pgfpathlineto{\pgfqpoint{0.020833in}{0.000000in}}%
\pgfusepath{stroke,fill}%
}%
\begin{pgfscope}%
\pgfsys@transformshift{0.249462in}{1.077840in}%
\pgfsys@useobject{currentmarker}{}%
\end{pgfscope}%
\end{pgfscope}%
\begin{pgfscope}%
\pgfsetbuttcap%
\pgfsetroundjoin%
\definecolor{currentfill}{rgb}{0.000000,0.000000,0.000000}%
\pgfsetfillcolor{currentfill}%
\pgfsetlinewidth{0.501875pt}%
\definecolor{currentstroke}{rgb}{0.000000,0.000000,0.000000}%
\pgfsetstrokecolor{currentstroke}%
\pgfsetdash{}{0pt}%
\pgfsys@defobject{currentmarker}{\pgfqpoint{0.000000in}{0.000000in}}{\pgfqpoint{0.020833in}{0.000000in}}{%
\pgfpathmoveto{\pgfqpoint{0.000000in}{0.000000in}}%
\pgfpathlineto{\pgfqpoint{0.020833in}{0.000000in}}%
\pgfusepath{stroke,fill}%
}%
\begin{pgfscope}%
\pgfsys@transformshift{0.249462in}{1.145568in}%
\pgfsys@useobject{currentmarker}{}%
\end{pgfscope}%
\end{pgfscope}%
\begin{pgfscope}%
\pgfsetbuttcap%
\pgfsetroundjoin%
\definecolor{currentfill}{rgb}{0.000000,0.000000,0.000000}%
\pgfsetfillcolor{currentfill}%
\pgfsetlinewidth{0.501875pt}%
\definecolor{currentstroke}{rgb}{0.000000,0.000000,0.000000}%
\pgfsetstrokecolor{currentstroke}%
\pgfsetdash{}{0pt}%
\pgfsys@defobject{currentmarker}{\pgfqpoint{0.000000in}{0.000000in}}{\pgfqpoint{0.020833in}{0.000000in}}{%
\pgfpathmoveto{\pgfqpoint{0.000000in}{0.000000in}}%
\pgfpathlineto{\pgfqpoint{0.020833in}{0.000000in}}%
\pgfusepath{stroke,fill}%
}%
\begin{pgfscope}%
\pgfsys@transformshift{0.249462in}{1.281023in}%
\pgfsys@useobject{currentmarker}{}%
\end{pgfscope}%
\end{pgfscope}%
\begin{pgfscope}%
\pgfsetbuttcap%
\pgfsetroundjoin%
\definecolor{currentfill}{rgb}{0.000000,0.000000,0.000000}%
\pgfsetfillcolor{currentfill}%
\pgfsetlinewidth{0.501875pt}%
\definecolor{currentstroke}{rgb}{0.000000,0.000000,0.000000}%
\pgfsetstrokecolor{currentstroke}%
\pgfsetdash{}{0pt}%
\pgfsys@defobject{currentmarker}{\pgfqpoint{0.000000in}{0.000000in}}{\pgfqpoint{0.020833in}{0.000000in}}{%
\pgfpathmoveto{\pgfqpoint{0.000000in}{0.000000in}}%
\pgfpathlineto{\pgfqpoint{0.020833in}{0.000000in}}%
\pgfusepath{stroke,fill}%
}%
\begin{pgfscope}%
\pgfsys@transformshift{0.249462in}{1.348751in}%
\pgfsys@useobject{currentmarker}{}%
\end{pgfscope}%
\end{pgfscope}%
\begin{pgfscope}%
\pgfsetbuttcap%
\pgfsetroundjoin%
\definecolor{currentfill}{rgb}{0.000000,0.000000,0.000000}%
\pgfsetfillcolor{currentfill}%
\pgfsetlinewidth{0.501875pt}%
\definecolor{currentstroke}{rgb}{0.000000,0.000000,0.000000}%
\pgfsetstrokecolor{currentstroke}%
\pgfsetdash{}{0pt}%
\pgfsys@defobject{currentmarker}{\pgfqpoint{0.000000in}{0.000000in}}{\pgfqpoint{0.020833in}{0.000000in}}{%
\pgfpathmoveto{\pgfqpoint{0.000000in}{0.000000in}}%
\pgfpathlineto{\pgfqpoint{0.020833in}{0.000000in}}%
\pgfusepath{stroke,fill}%
}%
\begin{pgfscope}%
\pgfsys@transformshift{0.249462in}{1.416479in}%
\pgfsys@useobject{currentmarker}{}%
\end{pgfscope}%
\end{pgfscope}%
\begin{pgfscope}%
\pgfsetbuttcap%
\pgfsetroundjoin%
\definecolor{currentfill}{rgb}{0.000000,0.000000,0.000000}%
\pgfsetfillcolor{currentfill}%
\pgfsetlinewidth{0.501875pt}%
\definecolor{currentstroke}{rgb}{0.000000,0.000000,0.000000}%
\pgfsetstrokecolor{currentstroke}%
\pgfsetdash{}{0pt}%
\pgfsys@defobject{currentmarker}{\pgfqpoint{0.000000in}{0.000000in}}{\pgfqpoint{0.020833in}{0.000000in}}{%
\pgfpathmoveto{\pgfqpoint{0.000000in}{0.000000in}}%
\pgfpathlineto{\pgfqpoint{0.020833in}{0.000000in}}%
\pgfusepath{stroke,fill}%
}%
\begin{pgfscope}%
\pgfsys@transformshift{0.249462in}{1.484207in}%
\pgfsys@useobject{currentmarker}{}%
\end{pgfscope}%
\end{pgfscope}%
\begin{pgfscope}%
\pgfsetbuttcap%
\pgfsetroundjoin%
\definecolor{currentfill}{rgb}{0.000000,0.000000,0.000000}%
\pgfsetfillcolor{currentfill}%
\pgfsetlinewidth{0.501875pt}%
\definecolor{currentstroke}{rgb}{0.000000,0.000000,0.000000}%
\pgfsetstrokecolor{currentstroke}%
\pgfsetdash{}{0pt}%
\pgfsys@defobject{currentmarker}{\pgfqpoint{0.000000in}{0.000000in}}{\pgfqpoint{0.020833in}{0.000000in}}{%
\pgfpathmoveto{\pgfqpoint{0.000000in}{0.000000in}}%
\pgfpathlineto{\pgfqpoint{0.020833in}{0.000000in}}%
\pgfusepath{stroke,fill}%
}%
\begin{pgfscope}%
\pgfsys@transformshift{0.249462in}{1.619663in}%
\pgfsys@useobject{currentmarker}{}%
\end{pgfscope}%
\end{pgfscope}%
\begin{pgfscope}%
\pgfsetbuttcap%
\pgfsetroundjoin%
\definecolor{currentfill}{rgb}{0.000000,0.000000,0.000000}%
\pgfsetfillcolor{currentfill}%
\pgfsetlinewidth{0.501875pt}%
\definecolor{currentstroke}{rgb}{0.000000,0.000000,0.000000}%
\pgfsetstrokecolor{currentstroke}%
\pgfsetdash{}{0pt}%
\pgfsys@defobject{currentmarker}{\pgfqpoint{0.000000in}{0.000000in}}{\pgfqpoint{0.020833in}{0.000000in}}{%
\pgfpathmoveto{\pgfqpoint{0.000000in}{0.000000in}}%
\pgfpathlineto{\pgfqpoint{0.020833in}{0.000000in}}%
\pgfusepath{stroke,fill}%
}%
\begin{pgfscope}%
\pgfsys@transformshift{0.249462in}{1.687391in}%
\pgfsys@useobject{currentmarker}{}%
\end{pgfscope}%
\end{pgfscope}%
\begin{pgfscope}%
\pgfpathrectangle{\pgfqpoint{0.249462in}{0.197377in}}{\pgfqpoint{2.325000in}{1.540000in}}%
\pgfusepath{clip}%
\pgfsetbuttcap%
\pgfsetroundjoin%
\pgfsetlinewidth{0.501875pt}%
\definecolor{currentstroke}{rgb}{0.501961,0.501961,0.501961}%
\pgfsetstrokecolor{currentstroke}%
\pgfsetstrokeopacity{0.800000}%
\pgfsetdash{{1.850000pt}{0.800000pt}}{0.000000pt}%
\pgfpathmoveto{\pgfqpoint{0.832140in}{0.197377in}}%
\pgfpathlineto{\pgfqpoint{0.832140in}{1.737377in}}%
\pgfusepath{stroke}%
\end{pgfscope}%
\begin{pgfscope}%
\pgfpathrectangle{\pgfqpoint{0.249462in}{0.197377in}}{\pgfqpoint{2.325000in}{1.540000in}}%
\pgfusepath{clip}%
\pgfsetbuttcap%
\pgfsetroundjoin%
\pgfsetlinewidth{0.501875pt}%
\definecolor{currentstroke}{rgb}{0.501961,0.501961,0.501961}%
\pgfsetstrokecolor{currentstroke}%
\pgfsetstrokeopacity{0.800000}%
\pgfsetdash{{1.850000pt}{0.800000pt}}{0.000000pt}%
\pgfpathmoveto{\pgfqpoint{1.403393in}{0.197377in}}%
\pgfpathlineto{\pgfqpoint{1.403393in}{1.737377in}}%
\pgfusepath{stroke}%
\end{pgfscope}%
\begin{pgfscope}%
\pgfpathrectangle{\pgfqpoint{0.249462in}{0.197377in}}{\pgfqpoint{2.325000in}{1.540000in}}%
\pgfusepath{clip}%
\pgfsetbuttcap%
\pgfsetroundjoin%
\pgfsetlinewidth{0.501875pt}%
\definecolor{currentstroke}{rgb}{0.501961,0.501961,0.501961}%
\pgfsetstrokecolor{currentstroke}%
\pgfsetstrokeopacity{0.800000}%
\pgfsetdash{{1.850000pt}{0.800000pt}}{0.000000pt}%
\pgfpathmoveto{\pgfqpoint{1.974646in}{0.197377in}}%
\pgfpathlineto{\pgfqpoint{1.974646in}{1.737377in}}%
\pgfusepath{stroke}%
\end{pgfscope}%
\begin{pgfscope}%
\pgfsetrectcap%
\pgfsetmiterjoin%
\pgfsetlinewidth{0.501875pt}%
\definecolor{currentstroke}{rgb}{0.000000,0.000000,0.000000}%
\pgfsetstrokecolor{currentstroke}%
\pgfsetdash{}{0pt}%
\pgfpathmoveto{\pgfqpoint{0.249462in}{0.197377in}}%
\pgfpathlineto{\pgfqpoint{0.249462in}{1.737377in}}%
\pgfusepath{stroke}%
\end{pgfscope}%
\begin{pgfscope}%
\pgfsetrectcap%
\pgfsetmiterjoin%
\pgfsetlinewidth{0.501875pt}%
\definecolor{currentstroke}{rgb}{0.000000,0.000000,0.000000}%
\pgfsetstrokecolor{currentstroke}%
\pgfsetdash{}{0pt}%
\pgfpathmoveto{\pgfqpoint{2.574462in}{0.197377in}}%
\pgfpathlineto{\pgfqpoint{2.574462in}{1.737377in}}%
\pgfusepath{stroke}%
\end{pgfscope}%
\begin{pgfscope}%
\pgfsetrectcap%
\pgfsetmiterjoin%
\pgfsetlinewidth{0.501875pt}%
\definecolor{currentstroke}{rgb}{0.000000,0.000000,0.000000}%
\pgfsetstrokecolor{currentstroke}%
\pgfsetdash{}{0pt}%
\pgfpathmoveto{\pgfqpoint{0.249462in}{0.197377in}}%
\pgfpathlineto{\pgfqpoint{2.574462in}{0.197377in}}%
\pgfusepath{stroke}%
\end{pgfscope}%
\begin{pgfscope}%
\pgfsetrectcap%
\pgfsetmiterjoin%
\pgfsetlinewidth{0.501875pt}%
\definecolor{currentstroke}{rgb}{0.000000,0.000000,0.000000}%
\pgfsetstrokecolor{currentstroke}%
\pgfsetdash{}{0pt}%
\pgfpathmoveto{\pgfqpoint{0.249462in}{1.737377in}}%
\pgfpathlineto{\pgfqpoint{2.574462in}{1.737377in}}%
\pgfusepath{stroke}%
\end{pgfscope}%
\begin{pgfscope}%
\pgfpathrectangle{\pgfqpoint{0.249462in}{0.197377in}}{\pgfqpoint{2.325000in}{1.540000in}}%
\pgfusepath{clip}%
\pgfsetbuttcap%
\pgfsetmiterjoin%
\definecolor{currentfill}{rgb}{0.000000,0.000000,0.000000}%
\pgfsetfillcolor{currentfill}%
\pgfsetlinewidth{0.000000pt}%
\definecolor{currentstroke}{rgb}{0.000000,0.000000,0.000000}%
\pgfsetstrokecolor{currentstroke}%
\pgfsetstrokeopacity{0.000000}%
\pgfsetdash{}{0pt}%
\pgfpathmoveto{\pgfqpoint{0.355144in}{0.197377in}}%
\pgfpathlineto{\pgfqpoint{0.440832in}{0.197377in}}%
\pgfpathlineto{\pgfqpoint{0.440832in}{1.423457in}}%
\pgfpathlineto{\pgfqpoint{0.355144in}{1.423457in}}%
\pgfpathlineto{\pgfqpoint{0.355144in}{0.197377in}}%
\pgfpathclose%
\pgfusepath{fill}%
\end{pgfscope}%
\begin{pgfscope}%
\pgfpathrectangle{\pgfqpoint{0.249462in}{0.197377in}}{\pgfqpoint{2.325000in}{1.540000in}}%
\pgfusepath{clip}%
\pgfsetbuttcap%
\pgfsetmiterjoin%
\definecolor{currentfill}{rgb}{0.000000,0.000000,0.000000}%
\pgfsetfillcolor{currentfill}%
\pgfsetlinewidth{0.000000pt}%
\definecolor{currentstroke}{rgb}{0.000000,0.000000,0.000000}%
\pgfsetstrokecolor{currentstroke}%
\pgfsetstrokeopacity{0.000000}%
\pgfsetdash{}{0pt}%
\pgfpathmoveto{\pgfqpoint{0.926397in}{0.197377in}}%
\pgfpathlineto{\pgfqpoint{1.012085in}{0.197377in}}%
\pgfpathlineto{\pgfqpoint{1.012085in}{0.197377in}}%
\pgfpathlineto{\pgfqpoint{0.926397in}{0.197377in}}%
\pgfpathlineto{\pgfqpoint{0.926397in}{0.197377in}}%
\pgfpathclose%
\pgfusepath{fill}%
\end{pgfscope}%
\begin{pgfscope}%
\pgfpathrectangle{\pgfqpoint{0.249462in}{0.197377in}}{\pgfqpoint{2.325000in}{1.540000in}}%
\pgfusepath{clip}%
\pgfsetbuttcap%
\pgfsetmiterjoin%
\definecolor{currentfill}{rgb}{0.000000,0.000000,0.000000}%
\pgfsetfillcolor{currentfill}%
\pgfsetlinewidth{0.000000pt}%
\definecolor{currentstroke}{rgb}{0.000000,0.000000,0.000000}%
\pgfsetstrokecolor{currentstroke}%
\pgfsetstrokeopacity{0.000000}%
\pgfsetdash{}{0pt}%
\pgfpathmoveto{\pgfqpoint{1.497650in}{0.197377in}}%
\pgfpathlineto{\pgfqpoint{1.583338in}{0.197377in}}%
\pgfpathlineto{\pgfqpoint{1.583338in}{1.047283in}}%
\pgfpathlineto{\pgfqpoint{1.497650in}{1.047283in}}%
\pgfpathlineto{\pgfqpoint{1.497650in}{0.197377in}}%
\pgfpathclose%
\pgfusepath{fill}%
\end{pgfscope}%
\begin{pgfscope}%
\pgfpathrectangle{\pgfqpoint{0.249462in}{0.197377in}}{\pgfqpoint{2.325000in}{1.540000in}}%
\pgfusepath{clip}%
\pgfsetbuttcap%
\pgfsetmiterjoin%
\definecolor{currentfill}{rgb}{0.000000,0.000000,0.000000}%
\pgfsetfillcolor{currentfill}%
\pgfsetlinewidth{0.000000pt}%
\definecolor{currentstroke}{rgb}{0.000000,0.000000,0.000000}%
\pgfsetstrokecolor{currentstroke}%
\pgfsetstrokeopacity{0.000000}%
\pgfsetdash{}{0pt}%
\pgfpathmoveto{\pgfqpoint{2.068903in}{0.197377in}}%
\pgfpathlineto{\pgfqpoint{2.154591in}{0.197377in}}%
\pgfpathlineto{\pgfqpoint{2.154591in}{1.541683in}}%
\pgfpathlineto{\pgfqpoint{2.068903in}{1.541683in}}%
\pgfpathlineto{\pgfqpoint{2.068903in}{0.197377in}}%
\pgfpathclose%
\pgfusepath{fill}%
\end{pgfscope}%
\begin{pgfscope}%
\pgfpathrectangle{\pgfqpoint{0.249462in}{0.197377in}}{\pgfqpoint{2.325000in}{1.540000in}}%
\pgfusepath{clip}%
\pgfsetbuttcap%
\pgfsetmiterjoin%
\definecolor{currentfill}{rgb}{1.000000,0.000000,0.000000}%
\pgfsetfillcolor{currentfill}%
\pgfsetlinewidth{0.000000pt}%
\definecolor{currentstroke}{rgb}{0.000000,0.000000,0.000000}%
\pgfsetstrokecolor{currentstroke}%
\pgfsetstrokeopacity{0.000000}%
\pgfsetdash{}{0pt}%
\pgfpathmoveto{\pgfqpoint{0.440832in}{0.197377in}}%
\pgfpathlineto{\pgfqpoint{0.526520in}{0.197377in}}%
\pgfpathlineto{\pgfqpoint{0.526520in}{1.511093in}}%
\pgfpathlineto{\pgfqpoint{0.440832in}{1.511093in}}%
\pgfpathlineto{\pgfqpoint{0.440832in}{0.197377in}}%
\pgfpathclose%
\pgfusepath{fill}%
\end{pgfscope}%
\begin{pgfscope}%
\pgfpathrectangle{\pgfqpoint{0.249462in}{0.197377in}}{\pgfqpoint{2.325000in}{1.540000in}}%
\pgfusepath{clip}%
\pgfsetbuttcap%
\pgfsetmiterjoin%
\definecolor{currentfill}{rgb}{1.000000,0.000000,0.000000}%
\pgfsetfillcolor{currentfill}%
\pgfsetlinewidth{0.000000pt}%
\definecolor{currentstroke}{rgb}{0.000000,0.000000,0.000000}%
\pgfsetstrokecolor{currentstroke}%
\pgfsetstrokeopacity{0.000000}%
\pgfsetdash{}{0pt}%
\pgfpathmoveto{\pgfqpoint{1.012085in}{0.197377in}}%
\pgfpathlineto{\pgfqpoint{1.097773in}{0.197377in}}%
\pgfpathlineto{\pgfqpoint{1.097773in}{0.197377in}}%
\pgfpathlineto{\pgfqpoint{1.012085in}{0.197377in}}%
\pgfpathlineto{\pgfqpoint{1.012085in}{0.197377in}}%
\pgfpathclose%
\pgfusepath{fill}%
\end{pgfscope}%
\begin{pgfscope}%
\pgfpathrectangle{\pgfqpoint{0.249462in}{0.197377in}}{\pgfqpoint{2.325000in}{1.540000in}}%
\pgfusepath{clip}%
\pgfsetbuttcap%
\pgfsetmiterjoin%
\definecolor{currentfill}{rgb}{1.000000,0.000000,0.000000}%
\pgfsetfillcolor{currentfill}%
\pgfsetlinewidth{0.000000pt}%
\definecolor{currentstroke}{rgb}{0.000000,0.000000,0.000000}%
\pgfsetstrokecolor{currentstroke}%
\pgfsetstrokeopacity{0.000000}%
\pgfsetdash{}{0pt}%
\pgfpathmoveto{\pgfqpoint{1.583338in}{0.197377in}}%
\pgfpathlineto{\pgfqpoint{1.669026in}{0.197377in}}%
\pgfpathlineto{\pgfqpoint{1.669026in}{1.250665in}}%
\pgfpathlineto{\pgfqpoint{1.583338in}{1.250665in}}%
\pgfpathlineto{\pgfqpoint{1.583338in}{0.197377in}}%
\pgfpathclose%
\pgfusepath{fill}%
\end{pgfscope}%
\begin{pgfscope}%
\pgfpathrectangle{\pgfqpoint{0.249462in}{0.197377in}}{\pgfqpoint{2.325000in}{1.540000in}}%
\pgfusepath{clip}%
\pgfsetbuttcap%
\pgfsetmiterjoin%
\definecolor{currentfill}{rgb}{1.000000,0.000000,0.000000}%
\pgfsetfillcolor{currentfill}%
\pgfsetlinewidth{0.000000pt}%
\definecolor{currentstroke}{rgb}{0.000000,0.000000,0.000000}%
\pgfsetstrokecolor{currentstroke}%
\pgfsetstrokeopacity{0.000000}%
\pgfsetdash{}{0pt}%
\pgfpathmoveto{\pgfqpoint{2.154591in}{0.197377in}}%
\pgfpathlineto{\pgfqpoint{2.240279in}{0.197377in}}%
\pgfpathlineto{\pgfqpoint{2.240279in}{1.636760in}}%
\pgfpathlineto{\pgfqpoint{2.154591in}{1.636760in}}%
\pgfpathlineto{\pgfqpoint{2.154591in}{0.197377in}}%
\pgfpathclose%
\pgfusepath{fill}%
\end{pgfscope}%
\begin{pgfscope}%
\pgfpathrectangle{\pgfqpoint{0.249462in}{0.197377in}}{\pgfqpoint{2.325000in}{1.540000in}}%
\pgfusepath{clip}%
\pgfsetbuttcap%
\pgfsetmiterjoin%
\definecolor{currentfill}{rgb}{0.000000,0.000000,1.000000}%
\pgfsetfillcolor{currentfill}%
\pgfsetlinewidth{0.000000pt}%
\definecolor{currentstroke}{rgb}{0.000000,0.000000,0.000000}%
\pgfsetstrokecolor{currentstroke}%
\pgfsetstrokeopacity{0.000000}%
\pgfsetdash{}{0pt}%
\pgfpathmoveto{\pgfqpoint{0.583645in}{0.197377in}}%
\pgfpathlineto{\pgfqpoint{0.669333in}{0.197377in}}%
\pgfpathlineto{\pgfqpoint{0.669333in}{1.521841in}}%
\pgfpathlineto{\pgfqpoint{0.583645in}{1.521841in}}%
\pgfpathlineto{\pgfqpoint{0.583645in}{0.197377in}}%
\pgfpathclose%
\pgfusepath{fill}%
\end{pgfscope}%
\begin{pgfscope}%
\pgfpathrectangle{\pgfqpoint{0.249462in}{0.197377in}}{\pgfqpoint{2.325000in}{1.540000in}}%
\pgfusepath{clip}%
\pgfsetbuttcap%
\pgfsetmiterjoin%
\definecolor{currentfill}{rgb}{0.000000,0.000000,1.000000}%
\pgfsetfillcolor{currentfill}%
\pgfsetlinewidth{0.000000pt}%
\definecolor{currentstroke}{rgb}{0.000000,0.000000,0.000000}%
\pgfsetstrokecolor{currentstroke}%
\pgfsetstrokeopacity{0.000000}%
\pgfsetdash{}{0pt}%
\pgfpathmoveto{\pgfqpoint{1.154898in}{0.197377in}}%
\pgfpathlineto{\pgfqpoint{1.240586in}{0.197377in}}%
\pgfpathlineto{\pgfqpoint{1.240586in}{0.197377in}}%
\pgfpathlineto{\pgfqpoint{1.154898in}{0.197377in}}%
\pgfpathlineto{\pgfqpoint{1.154898in}{0.197377in}}%
\pgfpathclose%
\pgfusepath{fill}%
\end{pgfscope}%
\begin{pgfscope}%
\pgfpathrectangle{\pgfqpoint{0.249462in}{0.197377in}}{\pgfqpoint{2.325000in}{1.540000in}}%
\pgfusepath{clip}%
\pgfsetbuttcap%
\pgfsetmiterjoin%
\definecolor{currentfill}{rgb}{0.000000,0.000000,1.000000}%
\pgfsetfillcolor{currentfill}%
\pgfsetlinewidth{0.000000pt}%
\definecolor{currentstroke}{rgb}{0.000000,0.000000,0.000000}%
\pgfsetstrokecolor{currentstroke}%
\pgfsetstrokeopacity{0.000000}%
\pgfsetdash{}{0pt}%
\pgfpathmoveto{\pgfqpoint{1.726151in}{0.197377in}}%
\pgfpathlineto{\pgfqpoint{1.811839in}{0.197377in}}%
\pgfpathlineto{\pgfqpoint{1.811839in}{1.287869in}}%
\pgfpathlineto{\pgfqpoint{1.726151in}{1.287869in}}%
\pgfpathlineto{\pgfqpoint{1.726151in}{0.197377in}}%
\pgfpathclose%
\pgfusepath{fill}%
\end{pgfscope}%
\begin{pgfscope}%
\pgfpathrectangle{\pgfqpoint{0.249462in}{0.197377in}}{\pgfqpoint{2.325000in}{1.540000in}}%
\pgfusepath{clip}%
\pgfsetbuttcap%
\pgfsetmiterjoin%
\definecolor{currentfill}{rgb}{0.000000,0.000000,1.000000}%
\pgfsetfillcolor{currentfill}%
\pgfsetlinewidth{0.000000pt}%
\definecolor{currentstroke}{rgb}{0.000000,0.000000,0.000000}%
\pgfsetstrokecolor{currentstroke}%
\pgfsetstrokeopacity{0.000000}%
\pgfsetdash{}{0pt}%
\pgfpathmoveto{\pgfqpoint{2.297404in}{0.197377in}}%
\pgfpathlineto{\pgfqpoint{2.383092in}{0.197377in}}%
\pgfpathlineto{\pgfqpoint{2.383092in}{1.664043in}}%
\pgfpathlineto{\pgfqpoint{2.297404in}{1.664043in}}%
\pgfpathlineto{\pgfqpoint{2.297404in}{0.197377in}}%
\pgfpathclose%
\pgfusepath{fill}%
\end{pgfscope}%
\begin{pgfscope}%
\pgfpathrectangle{\pgfqpoint{0.249462in}{0.197377in}}{\pgfqpoint{2.325000in}{1.540000in}}%
\pgfusepath{clip}%
\pgfsetbuttcap%
\pgfsetmiterjoin%
\definecolor{currentfill}{rgb}{0.000000,0.500000,0.000000}%
\pgfsetfillcolor{currentfill}%
\pgfsetlinewidth{0.000000pt}%
\definecolor{currentstroke}{rgb}{0.000000,0.000000,0.000000}%
\pgfsetstrokecolor{currentstroke}%
\pgfsetstrokeopacity{0.000000}%
\pgfsetdash{}{0pt}%
\pgfpathmoveto{\pgfqpoint{0.669333in}{0.197377in}}%
\pgfpathlineto{\pgfqpoint{0.755021in}{0.197377in}}%
\pgfpathlineto{\pgfqpoint{0.755021in}{1.244051in}}%
\pgfpathlineto{\pgfqpoint{0.669333in}{1.244051in}}%
\pgfpathlineto{\pgfqpoint{0.669333in}{0.197377in}}%
\pgfpathclose%
\pgfusepath{fill}%
\end{pgfscope}%
\begin{pgfscope}%
\pgfpathrectangle{\pgfqpoint{0.249462in}{0.197377in}}{\pgfqpoint{2.325000in}{1.540000in}}%
\pgfusepath{clip}%
\pgfsetbuttcap%
\pgfsetmiterjoin%
\definecolor{currentfill}{rgb}{0.000000,0.500000,0.000000}%
\pgfsetfillcolor{currentfill}%
\pgfsetlinewidth{0.000000pt}%
\definecolor{currentstroke}{rgb}{0.000000,0.000000,0.000000}%
\pgfsetstrokecolor{currentstroke}%
\pgfsetstrokeopacity{0.000000}%
\pgfsetdash{}{0pt}%
\pgfpathmoveto{\pgfqpoint{1.240586in}{0.197377in}}%
\pgfpathlineto{\pgfqpoint{1.326274in}{0.197377in}}%
\pgfpathlineto{\pgfqpoint{1.326274in}{0.197377in}}%
\pgfpathlineto{\pgfqpoint{1.240586in}{0.197377in}}%
\pgfpathlineto{\pgfqpoint{1.240586in}{0.197377in}}%
\pgfpathclose%
\pgfusepath{fill}%
\end{pgfscope}%
\begin{pgfscope}%
\pgfpathrectangle{\pgfqpoint{0.249462in}{0.197377in}}{\pgfqpoint{2.325000in}{1.540000in}}%
\pgfusepath{clip}%
\pgfsetbuttcap%
\pgfsetmiterjoin%
\definecolor{currentfill}{rgb}{0.000000,0.500000,0.000000}%
\pgfsetfillcolor{currentfill}%
\pgfsetlinewidth{0.000000pt}%
\definecolor{currentstroke}{rgb}{0.000000,0.000000,0.000000}%
\pgfsetstrokecolor{currentstroke}%
\pgfsetstrokeopacity{0.000000}%
\pgfsetdash{}{0pt}%
\pgfpathmoveto{\pgfqpoint{1.811839in}{0.197377in}}%
\pgfpathlineto{\pgfqpoint{1.897527in}{0.197377in}}%
\pgfpathlineto{\pgfqpoint{1.897527in}{0.959233in}}%
\pgfpathlineto{\pgfqpoint{1.811839in}{0.959233in}}%
\pgfpathlineto{\pgfqpoint{1.811839in}{0.197377in}}%
\pgfpathclose%
\pgfusepath{fill}%
\end{pgfscope}%
\begin{pgfscope}%
\pgfpathrectangle{\pgfqpoint{0.249462in}{0.197377in}}{\pgfqpoint{2.325000in}{1.540000in}}%
\pgfusepath{clip}%
\pgfsetbuttcap%
\pgfsetmiterjoin%
\definecolor{currentfill}{rgb}{0.000000,0.500000,0.000000}%
\pgfsetfillcolor{currentfill}%
\pgfsetlinewidth{0.000000pt}%
\definecolor{currentstroke}{rgb}{0.000000,0.000000,0.000000}%
\pgfsetstrokecolor{currentstroke}%
\pgfsetstrokeopacity{0.000000}%
\pgfsetdash{}{0pt}%
\pgfpathmoveto{\pgfqpoint{2.383092in}{0.197377in}}%
\pgfpathlineto{\pgfqpoint{2.468780in}{0.197377in}}%
\pgfpathlineto{\pgfqpoint{2.468780in}{1.372198in}}%
\pgfpathlineto{\pgfqpoint{2.383092in}{1.372198in}}%
\pgfpathlineto{\pgfqpoint{2.383092in}{0.197377in}}%
\pgfpathclose%
\pgfusepath{fill}%
\end{pgfscope}%
\end{pgfpicture}%
\makeatother%
\endgroup%

%% file: HIXray-object-size.pgf
\begingroup%
\makeatletter%
\begin{pgfpicture}%
\pgfpathrectangle{\pgfpointorigin}{\pgfqpoint{2.624462in}{1.787377in}}%
\pgfusepath{use as bounding box, clip}%
\begin{pgfscope}%
\pgfsetbuttcap%
\pgfsetmiterjoin%
\definecolor{currentfill}{rgb}{1.000000,1.000000,1.000000}%
\pgfsetfillcolor{currentfill}%
\pgfsetlinewidth{0.000000pt}%
\definecolor{currentstroke}{rgb}{1.000000,1.000000,1.000000}%
\pgfsetstrokecolor{currentstroke}%
\pgfsetdash{}{0pt}%
\pgfpathmoveto{\pgfqpoint{0.000000in}{0.000000in}}%
\pgfpathlineto{\pgfqpoint{2.624462in}{0.000000in}}%
\pgfpathlineto{\pgfqpoint{2.624462in}{1.787377in}}%
\pgfpathlineto{\pgfqpoint{0.000000in}{1.787377in}}%
\pgfpathlineto{\pgfqpoint{0.000000in}{0.000000in}}%
\pgfpathclose%
\pgfusepath{fill}%
\end{pgfscope}%
\begin{pgfscope}%
\pgfsetbuttcap%
\pgfsetmiterjoin%
\definecolor{currentfill}{rgb}{1.000000,1.000000,1.000000}%
\pgfsetfillcolor{currentfill}%
\pgfsetlinewidth{0.000000pt}%
\definecolor{currentstroke}{rgb}{0.000000,0.000000,0.000000}%
\pgfsetstrokecolor{currentstroke}%
\pgfsetstrokeopacity{0.000000}%
\pgfsetdash{}{0pt}%
\pgfpathmoveto{\pgfqpoint{0.249462in}{0.197377in}}%
\pgfpathlineto{\pgfqpoint{2.574462in}{0.197377in}}%
\pgfpathlineto{\pgfqpoint{2.574462in}{1.737377in}}%
\pgfpathlineto{\pgfqpoint{0.249462in}{1.737377in}}%
\pgfpathlineto{\pgfqpoint{0.249462in}{0.197377in}}%
\pgfpathclose%
\pgfusepath{fill}%
\end{pgfscope}%
\begin{pgfscope}%
\pgfpathrectangle{\pgfqpoint{0.249462in}{0.197377in}}{\pgfqpoint{2.325000in}{1.540000in}}%
\pgfusepath{clip}%
\pgfsetrectcap%
\pgfsetroundjoin%
\pgfsetlinewidth{0.501875pt}%
\definecolor{currentstroke}{rgb}{0.690196,0.690196,0.690196}%
\pgfsetstrokecolor{currentstroke}%
\pgfsetstrokeopacity{0.500000}%
\pgfsetdash{}{0pt}%
\pgfpathmoveto{\pgfqpoint{0.552226in}{0.197377in}}%
\pgfpathlineto{\pgfqpoint{0.552226in}{1.737377in}}%
\pgfusepath{stroke}%
\end{pgfscope}%
\begin{pgfscope}%
\pgfsetbuttcap%
\pgfsetroundjoin%
\definecolor{currentfill}{rgb}{0.000000,0.000000,0.000000}%
\pgfsetfillcolor{currentfill}%
\pgfsetlinewidth{0.501875pt}%
\definecolor{currentstroke}{rgb}{0.000000,0.000000,0.000000}%
\pgfsetstrokecolor{currentstroke}%
\pgfsetdash{}{0pt}%
\pgfsys@defobject{currentmarker}{\pgfqpoint{0.000000in}{0.000000in}}{\pgfqpoint{0.000000in}{0.041667in}}{%
\pgfpathmoveto{\pgfqpoint{0.000000in}{0.000000in}}%
\pgfpathlineto{\pgfqpoint{0.000000in}{0.041667in}}%
\pgfusepath{stroke,fill}%
}%
\begin{pgfscope}%
\pgfsys@transformshift{0.552226in}{0.197377in}%
\pgfsys@useobject{currentmarker}{}%
\end{pgfscope}%
\end{pgfscope}%
\begin{pgfscope}%
\definecolor{textcolor}{rgb}{0.000000,0.000000,0.000000}%
\pgfsetstrokecolor{textcolor}%
\pgfsetfillcolor{textcolor}%
\pgftext[x=0.552226in,y=0.148766in,,top]{\color{textcolor}{\rmfamily\fontsize{8.000000}{9.600000}\selectfont\catcode`\^=\active\def^{\ifmmode\sp\else\^{}\fi}\catcode`\%=\active\def
\end{pgfscope}%
\begin{pgfscope}%
\pgfpathrectangle{\pgfqpoint{0.249462in}{0.197377in}}{\pgfqpoint{2.325000in}{1.540000in}}%
\pgfusepath{clip}%
\pgfsetrectcap%
\pgfsetroundjoin%
\pgfsetlinewidth{0.501875pt}%
\definecolor{currentstroke}{rgb}{0.690196,0.690196,0.690196}%
\pgfsetstrokecolor{currentstroke}%
\pgfsetstrokeopacity{0.500000}%
\pgfsetdash{}{0pt}%
\pgfpathmoveto{\pgfqpoint{1.123479in}{0.197377in}}%
\pgfpathlineto{\pgfqpoint{1.123479in}{1.737377in}}%
\pgfusepath{stroke}%
\end{pgfscope}%
\begin{pgfscope}%
\pgfsetbuttcap%
\pgfsetroundjoin%
\definecolor{currentfill}{rgb}{0.000000,0.000000,0.000000}%
\pgfsetfillcolor{currentfill}%
\pgfsetlinewidth{0.501875pt}%
\definecolor{currentstroke}{rgb}{0.000000,0.000000,0.000000}%
\pgfsetstrokecolor{currentstroke}%
\pgfsetdash{}{0pt}%
\pgfsys@defobject{currentmarker}{\pgfqpoint{0.000000in}{0.000000in}}{\pgfqpoint{0.000000in}{0.041667in}}{%
\pgfpathmoveto{\pgfqpoint{0.000000in}{0.000000in}}%
\pgfpathlineto{\pgfqpoint{0.000000in}{0.041667in}}%
\pgfusepath{stroke,fill}%
}%
\begin{pgfscope}%
\pgfsys@transformshift{1.123479in}{0.197377in}%
\pgfsys@useobject{currentmarker}{}%
\end{pgfscope}%
\end{pgfscope}%
\begin{pgfscope}%
\definecolor{textcolor}{rgb}{0.000000,0.000000,0.000000}%
\pgfsetstrokecolor{textcolor}%
\pgfsetfillcolor{textcolor}%
\pgftext[x=1.123479in,y=0.148766in,,top]{\color{textcolor}{\rmfamily\fontsize{8.000000}{9.600000}\selectfont\catcode`\^=\active\def^{\ifmmode\sp\else\^{}\fi}\catcode`\%=\active\def
\end{pgfscope}%
\begin{pgfscope}%
\pgfpathrectangle{\pgfqpoint{0.249462in}{0.197377in}}{\pgfqpoint{2.325000in}{1.540000in}}%
\pgfusepath{clip}%
\pgfsetrectcap%
\pgfsetroundjoin%
\pgfsetlinewidth{0.501875pt}%
\definecolor{currentstroke}{rgb}{0.690196,0.690196,0.690196}%
\pgfsetstrokecolor{currentstroke}%
\pgfsetstrokeopacity{0.500000}%
\pgfsetdash{}{0pt}%
\pgfpathmoveto{\pgfqpoint{1.694732in}{0.197377in}}%
\pgfpathlineto{\pgfqpoint{1.694732in}{1.737377in}}%
\pgfusepath{stroke}%
\end{pgfscope}%
\begin{pgfscope}%
\pgfsetbuttcap%
\pgfsetroundjoin%
\definecolor{currentfill}{rgb}{0.000000,0.000000,0.000000}%
\pgfsetfillcolor{currentfill}%
\pgfsetlinewidth{0.501875pt}%
\definecolor{currentstroke}{rgb}{0.000000,0.000000,0.000000}%
\pgfsetstrokecolor{currentstroke}%
\pgfsetdash{}{0pt}%
\pgfsys@defobject{currentmarker}{\pgfqpoint{0.000000in}{0.000000in}}{\pgfqpoint{0.000000in}{0.041667in}}{%
\pgfpathmoveto{\pgfqpoint{0.000000in}{0.000000in}}%
\pgfpathlineto{\pgfqpoint{0.000000in}{0.041667in}}%
\pgfusepath{stroke,fill}%
}%
\begin{pgfscope}%
\pgfsys@transformshift{1.694732in}{0.197377in}%
\pgfsys@useobject{currentmarker}{}%
\end{pgfscope}%
\end{pgfscope}%
\begin{pgfscope}%
\definecolor{textcolor}{rgb}{0.000000,0.000000,0.000000}%
\pgfsetstrokecolor{textcolor}%
\pgfsetfillcolor{textcolor}%
\pgftext[x=1.694732in,y=0.148766in,,top]{\color{textcolor}{\rmfamily\fontsize{8.000000}{9.600000}\selectfont\catcode`\^=\active\def^{\ifmmode\sp\else\^{}\fi}\catcode`\%=\active\def
\end{pgfscope}%
\begin{pgfscope}%
\pgfpathrectangle{\pgfqpoint{0.249462in}{0.197377in}}{\pgfqpoint{2.325000in}{1.540000in}}%
\pgfusepath{clip}%
\pgfsetrectcap%
\pgfsetroundjoin%
\pgfsetlinewidth{0.501875pt}%
\definecolor{currentstroke}{rgb}{0.690196,0.690196,0.690196}%
\pgfsetstrokecolor{currentstroke}%
\pgfsetstrokeopacity{0.500000}%
\pgfsetdash{}{0pt}%
\pgfpathmoveto{\pgfqpoint{2.265985in}{0.197377in}}%
\pgfpathlineto{\pgfqpoint{2.265985in}{1.737377in}}%
\pgfusepath{stroke}%
\end{pgfscope}%
\begin{pgfscope}%
\pgfsetbuttcap%
\pgfsetroundjoin%
\definecolor{currentfill}{rgb}{0.000000,0.000000,0.000000}%
\pgfsetfillcolor{currentfill}%
\pgfsetlinewidth{0.501875pt}%
\definecolor{currentstroke}{rgb}{0.000000,0.000000,0.000000}%
\pgfsetstrokecolor{currentstroke}%
\pgfsetdash{}{0pt}%
\pgfsys@defobject{currentmarker}{\pgfqpoint{0.000000in}{0.000000in}}{\pgfqpoint{0.000000in}{0.041667in}}{%
\pgfpathmoveto{\pgfqpoint{0.000000in}{0.000000in}}%
\pgfpathlineto{\pgfqpoint{0.000000in}{0.041667in}}%
\pgfusepath{stroke,fill}%
}%
\begin{pgfscope}%
\pgfsys@transformshift{2.265985in}{0.197377in}%
\pgfsys@useobject{currentmarker}{}%
\end{pgfscope}%
\end{pgfscope}%
\begin{pgfscope}%
\definecolor{textcolor}{rgb}{0.000000,0.000000,0.000000}%
\pgfsetstrokecolor{textcolor}%
\pgfsetfillcolor{textcolor}%
\pgftext[x=2.265985in,y=0.148766in,,top]{\color{textcolor}{\rmfamily\fontsize{8.000000}{9.600000}\selectfont\catcode`\^=\active\def^{\ifmmode\sp\else\^{}\fi}\catcode`\%=\active\def
\end{pgfscope}%
\begin{pgfscope}%
\pgfpathrectangle{\pgfqpoint{0.249462in}{0.197377in}}{\pgfqpoint{2.325000in}{1.540000in}}%
\pgfusepath{clip}%
\pgfsetrectcap%
\pgfsetroundjoin%
\pgfsetlinewidth{0.501875pt}%
\definecolor{currentstroke}{rgb}{0.690196,0.690196,0.690196}%
\pgfsetstrokecolor{currentstroke}%
\pgfsetstrokeopacity{0.500000}%
\pgfsetdash{}{0pt}%
\pgfpathmoveto{\pgfqpoint{0.249462in}{0.197377in}}%
\pgfpathlineto{\pgfqpoint{2.574462in}{0.197377in}}%
\pgfusepath{stroke}%
\end{pgfscope}%
\begin{pgfscope}%
\pgfsetbuttcap%
\pgfsetroundjoin%
\definecolor{currentfill}{rgb}{0.000000,0.000000,0.000000}%
\pgfsetfillcolor{currentfill}%
\pgfsetlinewidth{0.501875pt}%
\definecolor{currentstroke}{rgb}{0.000000,0.000000,0.000000}%
\pgfsetstrokecolor{currentstroke}%
\pgfsetdash{}{0pt}%
\pgfsys@defobject{currentmarker}{\pgfqpoint{0.000000in}{0.000000in}}{\pgfqpoint{0.041667in}{0.000000in}}{%
\pgfpathmoveto{\pgfqpoint{0.000000in}{0.000000in}}%
\pgfpathlineto{\pgfqpoint{0.041667in}{0.000000in}}%
\pgfusepath{stroke,fill}%
}%
\begin{pgfscope}%
\pgfsys@transformshift{0.249462in}{0.197377in}%
\pgfsys@useobject{currentmarker}{}%
\end{pgfscope}%
\end{pgfscope}%
\begin{pgfscope}%
\definecolor{textcolor}{rgb}{0.000000,0.000000,0.000000}%
\pgfsetstrokecolor{textcolor}%
\pgfsetfillcolor{textcolor}%
\pgftext[x=0.050000in, y=0.158796in, left, base]{\color{textcolor}{\rmfamily\fontsize{8.000000}{9.600000}\selectfont\catcode`\^=\active\def^{\ifmmode\sp\else\^{}\fi}\catcode`\%=\active\def
\end{pgfscope}%
\begin{pgfscope}%
\pgfpathrectangle{\pgfqpoint{0.249462in}{0.197377in}}{\pgfqpoint{2.325000in}{1.540000in}}%
\pgfusepath{clip}%
\pgfsetrectcap%
\pgfsetroundjoin%
\pgfsetlinewidth{0.501875pt}%
\definecolor{currentstroke}{rgb}{0.690196,0.690196,0.690196}%
\pgfsetstrokecolor{currentstroke}%
\pgfsetstrokeopacity{0.500000}%
\pgfsetdash{}{0pt}%
\pgfpathmoveto{\pgfqpoint{0.249462in}{0.468472in}}%
\pgfpathlineto{\pgfqpoint{2.574462in}{0.468472in}}%
\pgfusepath{stroke}%
\end{pgfscope}%
\begin{pgfscope}%
\pgfsetbuttcap%
\pgfsetroundjoin%
\definecolor{currentfill}{rgb}{0.000000,0.000000,0.000000}%
\pgfsetfillcolor{currentfill}%
\pgfsetlinewidth{0.501875pt}%
\definecolor{currentstroke}{rgb}{0.000000,0.000000,0.000000}%
\pgfsetstrokecolor{currentstroke}%
\pgfsetdash{}{0pt}%
\pgfsys@defobject{currentmarker}{\pgfqpoint{0.000000in}{0.000000in}}{\pgfqpoint{0.041667in}{0.000000in}}{%
\pgfpathmoveto{\pgfqpoint{0.000000in}{0.000000in}}%
\pgfpathlineto{\pgfqpoint{0.041667in}{0.000000in}}%
\pgfusepath{stroke,fill}%
}%
\begin{pgfscope}%
\pgfsys@transformshift{0.249462in}{0.468472in}%
\pgfsys@useobject{currentmarker}{}%
\end{pgfscope}%
\end{pgfscope}%
\begin{pgfscope}%
\definecolor{textcolor}{rgb}{0.000000,0.000000,0.000000}%
\pgfsetstrokecolor{textcolor}%
\pgfsetfillcolor{textcolor}%
\pgftext[x=0.050000in, y=0.429891in, left, base]{\color{textcolor}{\rmfamily\fontsize{8.000000}{9.600000}\selectfont\catcode`\^=\active\def^{\ifmmode\sp\else\^{}\fi}\catcode`\%=\active\def
\end{pgfscope}%
\begin{pgfscope}%
\pgfpathrectangle{\pgfqpoint{0.249462in}{0.197377in}}{\pgfqpoint{2.325000in}{1.540000in}}%
\pgfusepath{clip}%
\pgfsetrectcap%
\pgfsetroundjoin%
\pgfsetlinewidth{0.501875pt}%
\definecolor{currentstroke}{rgb}{0.690196,0.690196,0.690196}%
\pgfsetstrokecolor{currentstroke}%
\pgfsetstrokeopacity{0.500000}%
\pgfsetdash{}{0pt}%
\pgfpathmoveto{\pgfqpoint{0.249462in}{0.739567in}}%
\pgfpathlineto{\pgfqpoint{2.574462in}{0.739567in}}%
\pgfusepath{stroke}%
\end{pgfscope}%
\begin{pgfscope}%
\pgfsetbuttcap%
\pgfsetroundjoin%
\definecolor{currentfill}{rgb}{0.000000,0.000000,0.000000}%
\pgfsetfillcolor{currentfill}%
\pgfsetlinewidth{0.501875pt}%
\definecolor{currentstroke}{rgb}{0.000000,0.000000,0.000000}%
\pgfsetstrokecolor{currentstroke}%
\pgfsetdash{}{0pt}%
\pgfsys@defobject{currentmarker}{\pgfqpoint{0.000000in}{0.000000in}}{\pgfqpoint{0.041667in}{0.000000in}}{%
\pgfpathmoveto{\pgfqpoint{0.000000in}{0.000000in}}%
\pgfpathlineto{\pgfqpoint{0.041667in}{0.000000in}}%
\pgfusepath{stroke,fill}%
}%
\begin{pgfscope}%
\pgfsys@transformshift{0.249462in}{0.739567in}%
\pgfsys@useobject{currentmarker}{}%
\end{pgfscope}%
\end{pgfscope}%
\begin{pgfscope}%
\definecolor{textcolor}{rgb}{0.000000,0.000000,0.000000}%
\pgfsetstrokecolor{textcolor}%
\pgfsetfillcolor{textcolor}%
\pgftext[x=0.050000in, y=0.700987in, left, base]{\color{textcolor}{\rmfamily\fontsize{8.000000}{9.600000}\selectfont\catcode`\^=\active\def^{\ifmmode\sp\else\^{}\fi}\catcode`\%=\active\def
\end{pgfscope}%
\begin{pgfscope}%
\pgfpathrectangle{\pgfqpoint{0.249462in}{0.197377in}}{\pgfqpoint{2.325000in}{1.540000in}}%
\pgfusepath{clip}%
\pgfsetrectcap%
\pgfsetroundjoin%
\pgfsetlinewidth{0.501875pt}%
\definecolor{currentstroke}{rgb}{0.690196,0.690196,0.690196}%
\pgfsetstrokecolor{currentstroke}%
\pgfsetstrokeopacity{0.500000}%
\pgfsetdash{}{0pt}%
\pgfpathmoveto{\pgfqpoint{0.249462in}{1.010662in}}%
\pgfpathlineto{\pgfqpoint{2.574462in}{1.010662in}}%
\pgfusepath{stroke}%
\end{pgfscope}%
\begin{pgfscope}%
\pgfsetbuttcap%
\pgfsetroundjoin%
\definecolor{currentfill}{rgb}{0.000000,0.000000,0.000000}%
\pgfsetfillcolor{currentfill}%
\pgfsetlinewidth{0.501875pt}%
\definecolor{currentstroke}{rgb}{0.000000,0.000000,0.000000}%
\pgfsetstrokecolor{currentstroke}%
\pgfsetdash{}{0pt}%
\pgfsys@defobject{currentmarker}{\pgfqpoint{0.000000in}{0.000000in}}{\pgfqpoint{0.041667in}{0.000000in}}{%
\pgfpathmoveto{\pgfqpoint{0.000000in}{0.000000in}}%
\pgfpathlineto{\pgfqpoint{0.041667in}{0.000000in}}%
\pgfusepath{stroke,fill}%
}%
\begin{pgfscope}%
\pgfsys@transformshift{0.249462in}{1.010662in}%
\pgfsys@useobject{currentmarker}{}%
\end{pgfscope}%
\end{pgfscope}%
\begin{pgfscope}%
\definecolor{textcolor}{rgb}{0.000000,0.000000,0.000000}%
\pgfsetstrokecolor{textcolor}%
\pgfsetfillcolor{textcolor}%
\pgftext[x=0.050000in, y=0.972082in, left, base]{\color{textcolor}{\rmfamily\fontsize{8.000000}{9.600000}\selectfont\catcode`\^=\active\def^{\ifmmode\sp\else\^{}\fi}\catcode`\%=\active\def
\end{pgfscope}%
\begin{pgfscope}%
\pgfpathrectangle{\pgfqpoint{0.249462in}{0.197377in}}{\pgfqpoint{2.325000in}{1.540000in}}%
\pgfusepath{clip}%
\pgfsetrectcap%
\pgfsetroundjoin%
\pgfsetlinewidth{0.501875pt}%
\definecolor{currentstroke}{rgb}{0.690196,0.690196,0.690196}%
\pgfsetstrokecolor{currentstroke}%
\pgfsetstrokeopacity{0.500000}%
\pgfsetdash{}{0pt}%
\pgfpathmoveto{\pgfqpoint{0.249462in}{1.281757in}}%
\pgfpathlineto{\pgfqpoint{2.574462in}{1.281757in}}%
\pgfusepath{stroke}%
\end{pgfscope}%
\begin{pgfscope}%
\pgfsetbuttcap%
\pgfsetroundjoin%
\definecolor{currentfill}{rgb}{0.000000,0.000000,0.000000}%
\pgfsetfillcolor{currentfill}%
\pgfsetlinewidth{0.501875pt}%
\definecolor{currentstroke}{rgb}{0.000000,0.000000,0.000000}%
\pgfsetstrokecolor{currentstroke}%
\pgfsetdash{}{0pt}%
\pgfsys@defobject{currentmarker}{\pgfqpoint{0.000000in}{0.000000in}}{\pgfqpoint{0.041667in}{0.000000in}}{%
\pgfpathmoveto{\pgfqpoint{0.000000in}{0.000000in}}%
\pgfpathlineto{\pgfqpoint{0.041667in}{0.000000in}}%
\pgfusepath{stroke,fill}%
}%
\begin{pgfscope}%
\pgfsys@transformshift{0.249462in}{1.281757in}%
\pgfsys@useobject{currentmarker}{}%
\end{pgfscope}%
\end{pgfscope}%
\begin{pgfscope}%
\definecolor{textcolor}{rgb}{0.000000,0.000000,0.000000}%
\pgfsetstrokecolor{textcolor}%
\pgfsetfillcolor{textcolor}%
\pgftext[x=0.050000in, y=1.243177in, left, base]{\color{textcolor}{\rmfamily\fontsize{8.000000}{9.600000}\selectfont\catcode`\^=\active\def^{\ifmmode\sp\else\^{}\fi}\catcode`\%=\active\def
\end{pgfscope}%
\begin{pgfscope}%
\pgfpathrectangle{\pgfqpoint{0.249462in}{0.197377in}}{\pgfqpoint{2.325000in}{1.540000in}}%
\pgfusepath{clip}%
\pgfsetrectcap%
\pgfsetroundjoin%
\pgfsetlinewidth{0.501875pt}%
\definecolor{currentstroke}{rgb}{0.690196,0.690196,0.690196}%
\pgfsetstrokecolor{currentstroke}%
\pgfsetstrokeopacity{0.500000}%
\pgfsetdash{}{0pt}%
\pgfpathmoveto{\pgfqpoint{0.249462in}{1.552852in}}%
\pgfpathlineto{\pgfqpoint{2.574462in}{1.552852in}}%
\pgfusepath{stroke}%
\end{pgfscope}%
\begin{pgfscope}%
\pgfsetbuttcap%
\pgfsetroundjoin%
\definecolor{currentfill}{rgb}{0.000000,0.000000,0.000000}%
\pgfsetfillcolor{currentfill}%
\pgfsetlinewidth{0.501875pt}%
\definecolor{currentstroke}{rgb}{0.000000,0.000000,0.000000}%
\pgfsetstrokecolor{currentstroke}%
\pgfsetdash{}{0pt}%
\pgfsys@defobject{currentmarker}{\pgfqpoint{0.000000in}{0.000000in}}{\pgfqpoint{0.041667in}{0.000000in}}{%
\pgfpathmoveto{\pgfqpoint{0.000000in}{0.000000in}}%
\pgfpathlineto{\pgfqpoint{0.041667in}{0.000000in}}%
\pgfusepath{stroke,fill}%
}%
\begin{pgfscope}%
\pgfsys@transformshift{0.249462in}{1.552852in}%
\pgfsys@useobject{currentmarker}{}%
\end{pgfscope}%
\end{pgfscope}%
\begin{pgfscope}%
\definecolor{textcolor}{rgb}{0.000000,0.000000,0.000000}%
\pgfsetstrokecolor{textcolor}%
\pgfsetfillcolor{textcolor}%
\pgftext[x=0.050000in, y=1.514272in, left, base]{\color{textcolor}{\rmfamily\fontsize{8.000000}{9.600000}\selectfont\catcode`\^=\active\def^{\ifmmode\sp\else\^{}\fi}\catcode`\%=\active\def
\end{pgfscope}%
\begin{pgfscope}%
\pgfsetbuttcap%
\pgfsetroundjoin%
\definecolor{currentfill}{rgb}{0.000000,0.000000,0.000000}%
\pgfsetfillcolor{currentfill}%
\pgfsetlinewidth{0.501875pt}%
\definecolor{currentstroke}{rgb}{0.000000,0.000000,0.000000}%
\pgfsetstrokecolor{currentstroke}%
\pgfsetdash{}{0pt}%
\pgfsys@defobject{currentmarker}{\pgfqpoint{0.000000in}{0.000000in}}{\pgfqpoint{0.020833in}{0.000000in}}{%
\pgfpathmoveto{\pgfqpoint{0.000000in}{0.000000in}}%
\pgfpathlineto{\pgfqpoint{0.020833in}{0.000000in}}%
\pgfusepath{stroke,fill}%
}%
\begin{pgfscope}%
\pgfsys@transformshift{0.249462in}{0.251596in}%
\pgfsys@useobject{currentmarker}{}%
\end{pgfscope}%
\end{pgfscope}%
\begin{pgfscope}%
\pgfsetbuttcap%
\pgfsetroundjoin%
\definecolor{currentfill}{rgb}{0.000000,0.000000,0.000000}%
\pgfsetfillcolor{currentfill}%
\pgfsetlinewidth{0.501875pt}%
\definecolor{currentstroke}{rgb}{0.000000,0.000000,0.000000}%
\pgfsetstrokecolor{currentstroke}%
\pgfsetdash{}{0pt}%
\pgfsys@defobject{currentmarker}{\pgfqpoint{0.000000in}{0.000000in}}{\pgfqpoint{0.020833in}{0.000000in}}{%
\pgfpathmoveto{\pgfqpoint{0.000000in}{0.000000in}}%
\pgfpathlineto{\pgfqpoint{0.020833in}{0.000000in}}%
\pgfusepath{stroke,fill}%
}%
\begin{pgfscope}%
\pgfsys@transformshift{0.249462in}{0.305815in}%
\pgfsys@useobject{currentmarker}{}%
\end{pgfscope}%
\end{pgfscope}%
\begin{pgfscope}%
\pgfsetbuttcap%
\pgfsetroundjoin%
\definecolor{currentfill}{rgb}{0.000000,0.000000,0.000000}%
\pgfsetfillcolor{currentfill}%
\pgfsetlinewidth{0.501875pt}%
\definecolor{currentstroke}{rgb}{0.000000,0.000000,0.000000}%
\pgfsetstrokecolor{currentstroke}%
\pgfsetdash{}{0pt}%
\pgfsys@defobject{currentmarker}{\pgfqpoint{0.000000in}{0.000000in}}{\pgfqpoint{0.020833in}{0.000000in}}{%
\pgfpathmoveto{\pgfqpoint{0.000000in}{0.000000in}}%
\pgfpathlineto{\pgfqpoint{0.020833in}{0.000000in}}%
\pgfusepath{stroke,fill}%
}%
\begin{pgfscope}%
\pgfsys@transformshift{0.249462in}{0.360034in}%
\pgfsys@useobject{currentmarker}{}%
\end{pgfscope}%
\end{pgfscope}%
\begin{pgfscope}%
\pgfsetbuttcap%
\pgfsetroundjoin%
\definecolor{currentfill}{rgb}{0.000000,0.000000,0.000000}%
\pgfsetfillcolor{currentfill}%
\pgfsetlinewidth{0.501875pt}%
\definecolor{currentstroke}{rgb}{0.000000,0.000000,0.000000}%
\pgfsetstrokecolor{currentstroke}%
\pgfsetdash{}{0pt}%
\pgfsys@defobject{currentmarker}{\pgfqpoint{0.000000in}{0.000000in}}{\pgfqpoint{0.020833in}{0.000000in}}{%
\pgfpathmoveto{\pgfqpoint{0.000000in}{0.000000in}}%
\pgfpathlineto{\pgfqpoint{0.020833in}{0.000000in}}%
\pgfusepath{stroke,fill}%
}%
\begin{pgfscope}%
\pgfsys@transformshift{0.249462in}{0.414253in}%
\pgfsys@useobject{currentmarker}{}%
\end{pgfscope}%
\end{pgfscope}%
\begin{pgfscope}%
\pgfsetbuttcap%
\pgfsetroundjoin%
\definecolor{currentfill}{rgb}{0.000000,0.000000,0.000000}%
\pgfsetfillcolor{currentfill}%
\pgfsetlinewidth{0.501875pt}%
\definecolor{currentstroke}{rgb}{0.000000,0.000000,0.000000}%
\pgfsetstrokecolor{currentstroke}%
\pgfsetdash{}{0pt}%
\pgfsys@defobject{currentmarker}{\pgfqpoint{0.000000in}{0.000000in}}{\pgfqpoint{0.020833in}{0.000000in}}{%
\pgfpathmoveto{\pgfqpoint{0.000000in}{0.000000in}}%
\pgfpathlineto{\pgfqpoint{0.020833in}{0.000000in}}%
\pgfusepath{stroke,fill}%
}%
\begin{pgfscope}%
\pgfsys@transformshift{0.249462in}{0.522691in}%
\pgfsys@useobject{currentmarker}{}%
\end{pgfscope}%
\end{pgfscope}%
\begin{pgfscope}%
\pgfsetbuttcap%
\pgfsetroundjoin%
\definecolor{currentfill}{rgb}{0.000000,0.000000,0.000000}%
\pgfsetfillcolor{currentfill}%
\pgfsetlinewidth{0.501875pt}%
\definecolor{currentstroke}{rgb}{0.000000,0.000000,0.000000}%
\pgfsetstrokecolor{currentstroke}%
\pgfsetdash{}{0pt}%
\pgfsys@defobject{currentmarker}{\pgfqpoint{0.000000in}{0.000000in}}{\pgfqpoint{0.020833in}{0.000000in}}{%
\pgfpathmoveto{\pgfqpoint{0.000000in}{0.000000in}}%
\pgfpathlineto{\pgfqpoint{0.020833in}{0.000000in}}%
\pgfusepath{stroke,fill}%
}%
\begin{pgfscope}%
\pgfsys@transformshift{0.249462in}{0.576910in}%
\pgfsys@useobject{currentmarker}{}%
\end{pgfscope}%
\end{pgfscope}%
\begin{pgfscope}%
\pgfsetbuttcap%
\pgfsetroundjoin%
\definecolor{currentfill}{rgb}{0.000000,0.000000,0.000000}%
\pgfsetfillcolor{currentfill}%
\pgfsetlinewidth{0.501875pt}%
\definecolor{currentstroke}{rgb}{0.000000,0.000000,0.000000}%
\pgfsetstrokecolor{currentstroke}%
\pgfsetdash{}{0pt}%
\pgfsys@defobject{currentmarker}{\pgfqpoint{0.000000in}{0.000000in}}{\pgfqpoint{0.020833in}{0.000000in}}{%
\pgfpathmoveto{\pgfqpoint{0.000000in}{0.000000in}}%
\pgfpathlineto{\pgfqpoint{0.020833in}{0.000000in}}%
\pgfusepath{stroke,fill}%
}%
\begin{pgfscope}%
\pgfsys@transformshift{0.249462in}{0.631129in}%
\pgfsys@useobject{currentmarker}{}%
\end{pgfscope}%
\end{pgfscope}%
\begin{pgfscope}%
\pgfsetbuttcap%
\pgfsetroundjoin%
\definecolor{currentfill}{rgb}{0.000000,0.000000,0.000000}%
\pgfsetfillcolor{currentfill}%
\pgfsetlinewidth{0.501875pt}%
\definecolor{currentstroke}{rgb}{0.000000,0.000000,0.000000}%
\pgfsetstrokecolor{currentstroke}%
\pgfsetdash{}{0pt}%
\pgfsys@defobject{currentmarker}{\pgfqpoint{0.000000in}{0.000000in}}{\pgfqpoint{0.020833in}{0.000000in}}{%
\pgfpathmoveto{\pgfqpoint{0.000000in}{0.000000in}}%
\pgfpathlineto{\pgfqpoint{0.020833in}{0.000000in}}%
\pgfusepath{stroke,fill}%
}%
\begin{pgfscope}%
\pgfsys@transformshift{0.249462in}{0.685348in}%
\pgfsys@useobject{currentmarker}{}%
\end{pgfscope}%
\end{pgfscope}%
\begin{pgfscope}%
\pgfsetbuttcap%
\pgfsetroundjoin%
\definecolor{currentfill}{rgb}{0.000000,0.000000,0.000000}%
\pgfsetfillcolor{currentfill}%
\pgfsetlinewidth{0.501875pt}%
\definecolor{currentstroke}{rgb}{0.000000,0.000000,0.000000}%
\pgfsetstrokecolor{currentstroke}%
\pgfsetdash{}{0pt}%
\pgfsys@defobject{currentmarker}{\pgfqpoint{0.000000in}{0.000000in}}{\pgfqpoint{0.020833in}{0.000000in}}{%
\pgfpathmoveto{\pgfqpoint{0.000000in}{0.000000in}}%
\pgfpathlineto{\pgfqpoint{0.020833in}{0.000000in}}%
\pgfusepath{stroke,fill}%
}%
\begin{pgfscope}%
\pgfsys@transformshift{0.249462in}{0.793786in}%
\pgfsys@useobject{currentmarker}{}%
\end{pgfscope}%
\end{pgfscope}%
\begin{pgfscope}%
\pgfsetbuttcap%
\pgfsetroundjoin%
\definecolor{currentfill}{rgb}{0.000000,0.000000,0.000000}%
\pgfsetfillcolor{currentfill}%
\pgfsetlinewidth{0.501875pt}%
\definecolor{currentstroke}{rgb}{0.000000,0.000000,0.000000}%
\pgfsetstrokecolor{currentstroke}%
\pgfsetdash{}{0pt}%
\pgfsys@defobject{currentmarker}{\pgfqpoint{0.000000in}{0.000000in}}{\pgfqpoint{0.020833in}{0.000000in}}{%
\pgfpathmoveto{\pgfqpoint{0.000000in}{0.000000in}}%
\pgfpathlineto{\pgfqpoint{0.020833in}{0.000000in}}%
\pgfusepath{stroke,fill}%
}%
\begin{pgfscope}%
\pgfsys@transformshift{0.249462in}{0.848005in}%
\pgfsys@useobject{currentmarker}{}%
\end{pgfscope}%
\end{pgfscope}%
\begin{pgfscope}%
\pgfsetbuttcap%
\pgfsetroundjoin%
\definecolor{currentfill}{rgb}{0.000000,0.000000,0.000000}%
\pgfsetfillcolor{currentfill}%
\pgfsetlinewidth{0.501875pt}%
\definecolor{currentstroke}{rgb}{0.000000,0.000000,0.000000}%
\pgfsetstrokecolor{currentstroke}%
\pgfsetdash{}{0pt}%
\pgfsys@defobject{currentmarker}{\pgfqpoint{0.000000in}{0.000000in}}{\pgfqpoint{0.020833in}{0.000000in}}{%
\pgfpathmoveto{\pgfqpoint{0.000000in}{0.000000in}}%
\pgfpathlineto{\pgfqpoint{0.020833in}{0.000000in}}%
\pgfusepath{stroke,fill}%
}%
\begin{pgfscope}%
\pgfsys@transformshift{0.249462in}{0.902224in}%
\pgfsys@useobject{currentmarker}{}%
\end{pgfscope}%
\end{pgfscope}%
\begin{pgfscope}%
\pgfsetbuttcap%
\pgfsetroundjoin%
\definecolor{currentfill}{rgb}{0.000000,0.000000,0.000000}%
\pgfsetfillcolor{currentfill}%
\pgfsetlinewidth{0.501875pt}%
\definecolor{currentstroke}{rgb}{0.000000,0.000000,0.000000}%
\pgfsetstrokecolor{currentstroke}%
\pgfsetdash{}{0pt}%
\pgfsys@defobject{currentmarker}{\pgfqpoint{0.000000in}{0.000000in}}{\pgfqpoint{0.020833in}{0.000000in}}{%
\pgfpathmoveto{\pgfqpoint{0.000000in}{0.000000in}}%
\pgfpathlineto{\pgfqpoint{0.020833in}{0.000000in}}%
\pgfusepath{stroke,fill}%
}%
\begin{pgfscope}%
\pgfsys@transformshift{0.249462in}{0.956443in}%
\pgfsys@useobject{currentmarker}{}%
\end{pgfscope}%
\end{pgfscope}%
\begin{pgfscope}%
\pgfsetbuttcap%
\pgfsetroundjoin%
\definecolor{currentfill}{rgb}{0.000000,0.000000,0.000000}%
\pgfsetfillcolor{currentfill}%
\pgfsetlinewidth{0.501875pt}%
\definecolor{currentstroke}{rgb}{0.000000,0.000000,0.000000}%
\pgfsetstrokecolor{currentstroke}%
\pgfsetdash{}{0pt}%
\pgfsys@defobject{currentmarker}{\pgfqpoint{0.000000in}{0.000000in}}{\pgfqpoint{0.020833in}{0.000000in}}{%
\pgfpathmoveto{\pgfqpoint{0.000000in}{0.000000in}}%
\pgfpathlineto{\pgfqpoint{0.020833in}{0.000000in}}%
\pgfusepath{stroke,fill}%
}%
\begin{pgfscope}%
\pgfsys@transformshift{0.249462in}{1.064881in}%
\pgfsys@useobject{currentmarker}{}%
\end{pgfscope}%
\end{pgfscope}%
\begin{pgfscope}%
\pgfsetbuttcap%
\pgfsetroundjoin%
\definecolor{currentfill}{rgb}{0.000000,0.000000,0.000000}%
\pgfsetfillcolor{currentfill}%
\pgfsetlinewidth{0.501875pt}%
\definecolor{currentstroke}{rgb}{0.000000,0.000000,0.000000}%
\pgfsetstrokecolor{currentstroke}%
\pgfsetdash{}{0pt}%
\pgfsys@defobject{currentmarker}{\pgfqpoint{0.000000in}{0.000000in}}{\pgfqpoint{0.020833in}{0.000000in}}{%
\pgfpathmoveto{\pgfqpoint{0.000000in}{0.000000in}}%
\pgfpathlineto{\pgfqpoint{0.020833in}{0.000000in}}%
\pgfusepath{stroke,fill}%
}%
\begin{pgfscope}%
\pgfsys@transformshift{0.249462in}{1.119100in}%
\pgfsys@useobject{currentmarker}{}%
\end{pgfscope}%
\end{pgfscope}%
\begin{pgfscope}%
\pgfsetbuttcap%
\pgfsetroundjoin%
\definecolor{currentfill}{rgb}{0.000000,0.000000,0.000000}%
\pgfsetfillcolor{currentfill}%
\pgfsetlinewidth{0.501875pt}%
\definecolor{currentstroke}{rgb}{0.000000,0.000000,0.000000}%
\pgfsetstrokecolor{currentstroke}%
\pgfsetdash{}{0pt}%
\pgfsys@defobject{currentmarker}{\pgfqpoint{0.000000in}{0.000000in}}{\pgfqpoint{0.020833in}{0.000000in}}{%
\pgfpathmoveto{\pgfqpoint{0.000000in}{0.000000in}}%
\pgfpathlineto{\pgfqpoint{0.020833in}{0.000000in}}%
\pgfusepath{stroke,fill}%
}%
\begin{pgfscope}%
\pgfsys@transformshift{0.249462in}{1.173319in}%
\pgfsys@useobject{currentmarker}{}%
\end{pgfscope}%
\end{pgfscope}%
\begin{pgfscope}%
\pgfsetbuttcap%
\pgfsetroundjoin%
\definecolor{currentfill}{rgb}{0.000000,0.000000,0.000000}%
\pgfsetfillcolor{currentfill}%
\pgfsetlinewidth{0.501875pt}%
\definecolor{currentstroke}{rgb}{0.000000,0.000000,0.000000}%
\pgfsetstrokecolor{currentstroke}%
\pgfsetdash{}{0pt}%
\pgfsys@defobject{currentmarker}{\pgfqpoint{0.000000in}{0.000000in}}{\pgfqpoint{0.020833in}{0.000000in}}{%
\pgfpathmoveto{\pgfqpoint{0.000000in}{0.000000in}}%
\pgfpathlineto{\pgfqpoint{0.020833in}{0.000000in}}%
\pgfusepath{stroke,fill}%
}%
\begin{pgfscope}%
\pgfsys@transformshift{0.249462in}{1.227538in}%
\pgfsys@useobject{currentmarker}{}%
\end{pgfscope}%
\end{pgfscope}%
\begin{pgfscope}%
\pgfsetbuttcap%
\pgfsetroundjoin%
\definecolor{currentfill}{rgb}{0.000000,0.000000,0.000000}%
\pgfsetfillcolor{currentfill}%
\pgfsetlinewidth{0.501875pt}%
\definecolor{currentstroke}{rgb}{0.000000,0.000000,0.000000}%
\pgfsetstrokecolor{currentstroke}%
\pgfsetdash{}{0pt}%
\pgfsys@defobject{currentmarker}{\pgfqpoint{0.000000in}{0.000000in}}{\pgfqpoint{0.020833in}{0.000000in}}{%
\pgfpathmoveto{\pgfqpoint{0.000000in}{0.000000in}}%
\pgfpathlineto{\pgfqpoint{0.020833in}{0.000000in}}%
\pgfusepath{stroke,fill}%
}%
\begin{pgfscope}%
\pgfsys@transformshift{0.249462in}{1.335976in}%
\pgfsys@useobject{currentmarker}{}%
\end{pgfscope}%
\end{pgfscope}%
\begin{pgfscope}%
\pgfsetbuttcap%
\pgfsetroundjoin%
\definecolor{currentfill}{rgb}{0.000000,0.000000,0.000000}%
\pgfsetfillcolor{currentfill}%
\pgfsetlinewidth{0.501875pt}%
\definecolor{currentstroke}{rgb}{0.000000,0.000000,0.000000}%
\pgfsetstrokecolor{currentstroke}%
\pgfsetdash{}{0pt}%
\pgfsys@defobject{currentmarker}{\pgfqpoint{0.000000in}{0.000000in}}{\pgfqpoint{0.020833in}{0.000000in}}{%
\pgfpathmoveto{\pgfqpoint{0.000000in}{0.000000in}}%
\pgfpathlineto{\pgfqpoint{0.020833in}{0.000000in}}%
\pgfusepath{stroke,fill}%
}%
\begin{pgfscope}%
\pgfsys@transformshift{0.249462in}{1.390195in}%
\pgfsys@useobject{currentmarker}{}%
\end{pgfscope}%
\end{pgfscope}%
\begin{pgfscope}%
\pgfsetbuttcap%
\pgfsetroundjoin%
\definecolor{currentfill}{rgb}{0.000000,0.000000,0.000000}%
\pgfsetfillcolor{currentfill}%
\pgfsetlinewidth{0.501875pt}%
\definecolor{currentstroke}{rgb}{0.000000,0.000000,0.000000}%
\pgfsetstrokecolor{currentstroke}%
\pgfsetdash{}{0pt}%
\pgfsys@defobject{currentmarker}{\pgfqpoint{0.000000in}{0.000000in}}{\pgfqpoint{0.020833in}{0.000000in}}{%
\pgfpathmoveto{\pgfqpoint{0.000000in}{0.000000in}}%
\pgfpathlineto{\pgfqpoint{0.020833in}{0.000000in}}%
\pgfusepath{stroke,fill}%
}%
\begin{pgfscope}%
\pgfsys@transformshift{0.249462in}{1.444414in}%
\pgfsys@useobject{currentmarker}{}%
\end{pgfscope}%
\end{pgfscope}%
\begin{pgfscope}%
\pgfsetbuttcap%
\pgfsetroundjoin%
\definecolor{currentfill}{rgb}{0.000000,0.000000,0.000000}%
\pgfsetfillcolor{currentfill}%
\pgfsetlinewidth{0.501875pt}%
\definecolor{currentstroke}{rgb}{0.000000,0.000000,0.000000}%
\pgfsetstrokecolor{currentstroke}%
\pgfsetdash{}{0pt}%
\pgfsys@defobject{currentmarker}{\pgfqpoint{0.000000in}{0.000000in}}{\pgfqpoint{0.020833in}{0.000000in}}{%
\pgfpathmoveto{\pgfqpoint{0.000000in}{0.000000in}}%
\pgfpathlineto{\pgfqpoint{0.020833in}{0.000000in}}%
\pgfusepath{stroke,fill}%
}%
\begin{pgfscope}%
\pgfsys@transformshift{0.249462in}{1.498633in}%
\pgfsys@useobject{currentmarker}{}%
\end{pgfscope}%
\end{pgfscope}%
\begin{pgfscope}%
\pgfsetbuttcap%
\pgfsetroundjoin%
\definecolor{currentfill}{rgb}{0.000000,0.000000,0.000000}%
\pgfsetfillcolor{currentfill}%
\pgfsetlinewidth{0.501875pt}%
\definecolor{currentstroke}{rgb}{0.000000,0.000000,0.000000}%
\pgfsetstrokecolor{currentstroke}%
\pgfsetdash{}{0pt}%
\pgfsys@defobject{currentmarker}{\pgfqpoint{0.000000in}{0.000000in}}{\pgfqpoint{0.020833in}{0.000000in}}{%
\pgfpathmoveto{\pgfqpoint{0.000000in}{0.000000in}}%
\pgfpathlineto{\pgfqpoint{0.020833in}{0.000000in}}%
\pgfusepath{stroke,fill}%
}%
\begin{pgfscope}%
\pgfsys@transformshift{0.249462in}{1.607071in}%
\pgfsys@useobject{currentmarker}{}%
\end{pgfscope}%
\end{pgfscope}%
\begin{pgfscope}%
\pgfsetbuttcap%
\pgfsetroundjoin%
\definecolor{currentfill}{rgb}{0.000000,0.000000,0.000000}%
\pgfsetfillcolor{currentfill}%
\pgfsetlinewidth{0.501875pt}%
\definecolor{currentstroke}{rgb}{0.000000,0.000000,0.000000}%
\pgfsetstrokecolor{currentstroke}%
\pgfsetdash{}{0pt}%
\pgfsys@defobject{currentmarker}{\pgfqpoint{0.000000in}{0.000000in}}{\pgfqpoint{0.020833in}{0.000000in}}{%
\pgfpathmoveto{\pgfqpoint{0.000000in}{0.000000in}}%
\pgfpathlineto{\pgfqpoint{0.020833in}{0.000000in}}%
\pgfusepath{stroke,fill}%
}%
\begin{pgfscope}%
\pgfsys@transformshift{0.249462in}{1.661290in}%
\pgfsys@useobject{currentmarker}{}%
\end{pgfscope}%
\end{pgfscope}%
\begin{pgfscope}%
\pgfsetbuttcap%
\pgfsetroundjoin%
\definecolor{currentfill}{rgb}{0.000000,0.000000,0.000000}%
\pgfsetfillcolor{currentfill}%
\pgfsetlinewidth{0.501875pt}%
\definecolor{currentstroke}{rgb}{0.000000,0.000000,0.000000}%
\pgfsetstrokecolor{currentstroke}%
\pgfsetdash{}{0pt}%
\pgfsys@defobject{currentmarker}{\pgfqpoint{0.000000in}{0.000000in}}{\pgfqpoint{0.020833in}{0.000000in}}{%
\pgfpathmoveto{\pgfqpoint{0.000000in}{0.000000in}}%
\pgfpathlineto{\pgfqpoint{0.020833in}{0.000000in}}%
\pgfusepath{stroke,fill}%
}%
\begin{pgfscope}%
\pgfsys@transformshift{0.249462in}{1.715509in}%
\pgfsys@useobject{currentmarker}{}%
\end{pgfscope}%
\end{pgfscope}%
\begin{pgfscope}%
\pgfpathrectangle{\pgfqpoint{0.249462in}{0.197377in}}{\pgfqpoint{2.325000in}{1.540000in}}%
\pgfusepath{clip}%
\pgfsetbuttcap%
\pgfsetroundjoin%
\pgfsetlinewidth{0.501875pt}%
\definecolor{currentstroke}{rgb}{0.501961,0.501961,0.501961}%
\pgfsetstrokecolor{currentstroke}%
\pgfsetstrokeopacity{0.800000}%
\pgfsetdash{{1.850000pt}{0.800000pt}}{0.000000pt}%
\pgfpathmoveto{\pgfqpoint{0.832140in}{0.197377in}}%
\pgfpathlineto{\pgfqpoint{0.832140in}{1.737377in}}%
\pgfusepath{stroke}%
\end{pgfscope}%
\begin{pgfscope}%
\pgfpathrectangle{\pgfqpoint{0.249462in}{0.197377in}}{\pgfqpoint{2.325000in}{1.540000in}}%
\pgfusepath{clip}%
\pgfsetbuttcap%
\pgfsetroundjoin%
\pgfsetlinewidth{0.501875pt}%
\definecolor{currentstroke}{rgb}{0.501961,0.501961,0.501961}%
\pgfsetstrokecolor{currentstroke}%
\pgfsetstrokeopacity{0.800000}%
\pgfsetdash{{1.850000pt}{0.800000pt}}{0.000000pt}%
\pgfpathmoveto{\pgfqpoint{1.403393in}{0.197377in}}%
\pgfpathlineto{\pgfqpoint{1.403393in}{1.737377in}}%
\pgfusepath{stroke}%
\end{pgfscope}%
\begin{pgfscope}%
\pgfpathrectangle{\pgfqpoint{0.249462in}{0.197377in}}{\pgfqpoint{2.325000in}{1.540000in}}%
\pgfusepath{clip}%
\pgfsetbuttcap%
\pgfsetroundjoin%
\pgfsetlinewidth{0.501875pt}%
\definecolor{currentstroke}{rgb}{0.501961,0.501961,0.501961}%
\pgfsetstrokecolor{currentstroke}%
\pgfsetstrokeopacity{0.800000}%
\pgfsetdash{{1.850000pt}{0.800000pt}}{0.000000pt}%
\pgfpathmoveto{\pgfqpoint{1.974646in}{0.197377in}}%
\pgfpathlineto{\pgfqpoint{1.974646in}{1.737377in}}%
\pgfusepath{stroke}%
\end{pgfscope}%
\begin{pgfscope}%
\pgfsetrectcap%
\pgfsetmiterjoin%
\pgfsetlinewidth{0.501875pt}%
\definecolor{currentstroke}{rgb}{0.000000,0.000000,0.000000}%
\pgfsetstrokecolor{currentstroke}%
\pgfsetdash{}{0pt}%
\pgfpathmoveto{\pgfqpoint{0.249462in}{0.197377in}}%
\pgfpathlineto{\pgfqpoint{0.249462in}{1.737377in}}%
\pgfusepath{stroke}%
\end{pgfscope}%
\begin{pgfscope}%
\pgfsetrectcap%
\pgfsetmiterjoin%
\pgfsetlinewidth{0.501875pt}%
\definecolor{currentstroke}{rgb}{0.000000,0.000000,0.000000}%
\pgfsetstrokecolor{currentstroke}%
\pgfsetdash{}{0pt}%
\pgfpathmoveto{\pgfqpoint{2.574462in}{0.197377in}}%
\pgfpathlineto{\pgfqpoint{2.574462in}{1.737377in}}%
\pgfusepath{stroke}%
\end{pgfscope}%
\begin{pgfscope}%
\pgfsetrectcap%
\pgfsetmiterjoin%
\pgfsetlinewidth{0.501875pt}%
\definecolor{currentstroke}{rgb}{0.000000,0.000000,0.000000}%
\pgfsetstrokecolor{currentstroke}%
\pgfsetdash{}{0pt}%
\pgfpathmoveto{\pgfqpoint{0.249462in}{0.197377in}}%
\pgfpathlineto{\pgfqpoint{2.574462in}{0.197377in}}%
\pgfusepath{stroke}%
\end{pgfscope}%
\begin{pgfscope}%
\pgfsetrectcap%
\pgfsetmiterjoin%
\pgfsetlinewidth{0.501875pt}%
\definecolor{currentstroke}{rgb}{0.000000,0.000000,0.000000}%
\pgfsetstrokecolor{currentstroke}%
\pgfsetdash{}{0pt}%
\pgfpathmoveto{\pgfqpoint{0.249462in}{1.737377in}}%
\pgfpathlineto{\pgfqpoint{2.574462in}{1.737377in}}%
\pgfusepath{stroke}%
\end{pgfscope}%
\begin{pgfscope}%
\pgfpathrectangle{\pgfqpoint{0.249462in}{0.197377in}}{\pgfqpoint{2.325000in}{1.540000in}}%
\pgfusepath{clip}%
\pgfsetbuttcap%
\pgfsetmiterjoin%
\definecolor{currentfill}{rgb}{0.000000,0.000000,0.000000}%
\pgfsetfillcolor{currentfill}%
\pgfsetlinewidth{0.000000pt}%
\definecolor{currentstroke}{rgb}{0.000000,0.000000,0.000000}%
\pgfsetstrokecolor{currentstroke}%
\pgfsetstrokeopacity{0.000000}%
\pgfsetdash{}{0pt}%
\pgfpathmoveto{\pgfqpoint{0.355144in}{0.197377in}}%
\pgfpathlineto{\pgfqpoint{0.440832in}{0.197377in}}%
\pgfpathlineto{\pgfqpoint{0.440832in}{1.664043in}}%
\pgfpathlineto{\pgfqpoint{0.355144in}{1.664043in}}%
\pgfpathlineto{\pgfqpoint{0.355144in}{0.197377in}}%
\pgfpathclose%
\pgfusepath{fill}%
\end{pgfscope}%
\begin{pgfscope}%
\pgfpathrectangle{\pgfqpoint{0.249462in}{0.197377in}}{\pgfqpoint{2.325000in}{1.540000in}}%
\pgfusepath{clip}%
\pgfsetbuttcap%
\pgfsetmiterjoin%
\definecolor{currentfill}{rgb}{0.000000,0.000000,0.000000}%
\pgfsetfillcolor{currentfill}%
\pgfsetlinewidth{0.000000pt}%
\definecolor{currentstroke}{rgb}{0.000000,0.000000,0.000000}%
\pgfsetstrokecolor{currentstroke}%
\pgfsetstrokeopacity{0.000000}%
\pgfsetdash{}{0pt}%
\pgfpathmoveto{\pgfqpoint{0.926397in}{0.197377in}}%
\pgfpathlineto{\pgfqpoint{1.012085in}{0.197377in}}%
\pgfpathlineto{\pgfqpoint{1.012085in}{0.197377in}}%
\pgfpathlineto{\pgfqpoint{0.926397in}{0.197377in}}%
\pgfpathlineto{\pgfqpoint{0.926397in}{0.197377in}}%
\pgfpathclose%
\pgfusepath{fill}%
\end{pgfscope}%
\begin{pgfscope}%
\pgfpathrectangle{\pgfqpoint{0.249462in}{0.197377in}}{\pgfqpoint{2.325000in}{1.540000in}}%
\pgfusepath{clip}%
\pgfsetbuttcap%
\pgfsetmiterjoin%
\definecolor{currentfill}{rgb}{0.000000,0.000000,0.000000}%
\pgfsetfillcolor{currentfill}%
\pgfsetlinewidth{0.000000pt}%
\definecolor{currentstroke}{rgb}{0.000000,0.000000,0.000000}%
\pgfsetstrokecolor{currentstroke}%
\pgfsetstrokeopacity{0.000000}%
\pgfsetdash{}{0pt}%
\pgfpathmoveto{\pgfqpoint{1.497650in}{0.197377in}}%
\pgfpathlineto{\pgfqpoint{1.583338in}{0.197377in}}%
\pgfpathlineto{\pgfqpoint{1.583338in}{0.799332in}}%
\pgfpathlineto{\pgfqpoint{1.497650in}{0.799332in}}%
\pgfpathlineto{\pgfqpoint{1.497650in}{0.197377in}}%
\pgfpathclose%
\pgfusepath{fill}%
\end{pgfscope}%
\begin{pgfscope}%
\pgfpathrectangle{\pgfqpoint{0.249462in}{0.197377in}}{\pgfqpoint{2.325000in}{1.540000in}}%
\pgfusepath{clip}%
\pgfsetbuttcap%
\pgfsetmiterjoin%
\definecolor{currentfill}{rgb}{0.000000,0.000000,0.000000}%
\pgfsetfillcolor{currentfill}%
\pgfsetlinewidth{0.000000pt}%
\definecolor{currentstroke}{rgb}{0.000000,0.000000,0.000000}%
\pgfsetstrokecolor{currentstroke}%
\pgfsetstrokeopacity{0.000000}%
\pgfsetdash{}{0pt}%
\pgfpathmoveto{\pgfqpoint{2.068903in}{0.197377in}}%
\pgfpathlineto{\pgfqpoint{2.154591in}{0.197377in}}%
\pgfpathlineto{\pgfqpoint{2.154591in}{1.664043in}}%
\pgfpathlineto{\pgfqpoint{2.068903in}{1.664043in}}%
\pgfpathlineto{\pgfqpoint{2.068903in}{0.197377in}}%
\pgfpathclose%
\pgfusepath{fill}%
\end{pgfscope}%
\begin{pgfscope}%
\pgfpathrectangle{\pgfqpoint{0.249462in}{0.197377in}}{\pgfqpoint{2.325000in}{1.540000in}}%
\pgfusepath{clip}%
\pgfsetbuttcap%
\pgfsetmiterjoin%
\definecolor{currentfill}{rgb}{1.000000,0.000000,0.000000}%
\pgfsetfillcolor{currentfill}%
\pgfsetlinewidth{0.000000pt}%
\definecolor{currentstroke}{rgb}{0.000000,0.000000,0.000000}%
\pgfsetstrokecolor{currentstroke}%
\pgfsetstrokeopacity{0.000000}%
\pgfsetdash{}{0pt}%
\pgfpathmoveto{\pgfqpoint{0.440832in}{0.197377in}}%
\pgfpathlineto{\pgfqpoint{0.526520in}{0.197377in}}%
\pgfpathlineto{\pgfqpoint{0.526520in}{1.642864in}}%
\pgfpathlineto{\pgfqpoint{0.440832in}{1.642864in}}%
\pgfpathlineto{\pgfqpoint{0.440832in}{0.197377in}}%
\pgfpathclose%
\pgfusepath{fill}%
\end{pgfscope}%
\begin{pgfscope}%
\pgfpathrectangle{\pgfqpoint{0.249462in}{0.197377in}}{\pgfqpoint{2.325000in}{1.540000in}}%
\pgfusepath{clip}%
\pgfsetbuttcap%
\pgfsetmiterjoin%
\definecolor{currentfill}{rgb}{1.000000,0.000000,0.000000}%
\pgfsetfillcolor{currentfill}%
\pgfsetlinewidth{0.000000pt}%
\definecolor{currentstroke}{rgb}{0.000000,0.000000,0.000000}%
\pgfsetstrokecolor{currentstroke}%
\pgfsetstrokeopacity{0.000000}%
\pgfsetdash{}{0pt}%
\pgfpathmoveto{\pgfqpoint{1.012085in}{0.197377in}}%
\pgfpathlineto{\pgfqpoint{1.097773in}{0.197377in}}%
\pgfpathlineto{\pgfqpoint{1.097773in}{0.197377in}}%
\pgfpathlineto{\pgfqpoint{1.012085in}{0.197377in}}%
\pgfpathlineto{\pgfqpoint{1.012085in}{0.197377in}}%
\pgfpathclose%
\pgfusepath{fill}%
\end{pgfscope}%
\begin{pgfscope}%
\pgfpathrectangle{\pgfqpoint{0.249462in}{0.197377in}}{\pgfqpoint{2.325000in}{1.540000in}}%
\pgfusepath{clip}%
\pgfsetbuttcap%
\pgfsetmiterjoin%
\definecolor{currentfill}{rgb}{1.000000,0.000000,0.000000}%
\pgfsetfillcolor{currentfill}%
\pgfsetlinewidth{0.000000pt}%
\definecolor{currentstroke}{rgb}{0.000000,0.000000,0.000000}%
\pgfsetstrokecolor{currentstroke}%
\pgfsetstrokeopacity{0.000000}%
\pgfsetdash{}{0pt}%
\pgfpathmoveto{\pgfqpoint{1.583338in}{0.197377in}}%
\pgfpathlineto{\pgfqpoint{1.669026in}{0.197377in}}%
\pgfpathlineto{\pgfqpoint{1.669026in}{0.793706in}}%
\pgfpathlineto{\pgfqpoint{1.583338in}{0.793706in}}%
\pgfpathlineto{\pgfqpoint{1.583338in}{0.197377in}}%
\pgfpathclose%
\pgfusepath{fill}%
\end{pgfscope}%
\begin{pgfscope}%
\pgfpathrectangle{\pgfqpoint{0.249462in}{0.197377in}}{\pgfqpoint{2.325000in}{1.540000in}}%
\pgfusepath{clip}%
\pgfsetbuttcap%
\pgfsetmiterjoin%
\definecolor{currentfill}{rgb}{1.000000,0.000000,0.000000}%
\pgfsetfillcolor{currentfill}%
\pgfsetlinewidth{0.000000pt}%
\definecolor{currentstroke}{rgb}{0.000000,0.000000,0.000000}%
\pgfsetstrokecolor{currentstroke}%
\pgfsetstrokeopacity{0.000000}%
\pgfsetdash{}{0pt}%
\pgfpathmoveto{\pgfqpoint{2.154591in}{0.197377in}}%
\pgfpathlineto{\pgfqpoint{2.240279in}{0.197377in}}%
\pgfpathlineto{\pgfqpoint{2.240279in}{1.648159in}}%
\pgfpathlineto{\pgfqpoint{2.154591in}{1.648159in}}%
\pgfpathlineto{\pgfqpoint{2.154591in}{0.197377in}}%
\pgfpathclose%
\pgfusepath{fill}%
\end{pgfscope}%
\begin{pgfscope}%
\pgfpathrectangle{\pgfqpoint{0.249462in}{0.197377in}}{\pgfqpoint{2.325000in}{1.540000in}}%
\pgfusepath{clip}%
\pgfsetbuttcap%
\pgfsetmiterjoin%
\definecolor{currentfill}{rgb}{0.000000,0.000000,1.000000}%
\pgfsetfillcolor{currentfill}%
\pgfsetlinewidth{0.000000pt}%
\definecolor{currentstroke}{rgb}{0.000000,0.000000,0.000000}%
\pgfsetstrokecolor{currentstroke}%
\pgfsetstrokeopacity{0.000000}%
\pgfsetdash{}{0pt}%
\pgfpathmoveto{\pgfqpoint{0.583645in}{0.197377in}}%
\pgfpathlineto{\pgfqpoint{0.669333in}{0.197377in}}%
\pgfpathlineto{\pgfqpoint{0.669333in}{1.574031in}}%
\pgfpathlineto{\pgfqpoint{0.583645in}{1.574031in}}%
\pgfpathlineto{\pgfqpoint{0.583645in}{0.197377in}}%
\pgfpathclose%
\pgfusepath{fill}%
\end{pgfscope}%
\begin{pgfscope}%
\pgfpathrectangle{\pgfqpoint{0.249462in}{0.197377in}}{\pgfqpoint{2.325000in}{1.540000in}}%
\pgfusepath{clip}%
\pgfsetbuttcap%
\pgfsetmiterjoin%
\definecolor{currentfill}{rgb}{0.000000,0.000000,1.000000}%
\pgfsetfillcolor{currentfill}%
\pgfsetlinewidth{0.000000pt}%
\definecolor{currentstroke}{rgb}{0.000000,0.000000,0.000000}%
\pgfsetstrokecolor{currentstroke}%
\pgfsetstrokeopacity{0.000000}%
\pgfsetdash{}{0pt}%
\pgfpathmoveto{\pgfqpoint{1.154898in}{0.197377in}}%
\pgfpathlineto{\pgfqpoint{1.240586in}{0.197377in}}%
\pgfpathlineto{\pgfqpoint{1.240586in}{0.197377in}}%
\pgfpathlineto{\pgfqpoint{1.154898in}{0.197377in}}%
\pgfpathlineto{\pgfqpoint{1.154898in}{0.197377in}}%
\pgfpathclose%
\pgfusepath{fill}%
\end{pgfscope}%
\begin{pgfscope}%
\pgfpathrectangle{\pgfqpoint{0.249462in}{0.197377in}}{\pgfqpoint{2.325000in}{1.540000in}}%
\pgfusepath{clip}%
\pgfsetbuttcap%
\pgfsetmiterjoin%
\definecolor{currentfill}{rgb}{0.000000,0.000000,1.000000}%
\pgfsetfillcolor{currentfill}%
\pgfsetlinewidth{0.000000pt}%
\definecolor{currentstroke}{rgb}{0.000000,0.000000,0.000000}%
\pgfsetstrokecolor{currentstroke}%
\pgfsetstrokeopacity{0.000000}%
\pgfsetdash{}{0pt}%
\pgfpathmoveto{\pgfqpoint{1.726151in}{0.197377in}}%
\pgfpathlineto{\pgfqpoint{1.811839in}{0.197377in}}%
\pgfpathlineto{\pgfqpoint{1.811839in}{0.766570in}}%
\pgfpathlineto{\pgfqpoint{1.726151in}{0.766570in}}%
\pgfpathlineto{\pgfqpoint{1.726151in}{0.197377in}}%
\pgfpathclose%
\pgfusepath{fill}%
\end{pgfscope}%
\begin{pgfscope}%
\pgfpathrectangle{\pgfqpoint{0.249462in}{0.197377in}}{\pgfqpoint{2.325000in}{1.540000in}}%
\pgfusepath{clip}%
\pgfsetbuttcap%
\pgfsetmiterjoin%
\definecolor{currentfill}{rgb}{0.000000,0.000000,1.000000}%
\pgfsetfillcolor{currentfill}%
\pgfsetlinewidth{0.000000pt}%
\definecolor{currentstroke}{rgb}{0.000000,0.000000,0.000000}%
\pgfsetstrokecolor{currentstroke}%
\pgfsetstrokeopacity{0.000000}%
\pgfsetdash{}{0pt}%
\pgfpathmoveto{\pgfqpoint{2.297404in}{0.197377in}}%
\pgfpathlineto{\pgfqpoint{2.383092in}{0.197377in}}%
\pgfpathlineto{\pgfqpoint{2.383092in}{1.574031in}}%
\pgfpathlineto{\pgfqpoint{2.297404in}{1.574031in}}%
\pgfpathlineto{\pgfqpoint{2.297404in}{0.197377in}}%
\pgfpathclose%
\pgfusepath{fill}%
\end{pgfscope}%
\begin{pgfscope}%
\pgfpathrectangle{\pgfqpoint{0.249462in}{0.197377in}}{\pgfqpoint{2.325000in}{1.540000in}}%
\pgfusepath{clip}%
\pgfsetbuttcap%
\pgfsetmiterjoin%
\definecolor{currentfill}{rgb}{0.000000,0.500000,0.000000}%
\pgfsetfillcolor{currentfill}%
\pgfsetlinewidth{0.000000pt}%
\definecolor{currentstroke}{rgb}{0.000000,0.000000,0.000000}%
\pgfsetstrokecolor{currentstroke}%
\pgfsetstrokeopacity{0.000000}%
\pgfsetdash{}{0pt}%
\pgfpathmoveto{\pgfqpoint{0.669333in}{0.197377in}}%
\pgfpathlineto{\pgfqpoint{0.755021in}{0.197377in}}%
\pgfpathlineto{\pgfqpoint{0.755021in}{1.501227in}}%
\pgfpathlineto{\pgfqpoint{0.669333in}{1.501227in}}%
\pgfpathlineto{\pgfqpoint{0.669333in}{0.197377in}}%
\pgfpathclose%
\pgfusepath{fill}%
\end{pgfscope}%
\begin{pgfscope}%
\pgfpathrectangle{\pgfqpoint{0.249462in}{0.197377in}}{\pgfqpoint{2.325000in}{1.540000in}}%
\pgfusepath{clip}%
\pgfsetbuttcap%
\pgfsetmiterjoin%
\definecolor{currentfill}{rgb}{0.000000,0.500000,0.000000}%
\pgfsetfillcolor{currentfill}%
\pgfsetlinewidth{0.000000pt}%
\definecolor{currentstroke}{rgb}{0.000000,0.000000,0.000000}%
\pgfsetstrokecolor{currentstroke}%
\pgfsetstrokeopacity{0.000000}%
\pgfsetdash{}{0pt}%
\pgfpathmoveto{\pgfqpoint{1.240586in}{0.197377in}}%
\pgfpathlineto{\pgfqpoint{1.326274in}{0.197377in}}%
\pgfpathlineto{\pgfqpoint{1.326274in}{0.197377in}}%
\pgfpathlineto{\pgfqpoint{1.240586in}{0.197377in}}%
\pgfpathlineto{\pgfqpoint{1.240586in}{0.197377in}}%
\pgfpathclose%
\pgfusepath{fill}%
\end{pgfscope}%
\begin{pgfscope}%
\pgfpathrectangle{\pgfqpoint{0.249462in}{0.197377in}}{\pgfqpoint{2.325000in}{1.540000in}}%
\pgfusepath{clip}%
\pgfsetbuttcap%
\pgfsetmiterjoin%
\definecolor{currentfill}{rgb}{0.000000,0.500000,0.000000}%
\pgfsetfillcolor{currentfill}%
\pgfsetlinewidth{0.000000pt}%
\definecolor{currentstroke}{rgb}{0.000000,0.000000,0.000000}%
\pgfsetstrokecolor{currentstroke}%
\pgfsetstrokeopacity{0.000000}%
\pgfsetdash{}{0pt}%
\pgfpathmoveto{\pgfqpoint{1.811839in}{0.197377in}}%
\pgfpathlineto{\pgfqpoint{1.897527in}{0.197377in}}%
\pgfpathlineto{\pgfqpoint{1.897527in}{0.707004in}}%
\pgfpathlineto{\pgfqpoint{1.811839in}{0.707004in}}%
\pgfpathlineto{\pgfqpoint{1.811839in}{0.197377in}}%
\pgfpathclose%
\pgfusepath{fill}%
\end{pgfscope}%
\begin{pgfscope}%
\pgfpathrectangle{\pgfqpoint{0.249462in}{0.197377in}}{\pgfqpoint{2.325000in}{1.540000in}}%
\pgfusepath{clip}%
\pgfsetbuttcap%
\pgfsetmiterjoin%
\definecolor{currentfill}{rgb}{0.000000,0.500000,0.000000}%
\pgfsetfillcolor{currentfill}%
\pgfsetlinewidth{0.000000pt}%
\definecolor{currentstroke}{rgb}{0.000000,0.000000,0.000000}%
\pgfsetstrokecolor{currentstroke}%
\pgfsetstrokeopacity{0.000000}%
\pgfsetdash{}{0pt}%
\pgfpathmoveto{\pgfqpoint{2.383092in}{0.197377in}}%
\pgfpathlineto{\pgfqpoint{2.468780in}{0.197377in}}%
\pgfpathlineto{\pgfqpoint{2.468780in}{1.506522in}}%
\pgfpathlineto{\pgfqpoint{2.383092in}{1.506522in}}%
\pgfpathlineto{\pgfqpoint{2.383092in}{0.197377in}}%
\pgfpathclose%
\pgfusepath{fill}%
\end{pgfscope}%
\end{pgfpicture}%
\makeatother%
\endgroup%

%% file: PIDray-object-size.pgf
\begingroup%
\makeatletter%
\begin{pgfpicture}%
\pgfpathrectangle{\pgfpointorigin}{\pgfqpoint{2.624462in}{1.787377in}}%
\pgfusepath{use as bounding box, clip}%
\begin{pgfscope}%
\pgfsetbuttcap%
\pgfsetmiterjoin%
\definecolor{currentfill}{rgb}{1.000000,1.000000,1.000000}%
\pgfsetfillcolor{currentfill}%
\pgfsetlinewidth{0.000000pt}%
\definecolor{currentstroke}{rgb}{1.000000,1.000000,1.000000}%
\pgfsetstrokecolor{currentstroke}%
\pgfsetdash{}{0pt}%
\pgfpathmoveto{\pgfqpoint{0.000000in}{0.000000in}}%
\pgfpathlineto{\pgfqpoint{2.624462in}{0.000000in}}%
\pgfpathlineto{\pgfqpoint{2.624462in}{1.787377in}}%
\pgfpathlineto{\pgfqpoint{0.000000in}{1.787377in}}%
\pgfpathlineto{\pgfqpoint{0.000000in}{0.000000in}}%
\pgfpathclose%
\pgfusepath{fill}%
\end{pgfscope}%
\begin{pgfscope}%
\pgfsetbuttcap%
\pgfsetmiterjoin%
\definecolor{currentfill}{rgb}{1.000000,1.000000,1.000000}%
\pgfsetfillcolor{currentfill}%
\pgfsetlinewidth{0.000000pt}%
\definecolor{currentstroke}{rgb}{0.000000,0.000000,0.000000}%
\pgfsetstrokecolor{currentstroke}%
\pgfsetstrokeopacity{0.000000}%
\pgfsetdash{}{0pt}%
\pgfpathmoveto{\pgfqpoint{0.249462in}{0.197377in}}%
\pgfpathlineto{\pgfqpoint{2.574462in}{0.197377in}}%
\pgfpathlineto{\pgfqpoint{2.574462in}{1.737377in}}%
\pgfpathlineto{\pgfqpoint{0.249462in}{1.737377in}}%
\pgfpathlineto{\pgfqpoint{0.249462in}{0.197377in}}%
\pgfpathclose%
\pgfusepath{fill}%
\end{pgfscope}%
\begin{pgfscope}%
\pgfpathrectangle{\pgfqpoint{0.249462in}{0.197377in}}{\pgfqpoint{2.325000in}{1.540000in}}%
\pgfusepath{clip}%
\pgfsetrectcap%
\pgfsetroundjoin%
\pgfsetlinewidth{0.501875pt}%
\definecolor{currentstroke}{rgb}{0.690196,0.690196,0.690196}%
\pgfsetstrokecolor{currentstroke}%
\pgfsetstrokeopacity{0.500000}%
\pgfsetdash{}{0pt}%
\pgfpathmoveto{\pgfqpoint{0.552226in}{0.197377in}}%
\pgfpathlineto{\pgfqpoint{0.552226in}{1.737377in}}%
\pgfusepath{stroke}%
\end{pgfscope}%
\begin{pgfscope}%
\pgfsetbuttcap%
\pgfsetroundjoin%
\definecolor{currentfill}{rgb}{0.000000,0.000000,0.000000}%
\pgfsetfillcolor{currentfill}%
\pgfsetlinewidth{0.501875pt}%
\definecolor{currentstroke}{rgb}{0.000000,0.000000,0.000000}%
\pgfsetstrokecolor{currentstroke}%
\pgfsetdash{}{0pt}%
\pgfsys@defobject{currentmarker}{\pgfqpoint{0.000000in}{0.000000in}}{\pgfqpoint{0.000000in}{0.041667in}}{%
\pgfpathmoveto{\pgfqpoint{0.000000in}{0.000000in}}%
\pgfpathlineto{\pgfqpoint{0.000000in}{0.041667in}}%
\pgfusepath{stroke,fill}%
}%
\begin{pgfscope}%
\pgfsys@transformshift{0.552226in}{0.197377in}%
\pgfsys@useobject{currentmarker}{}%
\end{pgfscope}%
\end{pgfscope}%
\begin{pgfscope}%
\definecolor{textcolor}{rgb}{0.000000,0.000000,0.000000}%
\pgfsetstrokecolor{textcolor}%
\pgfsetfillcolor{textcolor}%
\pgftext[x=0.552226in,y=0.148766in,,top]{\color{textcolor}{\rmfamily\fontsize{8.000000}{9.600000}\selectfont\catcode`\^=\active\def^{\ifmmode\sp\else\^{}\fi}\catcode`\%=\active\def
\end{pgfscope}%
\begin{pgfscope}%
\pgfpathrectangle{\pgfqpoint{0.249462in}{0.197377in}}{\pgfqpoint{2.325000in}{1.540000in}}%
\pgfusepath{clip}%
\pgfsetrectcap%
\pgfsetroundjoin%
\pgfsetlinewidth{0.501875pt}%
\definecolor{currentstroke}{rgb}{0.690196,0.690196,0.690196}%
\pgfsetstrokecolor{currentstroke}%
\pgfsetstrokeopacity{0.500000}%
\pgfsetdash{}{0pt}%
\pgfpathmoveto{\pgfqpoint{1.123479in}{0.197377in}}%
\pgfpathlineto{\pgfqpoint{1.123479in}{1.737377in}}%
\pgfusepath{stroke}%
\end{pgfscope}%
\begin{pgfscope}%
\pgfsetbuttcap%
\pgfsetroundjoin%
\definecolor{currentfill}{rgb}{0.000000,0.000000,0.000000}%
\pgfsetfillcolor{currentfill}%
\pgfsetlinewidth{0.501875pt}%
\definecolor{currentstroke}{rgb}{0.000000,0.000000,0.000000}%
\pgfsetstrokecolor{currentstroke}%
\pgfsetdash{}{0pt}%
\pgfsys@defobject{currentmarker}{\pgfqpoint{0.000000in}{0.000000in}}{\pgfqpoint{0.000000in}{0.041667in}}{%
\pgfpathmoveto{\pgfqpoint{0.000000in}{0.000000in}}%
\pgfpathlineto{\pgfqpoint{0.000000in}{0.041667in}}%
\pgfusepath{stroke,fill}%
}%
\begin{pgfscope}%
\pgfsys@transformshift{1.123479in}{0.197377in}%
\pgfsys@useobject{currentmarker}{}%
\end{pgfscope}%
\end{pgfscope}%
\begin{pgfscope}%
\definecolor{textcolor}{rgb}{0.000000,0.000000,0.000000}%
\pgfsetstrokecolor{textcolor}%
\pgfsetfillcolor{textcolor}%
\pgftext[x=1.123479in,y=0.148766in,,top]{\color{textcolor}{\rmfamily\fontsize{8.000000}{9.600000}\selectfont\catcode`\^=\active\def^{\ifmmode\sp\else\^{}\fi}\catcode`\%=\active\def
\end{pgfscope}%
\begin{pgfscope}%
\pgfpathrectangle{\pgfqpoint{0.249462in}{0.197377in}}{\pgfqpoint{2.325000in}{1.540000in}}%
\pgfusepath{clip}%
\pgfsetrectcap%
\pgfsetroundjoin%
\pgfsetlinewidth{0.501875pt}%
\definecolor{currentstroke}{rgb}{0.690196,0.690196,0.690196}%
\pgfsetstrokecolor{currentstroke}%
\pgfsetstrokeopacity{0.500000}%
\pgfsetdash{}{0pt}%
\pgfpathmoveto{\pgfqpoint{1.694732in}{0.197377in}}%
\pgfpathlineto{\pgfqpoint{1.694732in}{1.737377in}}%
\pgfusepath{stroke}%
\end{pgfscope}%
\begin{pgfscope}%
\pgfsetbuttcap%
\pgfsetroundjoin%
\definecolor{currentfill}{rgb}{0.000000,0.000000,0.000000}%
\pgfsetfillcolor{currentfill}%
\pgfsetlinewidth{0.501875pt}%
\definecolor{currentstroke}{rgb}{0.000000,0.000000,0.000000}%
\pgfsetstrokecolor{currentstroke}%
\pgfsetdash{}{0pt}%
\pgfsys@defobject{currentmarker}{\pgfqpoint{0.000000in}{0.000000in}}{\pgfqpoint{0.000000in}{0.041667in}}{%
\pgfpathmoveto{\pgfqpoint{0.000000in}{0.000000in}}%
\pgfpathlineto{\pgfqpoint{0.000000in}{0.041667in}}%
\pgfusepath{stroke,fill}%
}%
\begin{pgfscope}%
\pgfsys@transformshift{1.694732in}{0.197377in}%
\pgfsys@useobject{currentmarker}{}%
\end{pgfscope}%
\end{pgfscope}%
\begin{pgfscope}%
\definecolor{textcolor}{rgb}{0.000000,0.000000,0.000000}%
\pgfsetstrokecolor{textcolor}%
\pgfsetfillcolor{textcolor}%
\pgftext[x=1.694732in,y=0.148766in,,top]{\color{textcolor}{\rmfamily\fontsize{8.000000}{9.600000}\selectfont\catcode`\^=\active\def^{\ifmmode\sp\else\^{}\fi}\catcode`\%=\active\def
\end{pgfscope}%
\begin{pgfscope}%
\pgfpathrectangle{\pgfqpoint{0.249462in}{0.197377in}}{\pgfqpoint{2.325000in}{1.540000in}}%
\pgfusepath{clip}%
\pgfsetrectcap%
\pgfsetroundjoin%
\pgfsetlinewidth{0.501875pt}%
\definecolor{currentstroke}{rgb}{0.690196,0.690196,0.690196}%
\pgfsetstrokecolor{currentstroke}%
\pgfsetstrokeopacity{0.500000}%
\pgfsetdash{}{0pt}%
\pgfpathmoveto{\pgfqpoint{2.265985in}{0.197377in}}%
\pgfpathlineto{\pgfqpoint{2.265985in}{1.737377in}}%
\pgfusepath{stroke}%
\end{pgfscope}%
\begin{pgfscope}%
\pgfsetbuttcap%
\pgfsetroundjoin%
\definecolor{currentfill}{rgb}{0.000000,0.000000,0.000000}%
\pgfsetfillcolor{currentfill}%
\pgfsetlinewidth{0.501875pt}%
\definecolor{currentstroke}{rgb}{0.000000,0.000000,0.000000}%
\pgfsetstrokecolor{currentstroke}%
\pgfsetdash{}{0pt}%
\pgfsys@defobject{currentmarker}{\pgfqpoint{0.000000in}{0.000000in}}{\pgfqpoint{0.000000in}{0.041667in}}{%
\pgfpathmoveto{\pgfqpoint{0.000000in}{0.000000in}}%
\pgfpathlineto{\pgfqpoint{0.000000in}{0.041667in}}%
\pgfusepath{stroke,fill}%
}%
\begin{pgfscope}%
\pgfsys@transformshift{2.265985in}{0.197377in}%
\pgfsys@useobject{currentmarker}{}%
\end{pgfscope}%
\end{pgfscope}%
\begin{pgfscope}%
\definecolor{textcolor}{rgb}{0.000000,0.000000,0.000000}%
\pgfsetstrokecolor{textcolor}%
\pgfsetfillcolor{textcolor}%
\pgftext[x=2.265985in,y=0.148766in,,top]{\color{textcolor}{\rmfamily\fontsize{8.000000}{9.600000}\selectfont\catcode`\^=\active\def^{\ifmmode\sp\else\^{}\fi}\catcode`\%=\active\def
\end{pgfscope}%
\begin{pgfscope}%
\pgfpathrectangle{\pgfqpoint{0.249462in}{0.197377in}}{\pgfqpoint{2.325000in}{1.540000in}}%
\pgfusepath{clip}%
\pgfsetrectcap%
\pgfsetroundjoin%
\pgfsetlinewidth{0.501875pt}%
\definecolor{currentstroke}{rgb}{0.690196,0.690196,0.690196}%
\pgfsetstrokecolor{currentstroke}%
\pgfsetstrokeopacity{0.500000}%
\pgfsetdash{}{0pt}%
\pgfpathmoveto{\pgfqpoint{0.249462in}{0.197377in}}%
\pgfpathlineto{\pgfqpoint{2.574462in}{0.197377in}}%
\pgfusepath{stroke}%
\end{pgfscope}%
\begin{pgfscope}%
\pgfsetbuttcap%
\pgfsetroundjoin%
\definecolor{currentfill}{rgb}{0.000000,0.000000,0.000000}%
\pgfsetfillcolor{currentfill}%
\pgfsetlinewidth{0.501875pt}%
\definecolor{currentstroke}{rgb}{0.000000,0.000000,0.000000}%
\pgfsetstrokecolor{currentstroke}%
\pgfsetdash{}{0pt}%
\pgfsys@defobject{currentmarker}{\pgfqpoint{0.000000in}{0.000000in}}{\pgfqpoint{0.041667in}{0.000000in}}{%
\pgfpathmoveto{\pgfqpoint{0.000000in}{0.000000in}}%
\pgfpathlineto{\pgfqpoint{0.041667in}{0.000000in}}%
\pgfusepath{stroke,fill}%
}%
\begin{pgfscope}%
\pgfsys@transformshift{0.249462in}{0.197377in}%
\pgfsys@useobject{currentmarker}{}%
\end{pgfscope}%
\end{pgfscope}%
\begin{pgfscope}%
\definecolor{textcolor}{rgb}{0.000000,0.000000,0.000000}%
\pgfsetstrokecolor{textcolor}%
\pgfsetfillcolor{textcolor}%
\pgftext[x=0.050000in, y=0.158796in, left, base]{\color{textcolor}{\rmfamily\fontsize{8.000000}{9.600000}\selectfont\catcode`\^=\active\def^{\ifmmode\sp\else\^{}\fi}\catcode`\%=\active\def
\end{pgfscope}%
\begin{pgfscope}%
\pgfpathrectangle{\pgfqpoint{0.249462in}{0.197377in}}{\pgfqpoint{2.325000in}{1.540000in}}%
\pgfusepath{clip}%
\pgfsetrectcap%
\pgfsetroundjoin%
\pgfsetlinewidth{0.501875pt}%
\definecolor{currentstroke}{rgb}{0.690196,0.690196,0.690196}%
\pgfsetstrokecolor{currentstroke}%
\pgfsetstrokeopacity{0.500000}%
\pgfsetdash{}{0pt}%
\pgfpathmoveto{\pgfqpoint{0.249462in}{0.549102in}}%
\pgfpathlineto{\pgfqpoint{2.574462in}{0.549102in}}%
\pgfusepath{stroke}%
\end{pgfscope}%
\begin{pgfscope}%
\pgfsetbuttcap%
\pgfsetroundjoin%
\definecolor{currentfill}{rgb}{0.000000,0.000000,0.000000}%
\pgfsetfillcolor{currentfill}%
\pgfsetlinewidth{0.501875pt}%
\definecolor{currentstroke}{rgb}{0.000000,0.000000,0.000000}%
\pgfsetstrokecolor{currentstroke}%
\pgfsetdash{}{0pt}%
\pgfsys@defobject{currentmarker}{\pgfqpoint{0.000000in}{0.000000in}}{\pgfqpoint{0.041667in}{0.000000in}}{%
\pgfpathmoveto{\pgfqpoint{0.000000in}{0.000000in}}%
\pgfpathlineto{\pgfqpoint{0.041667in}{0.000000in}}%
\pgfusepath{stroke,fill}%
}%
\begin{pgfscope}%
\pgfsys@transformshift{0.249462in}{0.549102in}%
\pgfsys@useobject{currentmarker}{}%
\end{pgfscope}%
\end{pgfscope}%
\begin{pgfscope}%
\definecolor{textcolor}{rgb}{0.000000,0.000000,0.000000}%
\pgfsetstrokecolor{textcolor}%
\pgfsetfillcolor{textcolor}%
\pgftext[x=0.050000in, y=0.510522in, left, base]{\color{textcolor}{\rmfamily\fontsize{8.000000}{9.600000}\selectfont\catcode`\^=\active\def^{\ifmmode\sp\else\^{}\fi}\catcode`\%=\active\def
\end{pgfscope}%
\begin{pgfscope}%
\pgfpathrectangle{\pgfqpoint{0.249462in}{0.197377in}}{\pgfqpoint{2.325000in}{1.540000in}}%
\pgfusepath{clip}%
\pgfsetrectcap%
\pgfsetroundjoin%
\pgfsetlinewidth{0.501875pt}%
\definecolor{currentstroke}{rgb}{0.690196,0.690196,0.690196}%
\pgfsetstrokecolor{currentstroke}%
\pgfsetstrokeopacity{0.500000}%
\pgfsetdash{}{0pt}%
\pgfpathmoveto{\pgfqpoint{0.249462in}{0.900827in}}%
\pgfpathlineto{\pgfqpoint{2.574462in}{0.900827in}}%
\pgfusepath{stroke}%
\end{pgfscope}%
\begin{pgfscope}%
\pgfsetbuttcap%
\pgfsetroundjoin%
\definecolor{currentfill}{rgb}{0.000000,0.000000,0.000000}%
\pgfsetfillcolor{currentfill}%
\pgfsetlinewidth{0.501875pt}%
\definecolor{currentstroke}{rgb}{0.000000,0.000000,0.000000}%
\pgfsetstrokecolor{currentstroke}%
\pgfsetdash{}{0pt}%
\pgfsys@defobject{currentmarker}{\pgfqpoint{0.000000in}{0.000000in}}{\pgfqpoint{0.041667in}{0.000000in}}{%
\pgfpathmoveto{\pgfqpoint{0.000000in}{0.000000in}}%
\pgfpathlineto{\pgfqpoint{0.041667in}{0.000000in}}%
\pgfusepath{stroke,fill}%
}%
\begin{pgfscope}%
\pgfsys@transformshift{0.249462in}{0.900827in}%
\pgfsys@useobject{currentmarker}{}%
\end{pgfscope}%
\end{pgfscope}%
\begin{pgfscope}%
\definecolor{textcolor}{rgb}{0.000000,0.000000,0.000000}%
\pgfsetstrokecolor{textcolor}%
\pgfsetfillcolor{textcolor}%
\pgftext[x=0.050000in, y=0.862247in, left, base]{\color{textcolor}{\rmfamily\fontsize{8.000000}{9.600000}\selectfont\catcode`\^=\active\def^{\ifmmode\sp\else\^{}\fi}\catcode`\%=\active\def
\end{pgfscope}%
\begin{pgfscope}%
\pgfpathrectangle{\pgfqpoint{0.249462in}{0.197377in}}{\pgfqpoint{2.325000in}{1.540000in}}%
\pgfusepath{clip}%
\pgfsetrectcap%
\pgfsetroundjoin%
\pgfsetlinewidth{0.501875pt}%
\definecolor{currentstroke}{rgb}{0.690196,0.690196,0.690196}%
\pgfsetstrokecolor{currentstroke}%
\pgfsetstrokeopacity{0.500000}%
\pgfsetdash{}{0pt}%
\pgfpathmoveto{\pgfqpoint{0.249462in}{1.252552in}}%
\pgfpathlineto{\pgfqpoint{2.574462in}{1.252552in}}%
\pgfusepath{stroke}%
\end{pgfscope}%
\begin{pgfscope}%
\pgfsetbuttcap%
\pgfsetroundjoin%
\definecolor{currentfill}{rgb}{0.000000,0.000000,0.000000}%
\pgfsetfillcolor{currentfill}%
\pgfsetlinewidth{0.501875pt}%
\definecolor{currentstroke}{rgb}{0.000000,0.000000,0.000000}%
\pgfsetstrokecolor{currentstroke}%
\pgfsetdash{}{0pt}%
\pgfsys@defobject{currentmarker}{\pgfqpoint{0.000000in}{0.000000in}}{\pgfqpoint{0.041667in}{0.000000in}}{%
\pgfpathmoveto{\pgfqpoint{0.000000in}{0.000000in}}%
\pgfpathlineto{\pgfqpoint{0.041667in}{0.000000in}}%
\pgfusepath{stroke,fill}%
}%
\begin{pgfscope}%
\pgfsys@transformshift{0.249462in}{1.252552in}%
\pgfsys@useobject{currentmarker}{}%
\end{pgfscope}%
\end{pgfscope}%
\begin{pgfscope}%
\definecolor{textcolor}{rgb}{0.000000,0.000000,0.000000}%
\pgfsetstrokecolor{textcolor}%
\pgfsetfillcolor{textcolor}%
\pgftext[x=0.050000in, y=1.213972in, left, base]{\color{textcolor}{\rmfamily\fontsize{8.000000}{9.600000}\selectfont\catcode`\^=\active\def^{\ifmmode\sp\else\^{}\fi}\catcode`\%=\active\def
\end{pgfscope}%
\begin{pgfscope}%
\pgfpathrectangle{\pgfqpoint{0.249462in}{0.197377in}}{\pgfqpoint{2.325000in}{1.540000in}}%
\pgfusepath{clip}%
\pgfsetrectcap%
\pgfsetroundjoin%
\pgfsetlinewidth{0.501875pt}%
\definecolor{currentstroke}{rgb}{0.690196,0.690196,0.690196}%
\pgfsetstrokecolor{currentstroke}%
\pgfsetstrokeopacity{0.500000}%
\pgfsetdash{}{0pt}%
\pgfpathmoveto{\pgfqpoint{0.249462in}{1.604278in}}%
\pgfpathlineto{\pgfqpoint{2.574462in}{1.604278in}}%
\pgfusepath{stroke}%
\end{pgfscope}%
\begin{pgfscope}%
\pgfsetbuttcap%
\pgfsetroundjoin%
\definecolor{currentfill}{rgb}{0.000000,0.000000,0.000000}%
\pgfsetfillcolor{currentfill}%
\pgfsetlinewidth{0.501875pt}%
\definecolor{currentstroke}{rgb}{0.000000,0.000000,0.000000}%
\pgfsetstrokecolor{currentstroke}%
\pgfsetdash{}{0pt}%
\pgfsys@defobject{currentmarker}{\pgfqpoint{0.000000in}{0.000000in}}{\pgfqpoint{0.041667in}{0.000000in}}{%
\pgfpathmoveto{\pgfqpoint{0.000000in}{0.000000in}}%
\pgfpathlineto{\pgfqpoint{0.041667in}{0.000000in}}%
\pgfusepath{stroke,fill}%
}%
\begin{pgfscope}%
\pgfsys@transformshift{0.249462in}{1.604278in}%
\pgfsys@useobject{currentmarker}{}%
\end{pgfscope}%
\end{pgfscope}%
\begin{pgfscope}%
\definecolor{textcolor}{rgb}{0.000000,0.000000,0.000000}%
\pgfsetstrokecolor{textcolor}%
\pgfsetfillcolor{textcolor}%
\pgftext[x=0.050000in, y=1.565697in, left, base]{\color{textcolor}{\rmfamily\fontsize{8.000000}{9.600000}\selectfont\catcode`\^=\active\def^{\ifmmode\sp\else\^{}\fi}\catcode`\%=\active\def
\end{pgfscope}%
\begin{pgfscope}%
\pgfsetbuttcap%
\pgfsetroundjoin%
\definecolor{currentfill}{rgb}{0.000000,0.000000,0.000000}%
\pgfsetfillcolor{currentfill}%
\pgfsetlinewidth{0.501875pt}%
\definecolor{currentstroke}{rgb}{0.000000,0.000000,0.000000}%
\pgfsetstrokecolor{currentstroke}%
\pgfsetdash{}{0pt}%
\pgfsys@defobject{currentmarker}{\pgfqpoint{0.000000in}{0.000000in}}{\pgfqpoint{0.020833in}{0.000000in}}{%
\pgfpathmoveto{\pgfqpoint{0.000000in}{0.000000in}}%
\pgfpathlineto{\pgfqpoint{0.020833in}{0.000000in}}%
\pgfusepath{stroke,fill}%
}%
\begin{pgfscope}%
\pgfsys@transformshift{0.249462in}{0.285308in}%
\pgfsys@useobject{currentmarker}{}%
\end{pgfscope}%
\end{pgfscope}%
\begin{pgfscope}%
\pgfsetbuttcap%
\pgfsetroundjoin%
\definecolor{currentfill}{rgb}{0.000000,0.000000,0.000000}%
\pgfsetfillcolor{currentfill}%
\pgfsetlinewidth{0.501875pt}%
\definecolor{currentstroke}{rgb}{0.000000,0.000000,0.000000}%
\pgfsetstrokecolor{currentstroke}%
\pgfsetdash{}{0pt}%
\pgfsys@defobject{currentmarker}{\pgfqpoint{0.000000in}{0.000000in}}{\pgfqpoint{0.020833in}{0.000000in}}{%
\pgfpathmoveto{\pgfqpoint{0.000000in}{0.000000in}}%
\pgfpathlineto{\pgfqpoint{0.020833in}{0.000000in}}%
\pgfusepath{stroke,fill}%
}%
\begin{pgfscope}%
\pgfsys@transformshift{0.249462in}{0.373239in}%
\pgfsys@useobject{currentmarker}{}%
\end{pgfscope}%
\end{pgfscope}%
\begin{pgfscope}%
\pgfsetbuttcap%
\pgfsetroundjoin%
\definecolor{currentfill}{rgb}{0.000000,0.000000,0.000000}%
\pgfsetfillcolor{currentfill}%
\pgfsetlinewidth{0.501875pt}%
\definecolor{currentstroke}{rgb}{0.000000,0.000000,0.000000}%
\pgfsetstrokecolor{currentstroke}%
\pgfsetdash{}{0pt}%
\pgfsys@defobject{currentmarker}{\pgfqpoint{0.000000in}{0.000000in}}{\pgfqpoint{0.020833in}{0.000000in}}{%
\pgfpathmoveto{\pgfqpoint{0.000000in}{0.000000in}}%
\pgfpathlineto{\pgfqpoint{0.020833in}{0.000000in}}%
\pgfusepath{stroke,fill}%
}%
\begin{pgfscope}%
\pgfsys@transformshift{0.249462in}{0.461171in}%
\pgfsys@useobject{currentmarker}{}%
\end{pgfscope}%
\end{pgfscope}%
\begin{pgfscope}%
\pgfsetbuttcap%
\pgfsetroundjoin%
\definecolor{currentfill}{rgb}{0.000000,0.000000,0.000000}%
\pgfsetfillcolor{currentfill}%
\pgfsetlinewidth{0.501875pt}%
\definecolor{currentstroke}{rgb}{0.000000,0.000000,0.000000}%
\pgfsetstrokecolor{currentstroke}%
\pgfsetdash{}{0pt}%
\pgfsys@defobject{currentmarker}{\pgfqpoint{0.000000in}{0.000000in}}{\pgfqpoint{0.020833in}{0.000000in}}{%
\pgfpathmoveto{\pgfqpoint{0.000000in}{0.000000in}}%
\pgfpathlineto{\pgfqpoint{0.020833in}{0.000000in}}%
\pgfusepath{stroke,fill}%
}%
\begin{pgfscope}%
\pgfsys@transformshift{0.249462in}{0.637033in}%
\pgfsys@useobject{currentmarker}{}%
\end{pgfscope}%
\end{pgfscope}%
\begin{pgfscope}%
\pgfsetbuttcap%
\pgfsetroundjoin%
\definecolor{currentfill}{rgb}{0.000000,0.000000,0.000000}%
\pgfsetfillcolor{currentfill}%
\pgfsetlinewidth{0.501875pt}%
\definecolor{currentstroke}{rgb}{0.000000,0.000000,0.000000}%
\pgfsetstrokecolor{currentstroke}%
\pgfsetdash{}{0pt}%
\pgfsys@defobject{currentmarker}{\pgfqpoint{0.000000in}{0.000000in}}{\pgfqpoint{0.020833in}{0.000000in}}{%
\pgfpathmoveto{\pgfqpoint{0.000000in}{0.000000in}}%
\pgfpathlineto{\pgfqpoint{0.020833in}{0.000000in}}%
\pgfusepath{stroke,fill}%
}%
\begin{pgfscope}%
\pgfsys@transformshift{0.249462in}{0.724964in}%
\pgfsys@useobject{currentmarker}{}%
\end{pgfscope}%
\end{pgfscope}%
\begin{pgfscope}%
\pgfsetbuttcap%
\pgfsetroundjoin%
\definecolor{currentfill}{rgb}{0.000000,0.000000,0.000000}%
\pgfsetfillcolor{currentfill}%
\pgfsetlinewidth{0.501875pt}%
\definecolor{currentstroke}{rgb}{0.000000,0.000000,0.000000}%
\pgfsetstrokecolor{currentstroke}%
\pgfsetdash{}{0pt}%
\pgfsys@defobject{currentmarker}{\pgfqpoint{0.000000in}{0.000000in}}{\pgfqpoint{0.020833in}{0.000000in}}{%
\pgfpathmoveto{\pgfqpoint{0.000000in}{0.000000in}}%
\pgfpathlineto{\pgfqpoint{0.020833in}{0.000000in}}%
\pgfusepath{stroke,fill}%
}%
\begin{pgfscope}%
\pgfsys@transformshift{0.249462in}{0.812896in}%
\pgfsys@useobject{currentmarker}{}%
\end{pgfscope}%
\end{pgfscope}%
\begin{pgfscope}%
\pgfsetbuttcap%
\pgfsetroundjoin%
\definecolor{currentfill}{rgb}{0.000000,0.000000,0.000000}%
\pgfsetfillcolor{currentfill}%
\pgfsetlinewidth{0.501875pt}%
\definecolor{currentstroke}{rgb}{0.000000,0.000000,0.000000}%
\pgfsetstrokecolor{currentstroke}%
\pgfsetdash{}{0pt}%
\pgfsys@defobject{currentmarker}{\pgfqpoint{0.000000in}{0.000000in}}{\pgfqpoint{0.020833in}{0.000000in}}{%
\pgfpathmoveto{\pgfqpoint{0.000000in}{0.000000in}}%
\pgfpathlineto{\pgfqpoint{0.020833in}{0.000000in}}%
\pgfusepath{stroke,fill}%
}%
\begin{pgfscope}%
\pgfsys@transformshift{0.249462in}{0.988758in}%
\pgfsys@useobject{currentmarker}{}%
\end{pgfscope}%
\end{pgfscope}%
\begin{pgfscope}%
\pgfsetbuttcap%
\pgfsetroundjoin%
\definecolor{currentfill}{rgb}{0.000000,0.000000,0.000000}%
\pgfsetfillcolor{currentfill}%
\pgfsetlinewidth{0.501875pt}%
\definecolor{currentstroke}{rgb}{0.000000,0.000000,0.000000}%
\pgfsetstrokecolor{currentstroke}%
\pgfsetdash{}{0pt}%
\pgfsys@defobject{currentmarker}{\pgfqpoint{0.000000in}{0.000000in}}{\pgfqpoint{0.020833in}{0.000000in}}{%
\pgfpathmoveto{\pgfqpoint{0.000000in}{0.000000in}}%
\pgfpathlineto{\pgfqpoint{0.020833in}{0.000000in}}%
\pgfusepath{stroke,fill}%
}%
\begin{pgfscope}%
\pgfsys@transformshift{0.249462in}{1.076690in}%
\pgfsys@useobject{currentmarker}{}%
\end{pgfscope}%
\end{pgfscope}%
\begin{pgfscope}%
\pgfsetbuttcap%
\pgfsetroundjoin%
\definecolor{currentfill}{rgb}{0.000000,0.000000,0.000000}%
\pgfsetfillcolor{currentfill}%
\pgfsetlinewidth{0.501875pt}%
\definecolor{currentstroke}{rgb}{0.000000,0.000000,0.000000}%
\pgfsetstrokecolor{currentstroke}%
\pgfsetdash{}{0pt}%
\pgfsys@defobject{currentmarker}{\pgfqpoint{0.000000in}{0.000000in}}{\pgfqpoint{0.020833in}{0.000000in}}{%
\pgfpathmoveto{\pgfqpoint{0.000000in}{0.000000in}}%
\pgfpathlineto{\pgfqpoint{0.020833in}{0.000000in}}%
\pgfusepath{stroke,fill}%
}%
\begin{pgfscope}%
\pgfsys@transformshift{0.249462in}{1.164621in}%
\pgfsys@useobject{currentmarker}{}%
\end{pgfscope}%
\end{pgfscope}%
\begin{pgfscope}%
\pgfsetbuttcap%
\pgfsetroundjoin%
\definecolor{currentfill}{rgb}{0.000000,0.000000,0.000000}%
\pgfsetfillcolor{currentfill}%
\pgfsetlinewidth{0.501875pt}%
\definecolor{currentstroke}{rgb}{0.000000,0.000000,0.000000}%
\pgfsetstrokecolor{currentstroke}%
\pgfsetdash{}{0pt}%
\pgfsys@defobject{currentmarker}{\pgfqpoint{0.000000in}{0.000000in}}{\pgfqpoint{0.020833in}{0.000000in}}{%
\pgfpathmoveto{\pgfqpoint{0.000000in}{0.000000in}}%
\pgfpathlineto{\pgfqpoint{0.020833in}{0.000000in}}%
\pgfusepath{stroke,fill}%
}%
\begin{pgfscope}%
\pgfsys@transformshift{0.249462in}{1.340484in}%
\pgfsys@useobject{currentmarker}{}%
\end{pgfscope}%
\end{pgfscope}%
\begin{pgfscope}%
\pgfsetbuttcap%
\pgfsetroundjoin%
\definecolor{currentfill}{rgb}{0.000000,0.000000,0.000000}%
\pgfsetfillcolor{currentfill}%
\pgfsetlinewidth{0.501875pt}%
\definecolor{currentstroke}{rgb}{0.000000,0.000000,0.000000}%
\pgfsetstrokecolor{currentstroke}%
\pgfsetdash{}{0pt}%
\pgfsys@defobject{currentmarker}{\pgfqpoint{0.000000in}{0.000000in}}{\pgfqpoint{0.020833in}{0.000000in}}{%
\pgfpathmoveto{\pgfqpoint{0.000000in}{0.000000in}}%
\pgfpathlineto{\pgfqpoint{0.020833in}{0.000000in}}%
\pgfusepath{stroke,fill}%
}%
\begin{pgfscope}%
\pgfsys@transformshift{0.249462in}{1.428415in}%
\pgfsys@useobject{currentmarker}{}%
\end{pgfscope}%
\end{pgfscope}%
\begin{pgfscope}%
\pgfsetbuttcap%
\pgfsetroundjoin%
\definecolor{currentfill}{rgb}{0.000000,0.000000,0.000000}%
\pgfsetfillcolor{currentfill}%
\pgfsetlinewidth{0.501875pt}%
\definecolor{currentstroke}{rgb}{0.000000,0.000000,0.000000}%
\pgfsetstrokecolor{currentstroke}%
\pgfsetdash{}{0pt}%
\pgfsys@defobject{currentmarker}{\pgfqpoint{0.000000in}{0.000000in}}{\pgfqpoint{0.020833in}{0.000000in}}{%
\pgfpathmoveto{\pgfqpoint{0.000000in}{0.000000in}}%
\pgfpathlineto{\pgfqpoint{0.020833in}{0.000000in}}%
\pgfusepath{stroke,fill}%
}%
\begin{pgfscope}%
\pgfsys@transformshift{0.249462in}{1.516346in}%
\pgfsys@useobject{currentmarker}{}%
\end{pgfscope}%
\end{pgfscope}%
\begin{pgfscope}%
\pgfsetbuttcap%
\pgfsetroundjoin%
\definecolor{currentfill}{rgb}{0.000000,0.000000,0.000000}%
\pgfsetfillcolor{currentfill}%
\pgfsetlinewidth{0.501875pt}%
\definecolor{currentstroke}{rgb}{0.000000,0.000000,0.000000}%
\pgfsetstrokecolor{currentstroke}%
\pgfsetdash{}{0pt}%
\pgfsys@defobject{currentmarker}{\pgfqpoint{0.000000in}{0.000000in}}{\pgfqpoint{0.020833in}{0.000000in}}{%
\pgfpathmoveto{\pgfqpoint{0.000000in}{0.000000in}}%
\pgfpathlineto{\pgfqpoint{0.020833in}{0.000000in}}%
\pgfusepath{stroke,fill}%
}%
\begin{pgfscope}%
\pgfsys@transformshift{0.249462in}{1.692209in}%
\pgfsys@useobject{currentmarker}{}%
\end{pgfscope}%
\end{pgfscope}%
\begin{pgfscope}%
\pgfpathrectangle{\pgfqpoint{0.249462in}{0.197377in}}{\pgfqpoint{2.325000in}{1.540000in}}%
\pgfusepath{clip}%
\pgfsetbuttcap%
\pgfsetroundjoin%
\pgfsetlinewidth{0.501875pt}%
\definecolor{currentstroke}{rgb}{0.501961,0.501961,0.501961}%
\pgfsetstrokecolor{currentstroke}%
\pgfsetstrokeopacity{0.800000}%
\pgfsetdash{{1.850000pt}{0.800000pt}}{0.000000pt}%
\pgfpathmoveto{\pgfqpoint{0.832140in}{0.197377in}}%
\pgfpathlineto{\pgfqpoint{0.832140in}{1.737377in}}%
\pgfusepath{stroke}%
\end{pgfscope}%
\begin{pgfscope}%
\pgfpathrectangle{\pgfqpoint{0.249462in}{0.197377in}}{\pgfqpoint{2.325000in}{1.540000in}}%
\pgfusepath{clip}%
\pgfsetbuttcap%
\pgfsetroundjoin%
\pgfsetlinewidth{0.501875pt}%
\definecolor{currentstroke}{rgb}{0.501961,0.501961,0.501961}%
\pgfsetstrokecolor{currentstroke}%
\pgfsetstrokeopacity{0.800000}%
\pgfsetdash{{1.850000pt}{0.800000pt}}{0.000000pt}%
\pgfpathmoveto{\pgfqpoint{1.403393in}{0.197377in}}%
\pgfpathlineto{\pgfqpoint{1.403393in}{1.737377in}}%
\pgfusepath{stroke}%
\end{pgfscope}%
\begin{pgfscope}%
\pgfpathrectangle{\pgfqpoint{0.249462in}{0.197377in}}{\pgfqpoint{2.325000in}{1.540000in}}%
\pgfusepath{clip}%
\pgfsetbuttcap%
\pgfsetroundjoin%
\pgfsetlinewidth{0.501875pt}%
\definecolor{currentstroke}{rgb}{0.501961,0.501961,0.501961}%
\pgfsetstrokecolor{currentstroke}%
\pgfsetstrokeopacity{0.800000}%
\pgfsetdash{{1.850000pt}{0.800000pt}}{0.000000pt}%
\pgfpathmoveto{\pgfqpoint{1.974646in}{0.197377in}}%
\pgfpathlineto{\pgfqpoint{1.974646in}{1.737377in}}%
\pgfusepath{stroke}%
\end{pgfscope}%
\begin{pgfscope}%
\pgfsetrectcap%
\pgfsetmiterjoin%
\pgfsetlinewidth{0.501875pt}%
\definecolor{currentstroke}{rgb}{0.000000,0.000000,0.000000}%
\pgfsetstrokecolor{currentstroke}%
\pgfsetdash{}{0pt}%
\pgfpathmoveto{\pgfqpoint{0.249462in}{0.197377in}}%
\pgfpathlineto{\pgfqpoint{0.249462in}{1.737377in}}%
\pgfusepath{stroke}%
\end{pgfscope}%
\begin{pgfscope}%
\pgfsetrectcap%
\pgfsetmiterjoin%
\pgfsetlinewidth{0.501875pt}%
\definecolor{currentstroke}{rgb}{0.000000,0.000000,0.000000}%
\pgfsetstrokecolor{currentstroke}%
\pgfsetdash{}{0pt}%
\pgfpathmoveto{\pgfqpoint{2.574462in}{0.197377in}}%
\pgfpathlineto{\pgfqpoint{2.574462in}{1.737377in}}%
\pgfusepath{stroke}%
\end{pgfscope}%
\begin{pgfscope}%
\pgfsetrectcap%
\pgfsetmiterjoin%
\pgfsetlinewidth{0.501875pt}%
\definecolor{currentstroke}{rgb}{0.000000,0.000000,0.000000}%
\pgfsetstrokecolor{currentstroke}%
\pgfsetdash{}{0pt}%
\pgfpathmoveto{\pgfqpoint{0.249462in}{0.197377in}}%
\pgfpathlineto{\pgfqpoint{2.574462in}{0.197377in}}%
\pgfusepath{stroke}%
\end{pgfscope}%
\begin{pgfscope}%
\pgfsetrectcap%
\pgfsetmiterjoin%
\pgfsetlinewidth{0.501875pt}%
\definecolor{currentstroke}{rgb}{0.000000,0.000000,0.000000}%
\pgfsetstrokecolor{currentstroke}%
\pgfsetdash{}{0pt}%
\pgfpathmoveto{\pgfqpoint{0.249462in}{1.737377in}}%
\pgfpathlineto{\pgfqpoint{2.574462in}{1.737377in}}%
\pgfusepath{stroke}%
\end{pgfscope}%
\begin{pgfscope}%
\pgfpathrectangle{\pgfqpoint{0.249462in}{0.197377in}}{\pgfqpoint{2.325000in}{1.540000in}}%
\pgfusepath{clip}%
\pgfsetbuttcap%
\pgfsetmiterjoin%
\definecolor{currentfill}{rgb}{0.000000,0.000000,0.000000}%
\pgfsetfillcolor{currentfill}%
\pgfsetlinewidth{0.000000pt}%
\definecolor{currentstroke}{rgb}{0.000000,0.000000,0.000000}%
\pgfsetstrokecolor{currentstroke}%
\pgfsetstrokeopacity{0.000000}%
\pgfsetdash{}{0pt}%
\pgfpathmoveto{\pgfqpoint{0.355144in}{0.197377in}}%
\pgfpathlineto{\pgfqpoint{0.440832in}{0.197377in}}%
\pgfpathlineto{\pgfqpoint{0.440832in}{1.539531in}}%
\pgfpathlineto{\pgfqpoint{0.355144in}{1.539531in}}%
\pgfpathlineto{\pgfqpoint{0.355144in}{0.197377in}}%
\pgfpathclose%
\pgfusepath{fill}%
\end{pgfscope}%
\begin{pgfscope}%
\pgfpathrectangle{\pgfqpoint{0.249462in}{0.197377in}}{\pgfqpoint{2.325000in}{1.540000in}}%
\pgfusepath{clip}%
\pgfsetbuttcap%
\pgfsetmiterjoin%
\definecolor{currentfill}{rgb}{0.000000,0.000000,0.000000}%
\pgfsetfillcolor{currentfill}%
\pgfsetlinewidth{0.000000pt}%
\definecolor{currentstroke}{rgb}{0.000000,0.000000,0.000000}%
\pgfsetstrokecolor{currentstroke}%
\pgfsetstrokeopacity{0.000000}%
\pgfsetdash{}{0pt}%
\pgfpathmoveto{\pgfqpoint{0.926397in}{0.197377in}}%
\pgfpathlineto{\pgfqpoint{1.012085in}{0.197377in}}%
\pgfpathlineto{\pgfqpoint{1.012085in}{1.632271in}}%
\pgfpathlineto{\pgfqpoint{0.926397in}{1.632271in}}%
\pgfpathlineto{\pgfqpoint{0.926397in}{0.197377in}}%
\pgfpathclose%
\pgfusepath{fill}%
\end{pgfscope}%
\begin{pgfscope}%
\pgfpathrectangle{\pgfqpoint{0.249462in}{0.197377in}}{\pgfqpoint{2.325000in}{1.540000in}}%
\pgfusepath{clip}%
\pgfsetbuttcap%
\pgfsetmiterjoin%
\definecolor{currentfill}{rgb}{0.000000,0.000000,0.000000}%
\pgfsetfillcolor{currentfill}%
\pgfsetlinewidth{0.000000pt}%
\definecolor{currentstroke}{rgb}{0.000000,0.000000,0.000000}%
\pgfsetstrokecolor{currentstroke}%
\pgfsetstrokeopacity{0.000000}%
\pgfsetdash{}{0pt}%
\pgfpathmoveto{\pgfqpoint{1.497650in}{0.197377in}}%
\pgfpathlineto{\pgfqpoint{1.583338in}{0.197377in}}%
\pgfpathlineto{\pgfqpoint{1.583338in}{0.197377in}}%
\pgfpathlineto{\pgfqpoint{1.497650in}{0.197377in}}%
\pgfpathlineto{\pgfqpoint{1.497650in}{0.197377in}}%
\pgfpathclose%
\pgfusepath{fill}%
\end{pgfscope}%
\begin{pgfscope}%
\pgfpathrectangle{\pgfqpoint{0.249462in}{0.197377in}}{\pgfqpoint{2.325000in}{1.540000in}}%
\pgfusepath{clip}%
\pgfsetbuttcap%
\pgfsetmiterjoin%
\definecolor{currentfill}{rgb}{0.000000,0.000000,0.000000}%
\pgfsetfillcolor{currentfill}%
\pgfsetlinewidth{0.000000pt}%
\definecolor{currentstroke}{rgb}{0.000000,0.000000,0.000000}%
\pgfsetstrokecolor{currentstroke}%
\pgfsetstrokeopacity{0.000000}%
\pgfsetdash{}{0pt}%
\pgfpathmoveto{\pgfqpoint{2.068903in}{0.197377in}}%
\pgfpathlineto{\pgfqpoint{2.154591in}{0.197377in}}%
\pgfpathlineto{\pgfqpoint{2.154591in}{0.197377in}}%
\pgfpathlineto{\pgfqpoint{2.068903in}{0.197377in}}%
\pgfpathlineto{\pgfqpoint{2.068903in}{0.197377in}}%
\pgfpathclose%
\pgfusepath{fill}%
\end{pgfscope}%
\begin{pgfscope}%
\pgfpathrectangle{\pgfqpoint{0.249462in}{0.197377in}}{\pgfqpoint{2.325000in}{1.540000in}}%
\pgfusepath{clip}%
\pgfsetbuttcap%
\pgfsetmiterjoin%
\definecolor{currentfill}{rgb}{1.000000,0.000000,0.000000}%
\pgfsetfillcolor{currentfill}%
\pgfsetlinewidth{0.000000pt}%
\definecolor{currentstroke}{rgb}{0.000000,0.000000,0.000000}%
\pgfsetstrokecolor{currentstroke}%
\pgfsetstrokeopacity{0.000000}%
\pgfsetdash{}{0pt}%
\pgfpathmoveto{\pgfqpoint{0.440832in}{0.197377in}}%
\pgfpathlineto{\pgfqpoint{0.526520in}{0.197377in}}%
\pgfpathlineto{\pgfqpoint{0.526520in}{1.462248in}}%
\pgfpathlineto{\pgfqpoint{0.440832in}{1.462248in}}%
\pgfpathlineto{\pgfqpoint{0.440832in}{0.197377in}}%
\pgfpathclose%
\pgfusepath{fill}%
\end{pgfscope}%
\begin{pgfscope}%
\pgfpathrectangle{\pgfqpoint{0.249462in}{0.197377in}}{\pgfqpoint{2.325000in}{1.540000in}}%
\pgfusepath{clip}%
\pgfsetbuttcap%
\pgfsetmiterjoin%
\definecolor{currentfill}{rgb}{1.000000,0.000000,0.000000}%
\pgfsetfillcolor{currentfill}%
\pgfsetlinewidth{0.000000pt}%
\definecolor{currentstroke}{rgb}{0.000000,0.000000,0.000000}%
\pgfsetstrokecolor{currentstroke}%
\pgfsetstrokeopacity{0.000000}%
\pgfsetdash{}{0pt}%
\pgfpathmoveto{\pgfqpoint{1.012085in}{0.197377in}}%
\pgfpathlineto{\pgfqpoint{1.097773in}{0.197377in}}%
\pgfpathlineto{\pgfqpoint{1.097773in}{1.578173in}}%
\pgfpathlineto{\pgfqpoint{1.012085in}{1.578173in}}%
\pgfpathlineto{\pgfqpoint{1.012085in}{0.197377in}}%
\pgfpathclose%
\pgfusepath{fill}%
\end{pgfscope}%
\begin{pgfscope}%
\pgfpathrectangle{\pgfqpoint{0.249462in}{0.197377in}}{\pgfqpoint{2.325000in}{1.540000in}}%
\pgfusepath{clip}%
\pgfsetbuttcap%
\pgfsetmiterjoin%
\definecolor{currentfill}{rgb}{1.000000,0.000000,0.000000}%
\pgfsetfillcolor{currentfill}%
\pgfsetlinewidth{0.000000pt}%
\definecolor{currentstroke}{rgb}{0.000000,0.000000,0.000000}%
\pgfsetstrokecolor{currentstroke}%
\pgfsetstrokeopacity{0.000000}%
\pgfsetdash{}{0pt}%
\pgfpathmoveto{\pgfqpoint{1.583338in}{0.197377in}}%
\pgfpathlineto{\pgfqpoint{1.669026in}{0.197377in}}%
\pgfpathlineto{\pgfqpoint{1.669026in}{0.197377in}}%
\pgfpathlineto{\pgfqpoint{1.583338in}{0.197377in}}%
\pgfpathlineto{\pgfqpoint{1.583338in}{0.197377in}}%
\pgfpathclose%
\pgfusepath{fill}%
\end{pgfscope}%
\begin{pgfscope}%
\pgfpathrectangle{\pgfqpoint{0.249462in}{0.197377in}}{\pgfqpoint{2.325000in}{1.540000in}}%
\pgfusepath{clip}%
\pgfsetbuttcap%
\pgfsetmiterjoin%
\definecolor{currentfill}{rgb}{1.000000,0.000000,0.000000}%
\pgfsetfillcolor{currentfill}%
\pgfsetlinewidth{0.000000pt}%
\definecolor{currentstroke}{rgb}{0.000000,0.000000,0.000000}%
\pgfsetstrokecolor{currentstroke}%
\pgfsetstrokeopacity{0.000000}%
\pgfsetdash{}{0pt}%
\pgfpathmoveto{\pgfqpoint{2.154591in}{0.197377in}}%
\pgfpathlineto{\pgfqpoint{2.240279in}{0.197377in}}%
\pgfpathlineto{\pgfqpoint{2.240279in}{0.197377in}}%
\pgfpathlineto{\pgfqpoint{2.154591in}{0.197377in}}%
\pgfpathlineto{\pgfqpoint{2.154591in}{0.197377in}}%
\pgfpathclose%
\pgfusepath{fill}%
\end{pgfscope}%
\begin{pgfscope}%
\pgfpathrectangle{\pgfqpoint{0.249462in}{0.197377in}}{\pgfqpoint{2.325000in}{1.540000in}}%
\pgfusepath{clip}%
\pgfsetbuttcap%
\pgfsetmiterjoin%
\definecolor{currentfill}{rgb}{0.000000,0.000000,1.000000}%
\pgfsetfillcolor{currentfill}%
\pgfsetlinewidth{0.000000pt}%
\definecolor{currentstroke}{rgb}{0.000000,0.000000,0.000000}%
\pgfsetstrokecolor{currentstroke}%
\pgfsetstrokeopacity{0.000000}%
\pgfsetdash{}{0pt}%
\pgfpathmoveto{\pgfqpoint{0.583645in}{0.197377in}}%
\pgfpathlineto{\pgfqpoint{0.669333in}{0.197377in}}%
\pgfpathlineto{\pgfqpoint{0.669333in}{1.396986in}}%
\pgfpathlineto{\pgfqpoint{0.583645in}{1.396986in}}%
\pgfpathlineto{\pgfqpoint{0.583645in}{0.197377in}}%
\pgfpathclose%
\pgfusepath{fill}%
\end{pgfscope}%
\begin{pgfscope}%
\pgfpathrectangle{\pgfqpoint{0.249462in}{0.197377in}}{\pgfqpoint{2.325000in}{1.540000in}}%
\pgfusepath{clip}%
\pgfsetbuttcap%
\pgfsetmiterjoin%
\definecolor{currentfill}{rgb}{0.000000,0.000000,1.000000}%
\pgfsetfillcolor{currentfill}%
\pgfsetlinewidth{0.000000pt}%
\definecolor{currentstroke}{rgb}{0.000000,0.000000,0.000000}%
\pgfsetstrokecolor{currentstroke}%
\pgfsetstrokeopacity{0.000000}%
\pgfsetdash{}{0pt}%
\pgfpathmoveto{\pgfqpoint{1.154898in}{0.197377in}}%
\pgfpathlineto{\pgfqpoint{1.240586in}{0.197377in}}%
\pgfpathlineto{\pgfqpoint{1.240586in}{1.664043in}}%
\pgfpathlineto{\pgfqpoint{1.154898in}{1.664043in}}%
\pgfpathlineto{\pgfqpoint{1.154898in}{0.197377in}}%
\pgfpathclose%
\pgfusepath{fill}%
\end{pgfscope}%
\begin{pgfscope}%
\pgfpathrectangle{\pgfqpoint{0.249462in}{0.197377in}}{\pgfqpoint{2.325000in}{1.540000in}}%
\pgfusepath{clip}%
\pgfsetbuttcap%
\pgfsetmiterjoin%
\definecolor{currentfill}{rgb}{0.000000,0.000000,1.000000}%
\pgfsetfillcolor{currentfill}%
\pgfsetlinewidth{0.000000pt}%
\definecolor{currentstroke}{rgb}{0.000000,0.000000,0.000000}%
\pgfsetstrokecolor{currentstroke}%
\pgfsetstrokeopacity{0.000000}%
\pgfsetdash{}{0pt}%
\pgfpathmoveto{\pgfqpoint{1.726151in}{0.197377in}}%
\pgfpathlineto{\pgfqpoint{1.811839in}{0.197377in}}%
\pgfpathlineto{\pgfqpoint{1.811839in}{0.197377in}}%
\pgfpathlineto{\pgfqpoint{1.726151in}{0.197377in}}%
\pgfpathlineto{\pgfqpoint{1.726151in}{0.197377in}}%
\pgfpathclose%
\pgfusepath{fill}%
\end{pgfscope}%
\begin{pgfscope}%
\pgfpathrectangle{\pgfqpoint{0.249462in}{0.197377in}}{\pgfqpoint{2.325000in}{1.540000in}}%
\pgfusepath{clip}%
\pgfsetbuttcap%
\pgfsetmiterjoin%
\definecolor{currentfill}{rgb}{0.000000,0.000000,1.000000}%
\pgfsetfillcolor{currentfill}%
\pgfsetlinewidth{0.000000pt}%
\definecolor{currentstroke}{rgb}{0.000000,0.000000,0.000000}%
\pgfsetstrokecolor{currentstroke}%
\pgfsetstrokeopacity{0.000000}%
\pgfsetdash{}{0pt}%
\pgfpathmoveto{\pgfqpoint{2.297404in}{0.197377in}}%
\pgfpathlineto{\pgfqpoint{2.383092in}{0.197377in}}%
\pgfpathlineto{\pgfqpoint{2.383092in}{0.197377in}}%
\pgfpathlineto{\pgfqpoint{2.297404in}{0.197377in}}%
\pgfpathlineto{\pgfqpoint{2.297404in}{0.197377in}}%
\pgfpathclose%
\pgfusepath{fill}%
\end{pgfscope}%
\begin{pgfscope}%
\pgfpathrectangle{\pgfqpoint{0.249462in}{0.197377in}}{\pgfqpoint{2.325000in}{1.540000in}}%
\pgfusepath{clip}%
\pgfsetbuttcap%
\pgfsetmiterjoin%
\definecolor{currentfill}{rgb}{0.000000,0.500000,0.000000}%
\pgfsetfillcolor{currentfill}%
\pgfsetlinewidth{0.000000pt}%
\definecolor{currentstroke}{rgb}{0.000000,0.000000,0.000000}%
\pgfsetstrokecolor{currentstroke}%
\pgfsetstrokeopacity{0.000000}%
\pgfsetdash{}{0pt}%
\pgfpathmoveto{\pgfqpoint{0.669333in}{0.197377in}}%
\pgfpathlineto{\pgfqpoint{0.755021in}{0.197377in}}%
\pgfpathlineto{\pgfqpoint{0.755021in}{1.506042in}}%
\pgfpathlineto{\pgfqpoint{0.669333in}{1.506042in}}%
\pgfpathlineto{\pgfqpoint{0.669333in}{0.197377in}}%
\pgfpathclose%
\pgfusepath{fill}%
\end{pgfscope}%
\begin{pgfscope}%
\pgfpathrectangle{\pgfqpoint{0.249462in}{0.197377in}}{\pgfqpoint{2.325000in}{1.540000in}}%
\pgfusepath{clip}%
\pgfsetbuttcap%
\pgfsetmiterjoin%
\definecolor{currentfill}{rgb}{0.000000,0.500000,0.000000}%
\pgfsetfillcolor{currentfill}%
\pgfsetlinewidth{0.000000pt}%
\definecolor{currentstroke}{rgb}{0.000000,0.000000,0.000000}%
\pgfsetstrokecolor{currentstroke}%
\pgfsetstrokeopacity{0.000000}%
\pgfsetdash{}{0pt}%
\pgfpathmoveto{\pgfqpoint{1.240586in}{0.197377in}}%
\pgfpathlineto{\pgfqpoint{1.326274in}{0.197377in}}%
\pgfpathlineto{\pgfqpoint{1.326274in}{1.642576in}}%
\pgfpathlineto{\pgfqpoint{1.240586in}{1.642576in}}%
\pgfpathlineto{\pgfqpoint{1.240586in}{0.197377in}}%
\pgfpathclose%
\pgfusepath{fill}%
\end{pgfscope}%
\begin{pgfscope}%
\pgfpathrectangle{\pgfqpoint{0.249462in}{0.197377in}}{\pgfqpoint{2.325000in}{1.540000in}}%
\pgfusepath{clip}%
\pgfsetbuttcap%
\pgfsetmiterjoin%
\definecolor{currentfill}{rgb}{0.000000,0.500000,0.000000}%
\pgfsetfillcolor{currentfill}%
\pgfsetlinewidth{0.000000pt}%
\definecolor{currentstroke}{rgb}{0.000000,0.000000,0.000000}%
\pgfsetstrokecolor{currentstroke}%
\pgfsetstrokeopacity{0.000000}%
\pgfsetdash{}{0pt}%
\pgfpathmoveto{\pgfqpoint{1.811839in}{0.197377in}}%
\pgfpathlineto{\pgfqpoint{1.897527in}{0.197377in}}%
\pgfpathlineto{\pgfqpoint{1.897527in}{0.197377in}}%
\pgfpathlineto{\pgfqpoint{1.811839in}{0.197377in}}%
\pgfpathlineto{\pgfqpoint{1.811839in}{0.197377in}}%
\pgfpathclose%
\pgfusepath{fill}%
\end{pgfscope}%
\begin{pgfscope}%
\pgfpathrectangle{\pgfqpoint{0.249462in}{0.197377in}}{\pgfqpoint{2.325000in}{1.540000in}}%
\pgfusepath{clip}%
\pgfsetbuttcap%
\pgfsetmiterjoin%
\definecolor{currentfill}{rgb}{0.000000,0.500000,0.000000}%
\pgfsetfillcolor{currentfill}%
\pgfsetlinewidth{0.000000pt}%
\definecolor{currentstroke}{rgb}{0.000000,0.000000,0.000000}%
\pgfsetstrokecolor{currentstroke}%
\pgfsetstrokeopacity{0.000000}%
\pgfsetdash{}{0pt}%
\pgfpathmoveto{\pgfqpoint{2.383092in}{0.197377in}}%
\pgfpathlineto{\pgfqpoint{2.468780in}{0.197377in}}%
\pgfpathlineto{\pgfqpoint{2.468780in}{0.197377in}}%
\pgfpathlineto{\pgfqpoint{2.383092in}{0.197377in}}%
\pgfpathlineto{\pgfqpoint{2.383092in}{0.197377in}}%
\pgfpathclose%
\pgfusepath{fill}%
\end{pgfscope}%
\end{pgfpicture}%
\makeatother%
\endgroup%

%% file: Papadopoulos-BDS.bbl
\begin{thebibliography}{10}

\bibitem{zhang2023pidray}
L.~Zhang, L.~Jiang, R.~Ji, and H.~Fan, ``Pidray: A large-scale x-ray benchmark
  for real-world prohibited item detection,'' {\em International Journal of
  Computer Vision}, vol.~131, no.~12, pp.~3170--3192, 2023.

\bibitem{batsis2023illicit}
G.~Batsis, I.~Mademlis, and G.~T. Papadopoulos, ``Illicit item detection in
  x-ray images for security applications,'' in {\em 2023 IEEE Ninth
  International Conference on Big Data Computing Service and Applications
  (BigDataService)}, pp.~63--70, IEEE, 2023.

\bibitem{rafiei2023computer}
M.~Rafiei, J.~Raitoharju, and A.~Iosifidis, ``Computer vision on x-ray data in
  industrial production and security applications: A comprehensive survey,''
  {\em Ieee Access}, vol.~11, pp.~2445--2477, 2023.

\bibitem{mademlis2023visual}
I.~Mademlis, G.~Batsis, A.~A.~R. Chrysochoou, and G.~T. Papadopoulos, ``Visual
  inspection for illicit items in x-ray images using deep learning,'' in {\em
  2023 IEEE International Conference on Big Data (BigData)}, pp.~4081--4089,
  IEEE, 2023.

\bibitem{dosovitskiy2020image}
A.~Dosovitskiy, L.~Beyer, A.~Kolesnikov, D.~Weissenborn, X.~Zhai,
  T.~Unterthiner, M.~Dehghani, M.~Minderer, G.~Heigold, S.~Gelly, {\em et~al.},
  ``An image is worth 16x16 words: Transformers for image recognition at
  scale,'' {\em arXiv preprint arXiv:2010.11929}, 2020.

\bibitem{cani2024illicit}
J.~Cani, I.~Mademlis, A.~A.~R. Chrysochoou, and G.~T. Papadopoulos, ``Illicit
  object detection in x-ray images using vision transformers,'' in {\em 2024
  5th International Conference in Electronic Engineering, Information
  Technology \& Education (EEITE)}, pp.~1--6, IEEE, 2024.

\bibitem{carion2020end}
N.~Carion, F.~Massa, G.~Synnaeve, N.~Usunier, A.~Kirillov, and S.~Zagoruyko,
  ``End-to-end object detection with transformers,'' in {\em European
  conference on computer vision}, pp.~213--229, Springer, 2020.

\bibitem{roh2021sparse}
B.~Roh, J.~Shin, W.~Shin, and S.~Kim, ``Sparse detr: Efficient end-to-end
  object detection with learnable sparsity,'' {\em arXiv preprint
  arXiv:2111.14330}, 2021.

\bibitem{zhang2022dino}
H.~Zhang, F.~Li, S.~Liu, L.~Zhang, H.~Su, J.~Zhu, L.~M. Ni, and H.-Y. Shum,
  ``Dino: Detr with improved denoising anchor boxes for end-to-end object
  detection,'' {\em arXiv preprint arXiv:2203.03605}, 2022.

\bibitem{alimisis2025advances}
P.~Alimisis, I.~Mademlis, P.~Radoglou-Grammatikis, P.~Sarigiannidis, and G.~T.
  Papadopoulos, ``Advances in diffusion models for image data augmentation: A
  review of methods, models, evaluation metrics and future research
  directions,'' {\em Artificial Intelligence Review}, vol.~58, no.~4,
  pp.~1--55, 2025.

\bibitem{khan2023survey}
A.~Khan, Z.~Rauf, A.~Sohail, A.~R. Khan, H.~Asif, A.~Asif, and U.~Farooq, ``A
  survey of the vision transformers and their cnn-transformer based variants,''
  {\em Artificial Intelligence Review}, vol.~56, no.~Suppl 3, pp.~2917--2970,
  2023.

\bibitem{hendria2023combining}
W.~F. Hendria, Q.~T. Phan, F.~Adzaka, and C.~Jeong, ``Combining transformer and
  cnn for object detection in uav imagery,'' {\em ICT Express}, vol.~9, no.~2,
  pp.~258--263, 2023.

\bibitem{li2022next}
J.~Li, X.~Xia, W.~Li, H.~Li, X.~Wang, X.~Xiao, R.~Wang, M.~Zheng, and X.~Pan,
  ``{Next-ViT}: Next generation vision transformer for efficient deployment in
  realistic industrial scenarios,'' {\em arXiv preprint arXiv:2207.05501},
  2022.

\bibitem{wu2023object}
J.~Wu, X.~Xu, and J.~Yang, ``Object detection and x-ray security imaging: A
  survey,'' {\em IEEE Access}, vol.~11, pp.~45416--45441, 2023.

\bibitem{mademlis2024invisible}
I.~Mademlis, M.~Mancuso, C.~Paternoster, S.~Evangelatos, E.~Finlay, J.~Hughes,
  P.~Radoglou-Grammatikis, P.~Sarigiannidis, G.~Stavropoulos, K.~Votis, {\em
  et~al.}, ``The invisible arms race: digital trends in illicit goods
  trafficking and ai-enabled responses,'' {\em IEEE Transactions on Technology
  and Society}, 2024.

\bibitem{HGNetv2}
{Baidu Paddle Vision Team}, {\em {HGNetv2}}, 2023.

\bibitem{Jocher_Ultralytics_YOLO_2023}
G.~Jocher, J.~Qiu, and A.~Chaurasia, {\em {Ultralytics YOLO}}, Jan. 2023.

\bibitem{zhao2024detrs}
Y.~Zhao, W.~Lv, S.~Xu, J.~Wei, G.~Wang, Q.~Dang, Y.~Liu, and J.~Chen, ``Detrs
  beat yolos on real-time object detection,'' in {\em Proceedings of the
  IEEE/CVF conference on computer vision and pattern recognition},
  pp.~16965--16974, 2024.

\bibitem{tao2022exploring}
R.~Tao, H.~Li, T.~Wang, Y.~Wei, Y.~Ding, B.~Jin, H.~Zhi, X.~Liu, and A.~Liu,
  ``Exploring endogenous shift for cross-domain detection: A large-scale
  benchmark and perturbation suppression network,'' in {\em 2022 IEEE/CVF
  Conference on Computer Vision and Pattern Recognition (CVPR)},
  pp.~21157--21167, IEEE, 2022.

\bibitem{tao2021towards}
R.~Tao, Y.~Wei, X.~Jiang, H.~Li, H.~Qin, J.~Wang, Y.~Ma, L.~Zhang, and X.~Liu,
  ``Towards real-world x-ray security inspection: A high-quality benchmark and
  lateral inhibition module for prohibited items detection,'' in {\em
  Proceedings of the IEEE/CVF international conference on computer vision},
  pp.~10923--10932, 2021.

\bibitem{wang2021towards}
B.~Wang, L.~Zhang, L.~Wen, X.~Liu, and Y.~Wu, ``Towards real-world prohibited
  item detection: A large-scale x-ray benchmark,'' in {\em Proceedings of the
  IEEE/CVF international conference on computer vision}, pp.~5412--5421, 2021.

\bibitem{wei2020occluded}
Y.~Wei, R.~Tao, Z.~Wu, Y.~Ma, L.~Zhang, and X.~Liu, ``Occluded prohibited items
  detection: An x-ray security inspection benchmark and de-occlusion attention
  module,'' in {\em Proceedings of the 28th ACM international conference on
  multimedia}, pp.~138--146, 2020.

\bibitem{jing2023yolo}
B.~Jing, P.~Duan, L.~Chen, and Y.~Du, ``Em-yolo: An x-ray
  prohibited-item-detection method based on edge and material information
  fusion,'' {\em Sensors}, vol.~23, no.~20, p.~8555, 2023.

\bibitem{han2023sc}
L.~Han, C.~Ma, Y.~Liu, J.~Jia, and J.~Sun, ``Sc-yolov8: A security check model
  for the inspection of prohibited items in x-ray images,'' {\em Electronics},
  vol.~12, no.~20, p.~4208, 2023.

\bibitem{wang2024lightweight}
Z.~Wang, X.~Wang, Y.~Shi, H.~Qi, M.~Jia, and W.~Wang, ``Lightweight detection
  method for x-ray security inspection with occlusion,'' {\em Sensors},
  vol.~24, no.~3, p.~1002, 2024.

\bibitem{wang2024improved}
A.~Wang, P.~Yuan, H.~Wu, Y.~Iwahori, and Y.~Liu, ``Improved yolov8 for
  dangerous goods detection in x-ray security images,'' {\em Electronics},
  vol.~13, no.~16, p.~3238, 2024.

\bibitem{zhang2025lightweight}
H.~Zhang, W.~Teng, X.~He, H.~Que, and Y.~Zhang, ``Lightweight prohibited items
  detection model in x-ray images based on improved yolov7-tiny,'' {\em Journal
  of the Franklin Institute}, vol.~362, no.~1, p.~107421, 2025.

\bibitem{ren2025feature}
Y.~Ren, L.~Zhao, Y.~Zhang, Y.~Liu, J.~Yang, H.~Zhang, and B.~Lei, ``Feature
  knowledge distillation-based model lightweight for prohibited item detection
  in x-ray security inspection images,'' {\em Advanced Engineering
  Informatics}, vol.~65, p.~103125, 2025.

\bibitem{meng2024transformer}
X.~Meng, H.~Feng, Y.~Ren, H.~Zhang, W.~Zou, and X.~Ouyang, ``Transformer-based
  dual-view x-ray security inspection image analysis,'' {\em Engineering
  Applications of Artificial Intelligence}, vol.~138, p.~109382, 2024.

\bibitem{wu2024eslaxdet}
J.~Wu and X.~Xu, ``Eslaxdet: A new x-ray baggage security detection framework
  based on self-supervised vision transformers,'' {\em Engineering Applications
  of Artificial Intelligence}, vol.~127, p.~107440, 2024.

\bibitem{konstantakos2025self}
S.~Konstantakos, J.~Cani, I.~Mademlis, D.~I. Chalkiadaki, Y.~M. Asano,
  E.~Gavves, and G.~T. Papadopoulos, ``Self-supervised visual learning in the
  low-data regime: a comparative evaluation,'' {\em Neurocomputing}, vol.~620,
  p.~129199, 2025.

\bibitem{wu2022tinyvit}
K.~Wu, J.~Zhang, H.~Peng, M.~Liu, B.~Xiao, J.~Fu, and L.~Yuan, ``Tinyvit: Fast
  pretraining distillation for small vision transformers,'' in {\em European
  conference on computer vision}, pp.~68--85, Springer, 2022.

\bibitem{lin2025detection}
S.~Lin, T.~Jia, H.~Wang, B.~Ma, M.~Li, and D.~Chen, ``Detection of novel
  prohibited item categories for real-world security inspection,'' {\em
  Engineering Applications of Artificial Intelligence}, vol.~144, p.~110110,
  2025.

\bibitem{huang2024adaptxray}
Y.~Huang, H.~Gao, and X.~Li, ``Adaptxray: Vision transformer and adapter in
  x-ray images for prohibited items detection,'' in {\em 2024 IEEE
  International Conference on Image Processing (ICIP)}, pp.~402--408, IEEE,
  2024.

\bibitem{lin2014microsoft}
T.-Y. Lin, M.~Maire, S.~Belongie, J.~Hays, P.~Perona, D.~Ramanan,
  P.~Doll{\'a}r, and C.~L. Zitnick, ``Microsoft coco: Common objects in
  context,'' in {\em Computer vision--ECCV 2014: 13th European conference,
  zurich, Switzerland, September 6-12, 2014, proceedings, part v 13},
  pp.~740--755, Springer, 2014.

\end{thebibliography}
